%% file: main.tex
\documentclass[table, usenames, dvipsnames]{article} % For LaTeX2e
\usepackage{iclr2022_conference,times}

\usepackage{bbm}

\usepackage{hyperref}
\usepackage{url}
\usepackage{graphicx}

\usepackage{booktabs}       % professional-quality tables
\usepackage{amsfonts}       % blackboard math symbols
\usepackage{nicefrac}       % compact symbols for 1/2, etc.
\usepackage{microtype}      % microtypography
\usepackage{hyperref}
\usepackage{multirow, makecell}
\usepackage[font=small,labelfont=bf]{caption}
\usepackage[multiple]{footmisc}
\usepackage[shortlabels]{enumitem}
\usepackage{xltabular}
\usepackage{tabularx}
\usepackage{array}
\usepackage{float}
\usepackage{anyfontsize}
\usepackage{xcolor}
\usepackage{collcell,xfp}
\usepackage{xstring}
\usepackage{tikz}
\usepackage{tipa}
\usepackage[english, russian]{babel}
\usepackage[utf8]{inputenc}
\usepackage{scalerel}
\makeatletter
\def\@BTrule[#1]{%
  \ifx\longtable\undefined
    \let\@BTswitch\@BTnormal
  \else\ifx\hline\LT@hline
    \nobreak
    \let\@BTswitch\@BLTrule
  \else
     \let\@BTswitch\@BTnormal
  \fi\fi
  \global\@thisrulewidth=#1\relax
  \ifnum\@thisruleclass=\tw@\vskip\@aboverulesep\else
  \ifnum\@lastruleclass=\z@\vskip\@aboverulesep\else
  \ifnum\@lastruleclass=\@ne\vskip\doublerulesep\fi\fi\fi
  \@BTswitch}
\makeatother

\input{tables}

\input{tables_appendix}

\usepackage{dashrule}
\usepackage{stackrel}
\usepackage[T4,T1]{fontenc}
\usepackage{subcaption}

\usepackage{colortbl}

\definecolor{lgreen}{RGB}{73,174,137}
\definecolor{lred}{RGB}{182,49,54}
\definecolor{lorange}{RGB}{255, 128, 0}
\definecolor{lblue}{RGB}{0, 0, 255}
\definecolor{ao(english)}{rgb}{0.0, 0.5, 0.0}
\definecolor{cadmiumgreen}{rgb}{0.0, 0.42, 0.24}
\colorlet{scale10}{lgreen!100}
\colorlet{scale9}{lgreen!60}
\colorlet{scale8}{lgreen!40}
\colorlet{scale7}{lgreen!20}
\colorlet{scale6}{lgreen!10}
\colorlet{scale5}{lred!5}
\colorlet{scale4}{lred!20}
\colorlet{scale3}{lred!40}
\colorlet{scale2}{lred!60}
\colorlet{scale1}{lred!75}

\colorlet{orangescale0}{lorange!0}
\colorlet{orangescale1}{lorange!10}
\colorlet{orangescale2}{lorange!20}
\colorlet{orangescale3}{lorange!30}
\colorlet{orangescale4}{lorange!40}
\colorlet{orangescale5}{lorange!50}
\colorlet{orangescale6}{lorange!60}
\colorlet{orangescale7}{lorange!70}
\colorlet{orangescale8}{lorange!80}
\colorlet{orangescale9}{lorange!90}
\colorlet{orangescale10}{lorange!100}

\colorlet{bluescale0}{lblue!0}
\colorlet{bluescale1}{lblue!10}
\colorlet{bluescale2}{lblue!20}
\colorlet{bluescale3}{lblue!30}
\colorlet{bluescale4}{lblue!40}
\colorlet{bluescale5}{lblue!50}
\colorlet{bluescale6}{lblue!60}
\colorlet{bluescale7}{lblue!70}
\colorlet{bluescale8}{lblue!80}
\colorlet{bluescale9}{lblue!90}
\colorlet{bluescale10}{lblue!100}

\definecolor{forestgreen}{rgb}{0.13, 0.55, 0.13}

\newcommand{\enxx}{en$\rightarrow$xx}
\newcommand{\xxen}{xx$\rightarrow$en}

\newcommand{\bleu}{\textsc{Bleu}}
\newcommand{\chrf}{\textsc{ChrF}}
\newcommand{\flores}{\textsc{Flores-200}}
\newcommand{\ntlevalset}{\textsc{Gatones}}
\newcommand{\gatitos}{\textsc{Gatitos}}

\makeatletter
\newcommand{\myfnsymbol}[1]{%
  \expandafter\@myfnsymbol\csname c@#1\endcsname
}
\newcommand{\@myfnsymbol}[1]{%
  \ifcase #1
  \or 1% 1
  \or 2% 2
  \or \TextOrMath{\textasteriskcentered}{*}% 3
  \or \TextOrMath{\textdagger}{\dagger}% 4
  \fi
}
\newcommand{\equalcontributor}{\@myfnsymbol{3}}
\newcommand{\correspondingA}{\@myfnsymbol{4}}
\makeatother

\renewcommand{\thefootnote}{\fnsymbol{footnote}}

\title{\centering \LARGE BiLex R$_{\textnormal x}$: \textnormal{Lexical Data Augmentation}  \protect\\  \textnormal{for Massively Multilingual Machine Translation}}

\author{\hspace{24mm}
Alex Jones\textsuperscript{\equalcontributor, \correspondingA}, Isaac Caswell\textsuperscript{\equalcontributor}, Ishank Saxena, Orhan Firat \\
\hspace{57mm} Google Research \\
}

\iclrfinalcopy % Uncomment for camera-ready version, but NOT for submission.
\begin{document}
\selectlanguage{english}
\maketitle

\footnotetext[1]{Equal contribution. Correspondence to \texttt{icaswell@google.com, alexjones1925@gmail.com}.}
\footnotetext[2]{Work done while interning on the Translate team at Google.}

\renewcommand{\thefootnote}{\myfnsymbol{footnote}}

\makeatletter
\def\nomarkfootnote{\xdef\@thefnmark{}\@footnotetext}
\makeatother

\begin{abstract}

Neural machine translation (NMT) has progressed rapidly over the past several years, and modern models are able to achieve relatively high quality using only monolingual text data, an approach dubbed Unsupervised Machine Translation (UNMT). However, these models still struggle in a variety of ways, including aspects of translation that for a human are the easiest---for instance, correctly translating common nouns. This work explores a cheap and abundant resource to combat this problem: bilingual lexica (\textsc{BiLex}s). We test the efficacy of bilingual lexica in a real-world set-up, on 200-language translation models trained on web-crawled text. We present several findings:  (1) using lexical data augmentation, we demonstrate sizable performance gains for unsupervised translation; (2) we compare several families of data augmentation, demonstrating that they yield similar improvements, and can be combined for even greater improvements; (3) we demonstrate the importance of carefully curated lexica over larger, noisier ones, especially with larger models; and (4) we compare the efficacy of multilingual lexicon data versus human-translated parallel data. Finally, we open-source \gatitos{}\footnote{\url{https://github.com/google-research/url-nlp/tree/main/gatitos}}, a new multilingual lexicon for 26 low-resource languages, which had the highest performance among lexica in our experiments.

\end{abstract}

\section{Introduction}

Neural machine translation (NMT) has emerged as the dominant way of training machine translation models \citep{bahdanau-etal-2015-neural}, where translation is modeled as a sequence-to-sequence task to be learned by neural networks \citep{sutskever-etal-2014-sequence}. Massively multilingual machine translation (MMMT) refers to the concept of training a single machine translation model on many languages and language pairs using a shared set of parameters, and has also seen success in recent years \citep{firat-etal-2016-multi, wu-etal-2016-google, johnson-etal-2017-googles, aharoni-etal-2019-massively, m2m-100, nllb2022, bapna-etal-2022-building, siddhant-etal-2022-towards}. Training these models typically relies on large-scale parallel corpora mined from the web \citep{resnik-smith-2003-web, uszkoreit-etal-2010-large, espla-gomis-2009-bitextor, banon-etal-2020-paracrawl}.

However, beyond the traditional technique of training NMT models with human-translated parallel texts, a number of other strategies have shown success recently, especially on lower-resource languages. One of these techniques is self-supervised training using monolingual corpora \citep{siddhant-etal-2020-leveraging, cheng-etal-2021-self}. With this approach, NMT models are pretrained or jointly trained on a self-supervised task with monolingual data, such as the MASS \citep{song-etal-2019-mass} or BART \citep{lewis-etal-2020-bart, liu-etal-2020-multilingual-denoising} tasks, as well as the usual neural machine translation task. This training regime can aid the model in performing zero-shot translation \citep{bapna-etal-2022-building, siddhant-etal-2022-towards}, in cases where a language has monolingual data but no parallel data. Moreover, both the self-supervised task and the supervised MT task can be modeled as neural sequence-to-sequence (Seq2Seq) problems, meaning a single Seq2Seq model can be used for training on both tasks.

Another technique that has shown success in training MMMT models is bitext mining \citep{artetxe-etal-2018-massively, schwenk-etal-2021-wikimatrix, schwenk-etal-2021-ccmatrix, heffernan-etal-2022-bitext}, in which parallel or nearly-parallel sentence pairs are mined from sets of monolingual corpora using a cross-lingual sentence encoder that has been trained to generate language-agnostic embeddings of sentences in different languages. This approach has proven useful in expanding MT models to hundreds more languages and thousands of language pairs \citep{m2m-100, nllb2022}.

Other techniques that have proven useful for low-resource MT include back-translation \citep{sennrich-etal-2016-improving, caswell-etal-2019-tagged, feldman-coto-solano-2020-neural} and the incorporation of language models into MT training \citep{gulcehre-etal-2017-on, baziotis-etal-2020-language, freitag-etal-2022-natural}. There has also been extensive work on training completely unsupervised MT systems using monolingual corpora only \citep{artetxe-etal-2017-unsupervised, artetxe-etal-2019-effective}. For example, \citet{artetxe-etal-2017-unsupervised} uses a combination of denoising autoencoding with pretrained cross-lingual embeddings and on-the-fly back-translation to achieve reasonable MT performance with zero parallel data.

In our work, we supplement the approach that combines supervised and self-supervised training with multilingual lexica. The motivation for using this resource is as follows. Despite the successes of the approach combining supervised and self-supervised training, cross-lingual vocabulary alignment is still highly imperfect in these models, especially for low-resource and unsupervised languages (see \citet{bapna-etal-2022-building} for examples of some common failure modes). That is, training on all languages using a shared set of parameters is insufficient to induce perfect cross-lingual vocabulary alignment. And although multilingual lexica generally do not contain sentence-level parallel information, they can greatly extend cross-lingual vocabulary coverage at the word or phrase level by providing examples of word or phrase translations. (Of course, we are not the first to experiment with using multilingual lexica to improve NMT performance, or multilingual NLP applications more generally; Section \ref{sec:related-work} gives more details.) 

In our approach, we experiment with augmenting monolingual and parallel data with translations from multilingual lexica. Using the publicly available massively multilingual lexicon Panlex \citep{kamholz-etal-2014-panlex}, we demonstrate that this added lexical data leads to small but significant gains over a baseline model on average, even for high-resource languages; and with smaller but carefully curated bilingual lexica, the gains are substantially larger. In both cases, the gains are most significant for unsupervised and low-resource languages. Our contributions are as follows:

\begin{enumerate}[nosep]
    \item We provide a thorough comparison of several lexicon-based data augmentation variants for MT, all of which are simple, generalizable, and easy to implement;
    \item We test these approaches ``in the wild'', i.e. on in a highly multilingual, web-mined data regime such as production systems tend to use, with hundreds of languages and billions of monolingual and parallel sentences;
    \item We explore the effects of lexical data quality \textit{and} quantity;
    \item We demonstrate the efficacy of bilingual lexicon-based approaches as models scale in size;
    \item We open-source the high-quality multilingual \gatitos{} lexicon for low-resource languages.
\end{enumerate}

The \textbf{tl;dr} of this paper is that bilingual lexica help low-resource and zero-shot NMT in almost all cases, and that most training-time augmentation methods have similar efficacy, and can be combined to be more effective. When scaling up to larger and more expressive models, these methods retain their efficacy, but the quality of the translated bilingual lexica becomes more important than the sheer quantity of lexical data points used for data augmentation. For instance, small, high-quality lexica like \gatitos{} show about 5x larger \chrf{} improvement than larger, noisier lexica like Panlex.

\section{Related Work}\label{sec:related-work}

A number of works have looked at using multilingual lexicon data augmentation for NMT and other NLP tasks. The first class of augmentations that we experiment with is ``codeswitching,'' where words in the source sentence are swapped out for their dictionary translations to create mixed-language sentences. This approach has been used for a range of multilingual NLP tasks, including MT \citep{reid-artetxe-2022-paradise, yang-etal-2020-csp, liu-etal-2021-continual, lin-etal-2020-pre, lin2021bilingual, pan-etal-2021-contrastive, yang-etal-2021-multilingual, kumar2022dictnmt, yu2021simple, kuwanto2021low, xia2019generalized}. Many of these, however, only look at codeswitching between the source and target languages, e.g. substituting source words with dictionary translations into the target language, or word-for-word BiLex translations of the target to make synthetic back-translated data \citep{nag2020incorporating}. \citet{ijcai2020p0533} experiment with codeswitching on NLI, sentiment classification, document classification, dialogue state tracking, and spoken language understanding, \citet{malon2021overcoming} looks at codeswitching embeddings for language modeling, and \citet{wang-etal-2022-expanding} experiment on NER, POS tagging, and dependency parsing. Another similar work is \citet{chaudhary-etal-2020-dict}, in which the MLM task is modified such that instead of predicting masked source tokens in the \textit{source} language, the authors provided language embeddings to cue the model to predict the masked tokens in a \textit{different} language instead. Codeswitching augmentations go by a variety of different names, e.g. ``dictionary denoising'' \citep{reid-artetxe-2022-paradise}, ``Random Aligned Substitution'' \citep{lin-etal-2020-pre}, or ``code-mixing''. In our paper, we will stick to the term ``codeswitching,'' though we will try to point out where an identical or similar approach has been tried under a different name.

The second class of augmentations we experiment with involves prepending lexicon translations to source sentences as additional cross-lingual signal, as instead of swapping out words in the source sentence. This approach has been tried as well for enhancing MT performance, e.g. in \citet{song-etal-2019-code, maheshwari-etal-2022-dictdis, niehues2021continuous, michon-etal-2020-integrating, xing2020look, susanto2020lexically}, and for similar tasks like language modeling \citep{yu2021dictbert}. One potential advantage this approach has over the codeswitching method is that it can be applied at inference time as well: multilingual lexicon entries can be prepended to sentence queries to steer the model toward more accurate word translations. Outside of NMT models, this lexical prompting approach has also been applied to translation with LLMs: \citet{dipmt2023} provide LLMs with dictionary translations of some of the source sentence words, which the model can use to cover gaps in its vocabulary coverage (although the authors do not experiment with truly low-resource languages). With the rise in popularity of LLMs for MT and other tasks, this is an exciting area for further research.

\section{The Problem with UNMT}

Various automatic metrics like \bleu{} and \chrf{} have demonstrated that Unsupervised Machine Translation can produce results of relatively high quality. However, such metrics hide the fact that UNMT approaches tend to make very specific and unusual mistakes. \citet{bapna-etal-2022-building} demonstrate that unsupervised models are remarkably good at aspects of translation like fluency, but frequently make mistakes by confusing distributionally similar words. When training on parallel data, it is straightforward enough to learn a mapping from a token in one language to another; when training on monolingual data only, this mapping can only be inferred from considerable amounts of context, necessitating a very fine-grained representation of its meaning, and consequently large amounts of monolingual data. This issue is especially noticeable with nouns that fall into the same semantic category, like animal names. For instance, \citet{bapna-etal-2022-building} demonstrate how their models, despite getting good \chrf{} scores, translate `lion' as `hyena', `snake', `rabbit', and `seizures', depending on the language.

The interesting thing about these errors is that in some sense they are the easiest to fix. Any student of a language with a pocket dictionary handy could correct many of these mistakes.
And even a traditional machine translation system -- especially a phrase-based one -- would rarely if ever mistranslate ``cat'' as ``dog''. This error mode is the primary motivation for incorporating bilingual lexica into translation systems. For this reason, this paper focuses on unsupervised and slightly supervised (low-resource) languages.

\insertgroundingmistakes

\section{Training data}

Our models were trained entirely on web-mined, sentence-level text data.

\subsection{Monolingual data} \label{sec:mono-data}

The models used in this paper build off the existing data used by \citet{bapna-etal-2022-building}, which in turn uses the specialized low-resource data crawling techniques described in \citet{caswell2020language}. However, to make rapid training and development possible, we subsampled the monolingual data for the 100 highest-resource languages to 10\% of its original size. In sum, this totaled about 4B sentences, or about 80B tokens. For the large model experiments in section \ref{sec:bigger_models} of the Appendix, we used the full (not subsampled) data, totalling 27B sentences (540B tokens). All 208 languages in our experiments had monolingual data.

\subsection{Parallel data} \label{parallel-data}

The parallel data is also the same as that from \citet{bapna-etal-2022-building}, which itself is a slightly extended version of the corpus described in \citet{arivazhagan2019massively}. The present paper extends it slightly further with the addition of small amounts of parallel data for some lower-resource languages (on the order of 1,000 - 5,000 website-template tokens). The parallel data tend to be much noisier than the monolingual data, and misalignments, boilerplate, and nonlinguistic content are common. All parallel data are sampled to 10\% of their original size, resulting in 9B parallel sentences (162M tokens) into English and the same number out of English, as well as 700K non-English-centric sentence-pairs. The large models in Section \ref{sec:bigger_models} use the full dataset, which is ten times larger than this. 

\subsection{Multilingual lexica} \label{lex-data}

\subsection{\gatitos{}} \label{gatitos}

The \gatitos{} dataset is a new dataset open-sourced in this paper. It consists of 4000 short English segments translated into a 26 very low-resource languages. The English source text is a mixture of words from a variety of sources, including frequent tokens in the English language, words for numbers, months, days of the week, Swadesh words, names of the languages themselves (including the endonym), and a few short sentences. The tokens were manually reviewed by the authors of this paper to ensure they looked reasonable\footnote{Exactly two tokens snuck into the dataset that are definitively odd, namely `simp' and `sh'}. As the name implies, this dataset is mostly very short, consisting of 93\% single tokens. There are also some short sentences, though only 23 entries have over 5 tokens. We hope this dataset will complement existing publicly available multilingual lexicons like MUSE \citep{conneau2017word, lample2017unsupervised}

\subsection{Panlex} \label{panlex}

Panlex \citep{kamholz-etal-2014-panlex} is a free, open-access massive online database consisting of word and phrase translations for $5000+$ languages, sourced from $2500$ individual dictionaries. Panlex contains $\approx 1.3$B translations across all language pairs. For our experiments, we use a subset of the Panlex database covering 177 languages and containing $66$M word pairs. Languages were chosen largely by the availability of eval sets; details given in Appendix \ref{appendix:language-rationale}.

\section{Evaluation} \label{eval}

We use two translation evaluation sets: \flores{} \citep{nllb2022, flores101, guzman-etal-2019-two}, an open-sourced evaluation set consisting of 2009 English Wikipedia sentences translated by humans into 200 languages, and \ntlevalset{} (Google AuTOmatic NTL Eval Set), an in-house evaluation set of 1,200 English sentences translated into various languages, for comparison with \citet{bapna-etal-2022-building}. We use the SacreBLEU \citep{post-2018-call} implementation of \chrf{} \footnote{signature \texttt{nrefs:1|case:mixed|eff:yes|nc:6|nw:0|space:no|version:2.3.1}} for our evaluation metric. Higher-quality, embedding-based metrics like \textsc{Bleurt} \citep{sellam-2020-bleurt} are not available for these languages, and token-based metrics like \bleu{} are especially questionable on highly-inflecting languages, as explored in \citet{bapna-etal-2022-building} Furthermore, \chrf{} has generally been found to correlate better with human judgement than \bleu{}, even for high-resource languages \citep{kocmi-etal-2021-ship,freitag-2022-stop}

For this study, we only evaluate on English-centric directions. The reason for this is that, although both evaluation sets are multi-way parallel, they are both also English-original. Therefore, using either of them for non-English-centric evaluation is not especially meaningful -- and specifically, such an evaluation set cannot measure any improvement that a direct model would have over a pivoting approach (first translating to English, and then to the target). For example, any sentence with ``you'' in the English side erases the formality and gender distinctions that either of the other languages may have, so if a direct model is able to correctly preserve formality, the eval set can't measure this. 

Furthermore, this study places more weight on the \enxx{} direction than the \xxen{} direction. The main reason for this is that the \xxen{} direction is generally an easier direction for models to learn (since they see so much more English text), so the \enxx{} direction is usually the limiting reagent when it comes to model quality; as a result we care more about improving this direction. Similarly, the smaller models trained in this study will lag larger models much more on the \xxen{} direction, so our results in this direction are not as meaningful.

\section{Model} \label{model}

For our experiments we use a Transformer Big encoder-decoder model \citep{vaswani-etal-2017-attention} with approximately $475$M parameters. We train each model for 400K steps on 64 TPU v2 chips. Our models assign a 40\% weight to the translation task, and a 60\% weight to the MASS task. For models augmented with a monolingual data augmentation, we split the 60\% weight on the MASS task into a 30\% weight on the augmented data and a 30\% weight on the non-augmented data. Parallel data augmentations were done in an analogous way.  For the raw-token-pair augmentation, we add in this task with a 5\% weight and shrink the other weights accordingly. We use a task-specific token for each of the six tasks the models may see, namely translation, MASS, GlowupMono, GlowupParallel, CodeswitchMono, and CodeswitchParallel.

\section{Methods} \label{methods}

In this paper, we divide our augmentation approaches into two classes: ``codeswitching'' approaches, which involve substituting source sentence words for their dictionary translations, and ``\texttt{GLOWUP},'' (Guiding Lexical Output With Understandable Prompts) which entails prepending dictionary translations of source words to source sentences. The main difference between these approaches is whether dictionary translations are substituted for source text (in the case of codeswitching) or added to the sentence (in the case of \texttt{GLOWUP}). As a third augmentation, we experiment with training on raw lexicon token pairs directly, treating them like any other parallel data.

The novelty of our contribution lies not so much in any one of our methods, but rather in (1) the application of these methods to unsupervised Machine Translation; (2) the number of methods we apply in controlled experiments to discover the best augmentation, (3) the scale of our experiments, in terms of number of languages, data quantity, and model capacity; and (4) the application of these methods to ``in the wild'' web-crawled data. While a variety of papers (e.g. \citet{reid-artetxe-2022-paradise, yang-etal-2020-csp}) have explored specific augmentations on particular language pairs, we believe our paper is the first to undertake a rigorous comparison of different augmentation strategies across hundreds of languages in a real-world setting.

\subsection{Codeswitching} \label{codeswitching}

In our ``codeswitching'' augmentation strategy, words in the source sentence are swapped out for their dictionary translations to create mixed-language sentences. We experiment with this augmentation on both monolingual and parallel data. The details of this method are described below.

\subsubsection{Multilingual codeswitching autoencoding (MCA)} \label{mca}

Our multilingual codeswitching autoencoding (MCA) approach is similar to the ``dictionary denoising'' objective in \citet{reid-artetxe-2022-paradise}. Let $D$ represent a multilingual lexicon containing word or phrase translation pairs for many languages. Given a source sentence $x = (x_{1}, x_{2}, . . ., x_{n})$ from monolingual corpus $X_{mono}$, we substitute each token in $x$ for its dictionary translation with probability $p_{tr} = 0.4$.  (More implementation details can be seen in Appendix Section \ref{token-sampling})

Note that \citet{reid-artetxe-2022-paradise} also apply additional noise to $x$ on top of codeswitching, along the lines of (m)BART \citep{lewis-etal-2020-bart, liu-etal-2020-multilingual-denoising}. For simplicity and so we can better examine the effects of lexicon information in isolation, we do not do this. Furthermore, while MCA indeed creates ``noisy'' codeswitched sentences, we note that word translations provide a meaningful cross-lingual signal for the model in a way that other noising functions (e.g. deletion, random word substitution) do not, as these augmentations corrupt the source sentence without adding additional useful information. Although models have been shown to learn from these denoising tasks, we use only lexical augmentation because it more directly furthers our ends of improving cross-lingual vocabulary alignment across many languages.

It should also be noted that because we are choosing dictionary translations at random without attempting to do any sort of word sense disambiguation beforehand, some of the translations will actually express the wrong sense of the source word (e.g. translating the verb ``fool'' as ``sot'' in French, even though the correct translation should be ``tromper''). We do not try to avoid this phenomenon because we believe this is an acceptable type of source-side noise, and may actually \textit{help} our models learn to perform word sense disambiguation implicitly.

\subsubsection{Codeswitching MT} \label{codeswitch-parallel}

Our codeswitching MT task is essentially the same approach as described in Section \ref{mca}, except it applies to parallel rather than monolingual data. Given a source sentence $x$ from parallel corpus $X_{parallel}$, we perform the identical procedure described in section Section \ref{mca} to obtain multilingual codeswitched sentence $x'$. We then train the model on the translation task using sentence pairs $(x', y)$, where $(x, y)$ is a sentence pair in $X_{parallel}$. This method is effectively identical to the Random Aligned Substitution method proposed in \citet{lin-etal-2020-pre}. As with MCA, we use $p_{tr} = 0.4$ and apply the augmentation on half the available parallel data.

\begin{figure}
    \centering
    \includegraphics[width=1\textwidth]{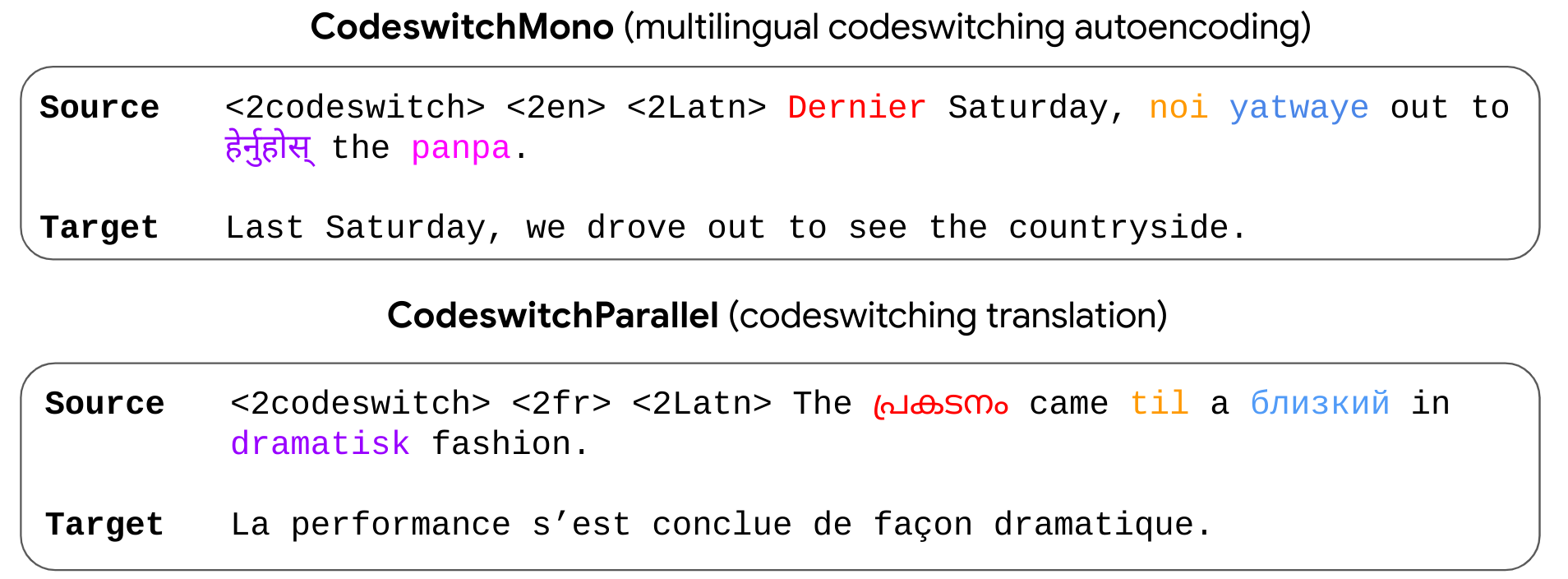}
    \caption{Examples of the monolingual (top) and parallel (bottom) codeswitching augmentation strategies. In both cases, random tokens in the source sentence are replaced with their translations in a random language from a multilingual lexicon. Color coding is used to indicate which source words have been swapped for their dictionary translation. The different colors are used simply to point out the fact that the words are from codeswitched words in each sentence come from different languages.}
    \label{fig:codeswitch_examples}
\end{figure}

\subsection{Lexical prompting (GLOWUP)}

The second class of lexical augmentations we experiment with is lexical prompting, which we call \texttt{GLOWUP}, short for Guiding Lexical Outputs With Understandable Prompts. The difference between these approaches and the codeswitching approaches described in Section \ref{codeswitching} is that instead of substituting source sentence words for their dictionary translations, we prepend those translations to the source sentences. In particular, we add $(src, transl)$ pairs to the beginning of source sentence, for some random fraction of the translatable words in that sentence. These hints can then be used to help the model guess the translation or the denoised sentence, depending on the flavor, as described below.

The \texttt{GLOWUP} task has the advantage that it can be used at inference time, can be used without retraining a model, and may be simpler to implement. However, it does result in longer and less balanced sequence lengths, which can pose problems for decoding.

\subsubsection{MASS with monolingual GLOWUP} \label{glowup-mono}

Our \texttt{GLOWUP} augmentation on monolingual data involves prepending lexical prompts to source sentences and then applying MASS on top of those sentences, after which the model attempts to reconstruct the original source sentence (including the prompts). Given multilingual lexicon $D$ and source sentence $x$, we sample some number of translatable tokens in $x$ as in the codeswitching augmentation. However, instead of substituting these tokens for their translations, we instead prepend these tokens and their translations to $x$. These $\{src, tgt\}$ translation pairs then form a ``\texttt{GLOWUP} prompt'' that is separated from the source sentence by indicator tokens. Unlike the codeswitching augmentation, we sample from a uniform distribution over $[0, k]$, where $k$ is the number of translatable tokens in $x$, when deciding how many tokens to translate. The reasons for this are described in Section \ref{glowup-mt}.

Finally, we apply MASS to mask random subsequences of $x'$, possibly including the \texttt{GLOWUP} prompt itself. We then train the model to reconstruct $x$. As with MCA, we apply the augmentation on half the monolingual data and train the rest on vanilla MASS.

\subsubsection{\texttt{GLOWUP}-MT} \label{glowup-mt}

Our \texttt{GLOWUP} augmentation on parallel data, which we call \texttt{GLOWUP}-MT, is effectively the same as the monolingual variant of the task, but without the MASS element. For a given sentence pair $(x, y)$ in the training corpus, the prompting is performed on the source sentence $x$, essentially to give it hints about how to produce $y$. The model is then trained on the translation task using $(x', y)$, with the task token \texttt{<2glowup>} instead of \texttt{<2translation>} 

One notable advantage that the \texttt{GLOWUP}-MT augmentation has over the codeswitching MT augmentation is that the \texttt{GLOWUP} variant may be applied at inference time as well. That is, given an unseen sentence, we can append source words and their translations by using a multilingual lexicon at inference time. For this reason, we opt to add word translations in the target language only, unlike the other augmentations discussed where we have added translations in multiple, random languages per sentence.

\begin{figure}
    \centering
    \includegraphics[width=1\textwidth]{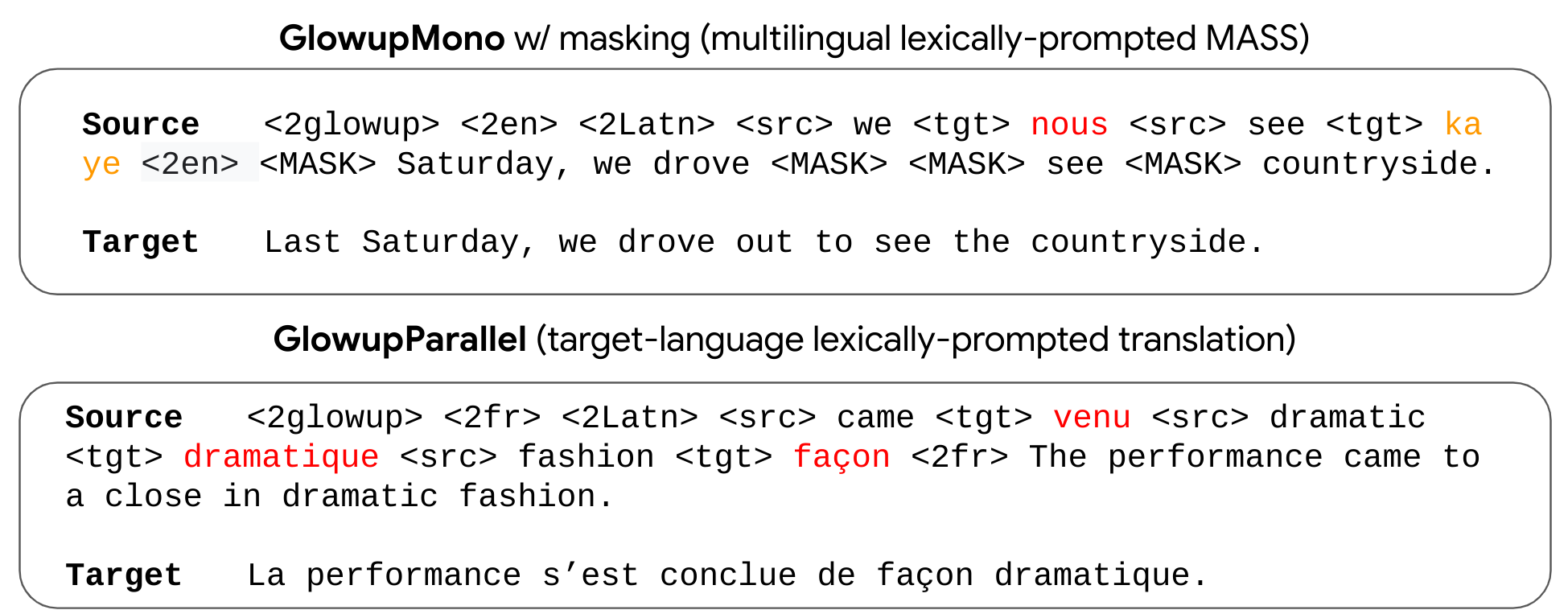}
    \caption{Examples of the monolingual (top) and parallel (bottom) \texttt{GLOWUP} augmentation strategies. In both cases, random tokens in the source sentence are prepended to the source sentence, along with their translations in a random language from a multilingual lexicon. As in Figure \ref{fig:codeswitch_examples}, differential color coding is done to draw attention to dictionary translations in different languages.}
    \label{fig:glowup_examples}
\end{figure}

\section{Experiments}

\subsection{Training regimes}

In our experiments, we train models with various combinations of the augmentations outlined above, as well as a baseline (with no data augmentation of any kind) and a model where we simply provide word pairs from the lexicon as additional parallel data. The details of our training regimes are discussed below.

\subsubsection{Baseline} \label{baseline}

We first train a baseline model with no data augmentation using the monolingual and parallel data described in Sections \ref{sec:mono-data} and \ref{parallel-data}, respectively. This model is essentially a smaller version of the model trained in \citet{bapna-etal-2022-building}. Using the model setup described in Section \ref{model}, we train the model on the MT task using all available parallel data and on the MASS task using all available monolingual data. Note that because this model is smaller and uses substantially less data than the one in \citet{bapna-etal-2022-building}, the performance is predictably not as high. We use this smaller model for the purposes of running experiments more quickly and efficiently, although we understand the relative gains of our approaches may not be identical with those of a larger model. For the same reasons, we also do not perform back-translation as done in \citet{bapna-etal-2022-building}. Larger models, which have $1.6$B parameters and are trained on substantially more parallel and monolingual data, are explored in Section \ref{sec:bigger_models}.

\subsubsection{Token-pair-only model (GatiPanlexTokenPairs)} \label{GatiPanlex}

In addition to the baseline model, we also experiment with the extremely simple approach of providing raw word pairs from multilingual lexica to the model as additional parallel data. That is, given a dictionary entry $s$ and its translation $t$, we provide the model with a ``sentence'' pair of the form $($\texttt{<2translation> <2lang> <2script>} $s, t)$. We use all $66$M token pairs from the Panlex subset described in Section \ref{panlex}, which covers $177$ of the $208$ languages present in the monolingual and parallel data, as well as all 400K token pairs from the \gatitos{} dataset described in section \ref{gatitos}, covering 26 languages. We call this token-pair baseline GatiPanlexTokenPairs.

\subsubsection{Single augmentation models}

We also train models on each of the augmentations described in Section \ref{methods}. As noted in that section, we only augment half the relevant data (monolingual, parallel, or both) before training each of these models, leaving the other half to be trained identically to the baseline (i.e. joint training on the MT task and MASS). Our naming conventions for our models are as follows:
\begin{enumerate}[nosep]
    \item \textbf{CodeswitchMono} is augmented with Monolingual Codeswitching (Section \ref{mca})
    \item \textbf{CodeswitchParallel} is augmented with Parallel Codeswitching (Section \ref{codeswitch-parallel})
    \item \textbf{GlowupMono} is augmented with Monolingual \texttt{GLOWUP} (Section \ref{glowup-mono})
    \item \textbf{GlowupParallel} is augmented with Parallel \texttt{GLOWUP} (Section \ref{glowup-mt})

\end{enumerate}

\subsubsection{Dual augmentation models}

We also train some models on a combination of data augmentation approaches to see whether a hybrid strategy may work better than any single augmentation. We train the following additional models:
\begin{enumerate}[nosep]

    \item \textbf{CodeswitchMonoParallel} and \textbf{GlowupMonoParallel} are identical to the CodeswitchMono and GlowupMono models, but also add the corresponding parallel augmentations.
    \item \textbf{CodeswitchMonoParallelGatiPanlex} and \textbf{GlowupMonoParallelGatiPanlex} take these models and further add the raw token pair objective (see Section \ref{GatiPanlex}).
 
\end{enumerate}

We leave experimentation with a hybrid codeswitch-\texttt{GLOWUP} approach (e.g. CodeswitchMonoGlowupMono) for future work.

\section{Results}

We evaluate all our models on the \flores{} dataset \citep{nllb2022, flores101, guzman-etal-2019-two}, which contains English-aligned parallel sentences for $200$ languages. In Appendix section \ref{appendix:gatones} we also report scores on the in-house \ntlevalset{} eval set, which is the eval set used by \citet{bapna-etal-2022-building}.

In analyzing our results across languages, we use the following resourcedness classifications:

\begin{enumerate}[nosep]
    \item \underline{H}igh-\underline{R}esource \underline{L}anguages (HRLs): $>2$B total training tokens in parallel data
    \item \underline{M}edium-\underline{R}esource \underline{L}anguages (MRLs): $360$M to $2$B training tokens in parallel data
    \item \underline{L}ow-\underline{R}esource \underline{L}anguages (LRLs): $1$ to $360$M training tokens in parallel data
    \item \underline{U}nsupe\underline{r}vised \underline{L}anguages (URLs): no parallel data
\end{enumerate}

\insertfloresenxxresults
\insertfloresxxenresults

The results relative to the baseline, in $\Delta$\chrf{}, are summarized in Tables \ref{tab:floresenxx} (\enxx{}) and \ref{tab:floresxxen} (\xxen{}). A few trends jump out. Firstly, all models trained with only monolingual data augmentations see consistent performance gains over the baseline. Conversely, models with only parallel data augmentations show performance degradations. Predictably, models mixing monolingual and parallel data augmentations fare in-between those poles.

These general trends are the same between \enxx{} and \xxen{} directions, though the gains are generally lower in the \xxen{} direction. As noted in Section \ref{eval}, this is expected, and this direction is less of a priority for translation improvements. Results on the \ntlevalset{} eval set, in Appendix Section \ref{appendix:gatones}, show the same trends, though the performance gains tend to be larger for all augmentations, with CodeSwitchMonoGatiPanlex gaining $+2.3$ \chrf{} for URLs. Interestingly, though the Codeswitch augmentation seems to have higher overall performance on unsupervised languages in the \enxx{} direction, \texttt{GLOWUP} usually has the edge for higher-resource languages and the \xxen{} direction.

Despite these trends seeming robust across models, the effect sizes are relatively small, maxing out at about $+1.5$ average $\Delta\chrf{}$ gain.  However, the picture changes dramatically when only looking at the subset of languages that has the higher-quality \gatitos{} training lexicons. For these 26 languages, every augmentation, even the parallel ones, have large performance gains. The winning augmentations for URLs remain CodeSwitchMonoGatiPanlex and GlowupMonoGatiPanlex, the former having an average gain of $+7.0 \chrf{}$ on \flores and $+8.0 \chrf{}$ on \ntlevalset{}, and the latter having $+5.6 \chrf{}$ on \flores{} and $+9.5 \chrf{}$ on \ntlevalset{}.

Finally, although it was not the first place model in any category, the GatiPanlexTokenPairs has large gains in all directions over baseline, and only falls short of the more complex augmentations by a small margin. Furthermore, when used in conjunction with either Glowup or Codeswitch, it further improves performance across all categories. This may be one of the most useful long-term findings of this study: raw token pairs perform roughly on par with all the other fancy augmentations!

\subsection{Scaling up: bigger models, more data} \label{sec:bigger_models}

Sometimes, results on smaller models do not transfer to larger models. To this end, we train larger transformer models with $1.6$B parameters using $10\times$ the parallel data and $10\times$ the high-resource monolingual data that was used for training the smaller models.

We trained three large models: a baseline, a token-pair model, and a token-pairs + CodeswitchMono model. The \chrf{} scores for these models can be seen in Table \ref{tab:bigbig}. There are a number of obvious differences between these results and the results on the smaller ($475$M parameter) models trained with $\approx \frac{1}{10}$th the data. First, the positive impact of the data augmentations is smaller in all categories, indicating that the gains previously seen from augmentations are partially washed out in the larger-data, larger model regime. On \flores{} \enxx{}, both models are very close to baseline---within the realm of noise. On \ntlevalset{} \enxx{}, they see consistent small gains of around $+0.5$ \chrf{}, much smaller than previously. For both eval sets in the \xxen{} direction, there are consistent small losses.
%with the exception of the surprising performance of GatiPanlexTokenPairsBig on unsupervised languages.

Is this the Bitter Lesson \citep{sutton_2019} getting us again? Perhaps---but the picture is less bleak than it first appears. When we look at the subset of the languages where we have a higher assurance that the bilingual lexica are higher-quality---namely, those that use \gatitos{} bilingual lexica---we still see consistent wins, even in the \xxen{} direction. For these 26 languages, all models see consistent gains, and as before, the gains are bigger on unsupervised languages, and adding in the Codeswitch augmentation gives a consistent leg up on the unsupervised languages.   

Overall, the takeaway from these experiments is that one has to ensure that the data is of high quality when applying lexical data augmentation at such a large scale. While we saw substantial improvements for many languages, these were balanced out by losses for other languages (especially those with only Panlex, but not \gatitos{}, data). 

\insertbigbig

\normalsize

\subsection{Glowup Decoding}

In principle, one of the advantages of the GlowupParallel approach is that lexica can be used at inference time. Therefore, we experimented with decoding the eval sets not with the translation task ID, but with the Glowup task ID, along with the relevant lookups from the lexica. Unfortunately, these decodes failed impressively, with performance degrading the more prompts that were included. Model decodes often had long sequences of control tokens. Further work should not disregard this direction; indeed, a variant of this likely has particular promise in the world of foundation models. The current approach likely just needs some tweaks to eliminate this sort of out-of-domain decoding errors we were seeing, but we leave an investigation of this hypothesis for future work.

\subsection{Oracles: what happens if we use trusted parallel text?}

Parallel text is much more costly to produce than bilingual lexica, but also contains many more useful signals, including examples of word usage in context. But how much more helpful is it, really, than bilingual lexica? The answer seems to be ``much more helpful''.

To measure this, we trained a model with a variety of public parallel datasources for a few of our lowest-resource languages. The datasets we used were HornMT \citep{hornMT}, SALT (Sunbird AI Language Translation) dataset \citep{akera2022machine}, FFR: Fon-French Neural Machine Translation \citep{emezue-dossou-2020-ffr}, Tatoeba \citep{tatoeba}, The Makerere MT Corpus: English to Luganda parallel corpus \citep{makereremt2021}, Commonvoice \citep{ardila-etal-2020-common}, Kencorpus: Kenyan Languages Corpus \citep{wanjawa2022kencorpus}, Chuvash-Russian parallel corpus\citep{chuvashcorpus}, Abkhaz Corpus \citep{abkhazcorpus}, Bashkir Corpus\citep{bashkircorpus}, Sprotin Faroese Corpus\citep{faroesecorpus}, Jojajovai Guarani-Spanish Parallel Corpus \citep{chiruzzo-etal-2022-jojajovai}, and the NLLB Seed data \citep{nllb2022}. We only used datasets with a license permitting commercial use, and therefore avoided other excellent datasets such as AmericasNLP \citep{mager-etal-2021-findings}. Note that these are high-quality, trusted datasets, prepared by community members -- a very different resource than the web-mined parallel data that the model is otherwise trained on.

To keep the experiment more controlled, we compared performance only on languages that had token-pair data from the \gatitos{} datasource, since this empirically seems to be of higher quality. Note that, although these languages are technically not zero-resource (unsupervised) in the baseline, they only had a few thousand short parallel sentences from a translation memory, so the findings here are similar to as if they had been on true 0-resource languages.

The results are in Table \ref{tab:vsparallel}. We compare four models: the standard baseline and GatiPanlexTokenPairs model, as well as the ``Parallel'' model (which adds the external parallel translation task with a 5\% weight) and the ``Parallel + GatiPanlexTokenPairs'' model, which uses both the token-pairs and the parallel data, with a combined weight of 5\%. On both eval sets, we see that using Bilex  yields a gain of around +3.5 \chrf{}, but using true parallel yields a much larger gain of about +10 \chrf{}. Using the bilingual lexica on top of the parallel data yields a further gain of about +0.5 \chrf{}, demonstrating that, though many of the gains have been washed out by the true parallel data, there are still modest gains to be had from bilex training.

Between these two resources, trusted parallel data is clearly more effective. However, the number of tokens (counting only the target language) is also much greater for the parallel data. On the extreme end, Tsonga (`ts') has 300 times more tokens in the parallel data than in the bilingual lexicon. In this light, it is impressive that bilex alone adds around +4 \chrf{}, and the parallel data only adds another +7 on top of that. Future work should look at quality gains conditioned on the number of tokens between parallel data and bilingual lexica.
%However, for the time being we can note that we can see impressive gains using only a small number of tokens in a bilingual lexicon-- and if one were to make sentence-level dataset of equal size to the 4,500-token \gatitos{} dataset, it would have only around 250 sentences.

\insertvsparallel

\insertregressiontab
\subsection{How many token pairs do I really need?}\label{subsec:token_pair_amount}

We examine the relationship between the number of lexical token pairs provided during training and MT performance (\chrf{}). First, we perform regressions using $\Delta\chrf{}$ over baseline as the outcome variable and three predictor variables: (1) number of Panlex entries, (2) number of \gatitos{} entries, and (3) number of monolingual sentences. We include monolingual sentences in the regression to control for it as a confound. To eliminate parallel data quantity as a confound, we only perform this analysis on URLs.

The regression results are given in Table \ref{tab:tokensregression}. As expected, both the number of Panlex word pairs for a given language (No. Panlex) and the number of \gatitos{} word pairs (No. \gatitos{}) have a positive $\beta$ coefficient. However, note that the $\beta$ for No. \gatitos{} is $\approx 3\times$ as large as the $\beta$ for No. Panlex on \flores{}, and $\approx 39\times$ as large for \ntlevalset{}. This aligns with our observation that the \gatitos{} dictionaries are more efficient for improving MT than Panlex, probably due to their higher quality.

The most practical question we can seek to answer is, \textbf{If I can spend \$X on translating tokens, how much quality increase can I expect?} To investigate this ``bang for buck'' question in a more controlled way, we observe the effects of the \gatitos{} dataset in isolation, without Panlex. We train an ``GatiTokenPairs'' model, which is identical to the ``GatiPanlexTokenPairs'' model, except the token-pair task has only \gatitos{} data. Thus, this tells us specifically what gains we can expect if we are to get 4,500 tokens' worth of a bilingual lexicon per language.
% For simplicity we don't train any \gatitos{}{}-only models with our augmentations, though by comparing the GatiPanlexTokenPairs model with the CodeswitchMonoGatiPanlex model on these languages we can infer that they would perform abot +0.7 \chrf{} better.

The results are in Table \ref{tab:bangforbuck}, reported in delta versus the baseline model for both \flores{} and the \ntlevalset{} eval set in the \enxx{} direction. The improvement for unsupervised languages is around +5.0 \chrf{} for both eval sets; the improvement for languages with some parallel data is less but still noticeable, hovering around +2 \chrf{}. The largest improvement is in Goan Konkani (+11.0 \chrf{}), with Mizo, Ilocano, and Bambara close on its heels with gains of around +8 \chrf{}. Only Maithili, which has interesting properties a a close dialect of Hindustani, sees a loss on both eval sets. The gains are not obviously related to the total number of tokens per language. 

As an aside, it is heartening that \flores{} and \ntlevalset{} seem to agree very nicely, despite their different domains (Wikipedia versus web + question-answers).

\insertbangforbuck

\section{Conclusions}

In this paper we explore the ways that that augmenting training data with bilingual lexicon information can improve the performance of machine translation models on low-resource and unsupervised languages, and open-source the \gatitos{} dataset, which leads to average gains of about +8 \chrf{} alone on unsupervised languages. We perform extensive experimentation with three main types of lexical augmentation: codeswitching, lexical prompting, and raw token-pair training. The results show that applying any of these augmentations to monolingual data yields substantial improvements, and that they can be combined for greater effect. The leader (by a small margin) is the combination of CodeswitchMono and raw token-pair training. These results hold when scaling up model and data size, but in the settings with more data and larger models, the quality of the bilingual lexica plays a relatively bigger role, and augmentation with the noisier Panlex begins to lag in quality behind the much smaller, yet higher-quality, \gatitos{} dataset.

Future work will likely want to focus on prompting foundation models with bilingual lexica. Large Language Models show promise on machine translation for high-resource languages \citep{jiao-etal-2023-chatgpt}, but their capabilities on low-resource languages have yet to be thoroughly explored. Additionally, a more thorough investigation of the trade-off between cost and quality for tiny datasets can be explored: if one only has X dollars, or X dedicated volunteers with k total hours, should they spend their time translating text, making monolingual text, or making bilingual lexica? 

\section{Acknowledgements}

We would like to thank Aditi Chaudhary, Ankur Bapna, Daan van Esch, and Machel Reid for their comments and suggestions on this project.

\newpage
\bibliography{iclr2022_conference,anthology}
\bibliographystyle{iclr2022_conference}

\newpage

\appendix

\insertexampletranslations

\section{Results on \ntlevalset{}} \label{appendix:gatones}

\insertntlenxxresults
\insertntlxxenresults

\subsection{ \ntlevalset{} Eval set}

The main paper reports the scores on the more widely-used \flores{} eval set; this section reports on the other dataset that we evaluate our models on. This is an in-house test set for the $1000$ Languages Initiative, which we call \ntlevalset{} (Google AuTOmatic NTL Eval Set). The dataset contains $63$ languages, most of which are unsupervised or low-resource (although there are a small number of MRLs).

\subsubsection{Summary}

\paragraph{\enxx{}}
The evaluation results on \ntlevalset{} for the \enxx{} direction are summarized in Table \ref{tab:ntlenxx}. The trends are mostly the same as what we saw for the \flores{} \enxx{} evaluation. Once again, the models trained with augmentation on monolingual data but not parallel data generally yielded improvements (the sole exception being GlowupMono, which decreased performance on URLs by $-0.2$ \chrf{} relative to the baseline), and the parallel data augmentations didn't do as well. CodeswitchMonoGatiPanlex emerges, even more definitively than on \flores{}, as the best model, winning all categories except MRLs (on which CodeswitchMono performs best).

However, there are some minor differences between this evaluation and the \flores{} one when we look more closely. First, although CodeswitchParallel again does quite poorly, GlowupParallel doesn't do too badly. Still, it is far from being on par with CodeswitchMonoGatiPanlex or even just GatiPanlexTokenPairs. GlowupMonoParallel doesn't do too poorly either, actually beating out GlowupMono on URLs by $1.3$ \chrf{}, but it doesn't help nearly as much as CodeswitchMonoGatiPanlex (or, for that matter, the two augmentations from which this hybrid model is trained). Overall, this part of the evaluation still provides a compelling case for CodeswitchMonoGatiPanlex as the single most effective augmentation method for URLs and LRLs.

\paragraph{\xxen{}}
The evaluation results for the \xxen{} direction are given in Table \ref{tab:ntlxxen}. The single most important takeaway from this part of the analysis is the same as it was for the \flores{} evaluation: the plain GatiPanlexTokenPairs model helps URLs the most in this direction, with a $\Delta \chrf{}$ of $+1.1$ over the baseline. Yet again, the improvements are smaller in this direction than for \enxx{}. The only other thing that stands out about this part of the evaluation is that the GlowupMono augmentation doesn't seem to be as helpful according to this test set as for the \flores{} set. Although GlowupMonoParallel and GlowupMonoGatiPanlex do reasonably well, their improvements are significantly smaller than the improvement from using GatiPanlexTokenPairs alone, and the GlowupMono augmentation by itself actually results in losses on URLs. So taking the \ntlevalset{} and \flores{} results together, it seems that adding raw token pairs as additional parallel data is the best way, out of the techniques we tried, to improve performance in the \xxen{} direction for very low-resource languages.

\section{ Effects on distributionally similar noun mistranslation}

Part of the motivation for using bilingual lexicons for unsupervised translation was to see whether we could repair the common error mode of mis-translating distributionally similar nouns. \citet{bapna-etal-2022-building} note that this error mode is particularly common for two categories of nouns: animals and colors.

To measure improvement on this phenomenon, we define the \textit{token hit-rate} as the following: for some set of desired tokens $D$, let $R_D$ be the subset of the eval set such that each reference contains at least one token in $D$. The hit-rate is then the percentage of times in $R_D$ that the model correctly generated one of the desired tokens in $D$. For instance, if the desired tokens are ``kitten'' and ``puma'',  $R_D$ is the set of references containing one of these words, e.g. ``The \textbf{kitten} lies'' and ``A \textbf{Puma} eats hot chip''. If the model produces ``\textbf{kitten} lie on floor'' and ``\textbf{Crocodile} charge they phone'' from the corresponding sources, it has a hit-rate of 50\%, since it correctly got ``kitten'', but not ``puma''.

Table \ref{tab:hitrate} looks at the token hit-rate for the models \textbf{BaselineBig} and \textbf{CodeswitchMonoGatiPanlexBig}, for the categories of animals occurring in \gatitos{} (\textit{bear, bee, bird, butterfly, cat, chicken, deer, dog, elephant, fish, frog, goat, horse, insect, lion, monkey, parrot, pig, rabbit, sheep, snail, snake, tiger, turkey, turtle}), animals NOT appearing in \gatitos{} (\textit{ant, antelope, buffalo, cheetah, crocodile, dolphin, dormouse, gorilla, jellyfish, koala, leopard, moose, mosquito, newt, ocelot, otter, reindeer, robin, scorpion, shark, sloth, spider, springbok, tortoise, velociraptor}), colors (\textit{black, white, red, blue, yellow, green, purple, orange, grey}), and numbers (\textit{one, two, three, four, five, six, seven, eight, nine, ten, hundred, million}). All numbers and colors appear in \gatitos{}. Numbers are included as a weak control, since the model will tend to make fewer mistakes on them, though such UNMT-style mistakes do occur.

As expected, the \gatitos{}-augmented model performs better on these tokens. Two things are worth noting. First, the model improves noticeably on the complementary distribution---words that do not appear in the lexicon training data---but unsurprisingly improves more on the words that are present in \gatitos{}. Second, the improvements are not as large as expected: why is it not now getting 100\% accuracy? Digging into the outputs, it seems that this is mainly due to the high baseline of (a) undertranslation; (b) hallucination; and (c) copying, as we expect from a model trained without various other tricks like back-translation (see Section \ref{sec:common_errors} for an analysis of common error types). This point is underscored by the models' imperfect performance on the ``easy'' class of numbers.

\inserthitrate

\begin{figure}
    \centering
    \includegraphics[width=1\textwidth]{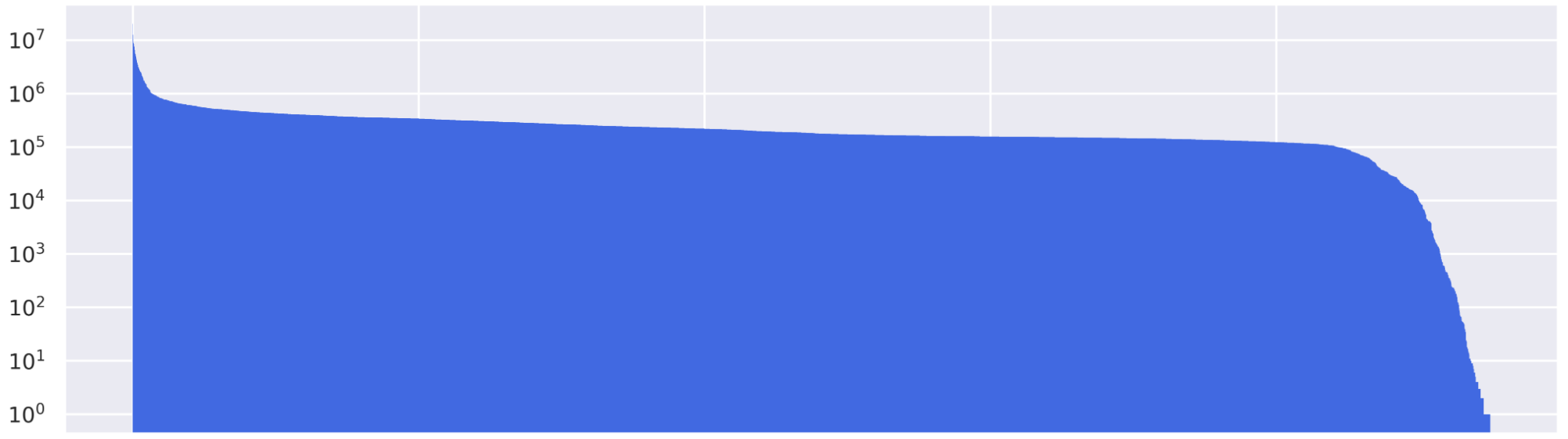}
    \caption{Number of word pairs available in Panlex for each of $4750$ BCP-47 languages (log scale).}
    \label{fig:panlex_langs}
\end{figure}

\section{Biggest winners}

We also look at the top $5$ languages that were the biggest gainers over the baseline for each model. In some cases these may represent remarkable successes of a particular approach---though in other cases they may represent noisy outliers, as is to be expected when evaluating 200 languages.

\subsection{\flores{}}
\paragraph{\enxx{}}
The biggest winners for each model in the \enxx{} direction for the \flores{} evaluation set are given in Table \ref{tab:floresenxxwinners}.

There are seven languages that gained at least $5$ \chrf{} over the baseline on at least one model trained with data augmentation. These languages are:

\begin{enumerate}[nosep]
    \item Bhojpuri (bho): up to $+14.5$ \chrf{}
    \item Ilocano (ilo): up to $+9.1$ \chrf{}
    \item Serbian (sr): up to $+8.3$ \chrf{}
    \item Bambara (bm): up to $+8.1$ \chrf{}
    \item Tibetan (bo): up to $+8.0$ \chrf{}
    \item Nuer (nus): up to $+6.8$ \chrf{}
    \item Mizo (lus): up to $+6.2$ \chrf{}
\end{enumerate}
    
Unsurprisingly, most of these languages are unsupervised or low-resource, except for Serbian which is medium-resource in our dataset. Of the seven languages listed above, we use Panlex data for Ilocano, Serbian, Bambara, Tibetan, Nuer, and Mizo, and there is \gatitos{} data for Bhojpuri, Ilocano, Bambara, and Mizo. As will be discussed in Section \ref{subsec:token_pair_amount}, the \gatitos{} bilingual lexica are clearly a very useful resource for MT, although evidently Panlex alone can help as well. Another interesting finding is that Nuer, which has \textit{no} English-aligned entries in Panlex but $\approx 20$K non-English-aligned entries, still sees large improvements when translating from English. This is evidence that lexicon data can improve performance even in the zero-shot case, where e.g. the model learns better vocabulary alignment between English and Nuer despite not receiving explicit alignment information during training. In Section \ref{subsec:token_pair_amount}, we look at the relationship between the number of lexical data points for a language and the \chrf{} improvement, which provides some insight (albeit not perfect clarity) into why these particular languages did well.

\paragraph{\xxen{}}

Table \ref{tab:floresxxenwinners} shows the top $5$ biggest winners per model for the \xxen{} direction. Clearly there is a lot of overlap with the \enxx{} direction, although there are some differences. Also, note again that the magnitude of improvement in this direction is smaller, likely because the baseline performance is higher and there is less improvement to be made simply by better aligning vocabulary cross-linguistically. Some of the biggest winners in this direction that weren't already discussed for the \enxx{} direction are:

\begin{enumerate}[nosep]
    \item Tsonga (ts): up to $+3.1$ \chrf{}
    \item Guarani (gn): up to $+2.8$ \chrf{}
    \item Bashkir (ba): up to $+2.5$ \chrf{}
    \item Minangkabau (min): up to $+2.5$ \chrf{}
\end{enumerate}

\insertfloresenxxwinners
\insertfloresxxenwinners

\subsection{\ntlevalset{}}

\paragraph{\enxx{}}
The biggest winners on \ntlevalset{} in the \enxx{} direction are given in Table \ref{tab:ntlenxxwinners}. Though there is some overlap with the biggest winners on the \flores{} dataset (e.g. Ilocano, Bambara, Mizo, Bhojpuri), a number of different languages perform well too, some of which simply aren't included in the \flores{} set. The languages which gain $> 5.0$ \chrf{} on this part of the evaluation are:

\begin{enumerate}[nosep]
    \item Adyghe (ady): up to $+14.1$ \chrf{}
    \item Kedah Malay (meo): up to $+12.6$ \chrf{}
    \item Goan Konkani (gom): up to $+11.6$ \chrf{}
    \item Bhojpuri (bho): up to $+10.4$ \chrf{}
    \item Ilocano (ilo): up to $+9.8$ \chrf{}
    \item Avar (av): up to $+9.5$ \chrf{}
    \item Bambara (bm): up to $+9.0$ \chrf{}
    \item Mizo (lus): up to $+8.9$ \chrf{}
    \item Madurese (mad): up to $+6.6$ \chrf{}
    \item Assamese (as): up to $+6.3$ \chrf{}
    \item Pattani Malay (mfa): up to $+5.7$ \chrf{}
    \item Kalaallisut (kl): up to $+5.4$ \chrf{}
    \item Tibetan (bo): up to $+5.2$ \chrf{}
\end{enumerate}

\paragraph{\xxen{}}
The biggest winners in the \xxen{} direction are given in Table \ref{tab:ntlxxenwinners}. Many of the biggest winners overlap with the \enxx{} direction, but some of the languages that haven't yet been mentioned are:

\begin{enumerate}[nosep]
    \item Manipuri (mni\{-Mtei\}): up to $+7.7$ \chrf{}
    \item Dogri (doi): up to $+5.4$ \chrf{}
    \item Dhivehi (dv): up to $+3.4$ \chrf{}
    \item Tigrinya (ti): up to $+3.0$ \chrf{}
\end{enumerate}

\insertntlenxxwinners
\insertntlxxenwinners

\subsection{Big model winners}

The biggest winners for the $1.6$B parameter models are given in Tables \ref{tab:floresenxxwinnersBIG}, \ref{tab:floresxxenwinnersBIG}, \ref{tab:ntlenxxwinnersBIG}, and \ref{tab:ntlxxenwinnersBIG}.

\insertfloresenxxBIGwinners
\insertfloresxxenBIGwinners
\insertntlenxxBIGwinners
\insertntlxxenBIGwinners

\subsection{Empirical study of the quality of Panlex}

One way to judge the quality of a dataset is to review it manually, as in \citep{kreutzer-etal-2022-quality}; another is to see the empirical effects on model quality of training on it. As a byproduct of using Panlex for this project, can we judge the quality of Panlex for different languages?

To reduce noise, we average the scores on the three main uses of the bilexes,  namely the TokenPairs model, the Glowup model, and the Codeswitch model. We average the \flores{} and \ntlevalset{} scores.  We then compare those scores to the baseline model for both \enxx{} and \xxen{}. For the purposes of this analysis, we treat any absolute delta of under 0.3 \chrf{} to be noise.  The result is displayed in Tables \ref{tab:panlexqsmall} and \ref{tab:panlexqbig}.

One would like to say that the upper left-hand corner represents languages with unequivocally high-quality lexical data, and the lower right-hand corner represents languages with poor quality lexical data. Alas, however, this picture is rather muddied when we scale up to larger models, as we see that many languages jump from one bucket to another. Nonetheless, we do see the trend that \gatitos{} languages tend to cluster to the upper left-hand corner in both cases, and that Shan (`shn`) and Latin (`la`) do poorly in all cases, and should likely be avoided by practitioners.

\insertpanlexq

\textbf{Teasing out the confound of the mixed \gatitos{} and Panlex data:} For the 26 \gatitos{} languages, it is harder to trust the previous analysis. However, we can compare the scores of these languages between the GatiPanlexTokenPairs model and the GatiTokenPairs model. The second of these models is trained on a strict subset of the data that the first is. If a language performs better with this subset of the data, we can presume that the Panlex data was on average lower quality; if a language performs better on the superset, the Panlex data might still be lower quality, but its quantity at least makes up for performance to some degree. The languages that do over +0.3 \chrf{} better on the subset data are \texttt{ts, dv, bm, lus, ff, and ckb}, suggesting that those may have poorer-quality Panlex data, with the largest difference being \texttt{lus} at +2.7 \chrf{}; those that do better on the superset are \texttt{gom, mni-Mtei, kri, ln, doi, ay, sa, ti, mai}, and \texttt{as}, suggesting that Panlex still adds useful signal there.

The picture that begins to come together is that Panlex often has some useful signal, but also contains considerable amounts of noise. For a less expressive model that is already not able to reach very high quality, some noise in the lexicons does not hurt, and Panlex can help the model get off the ground for the lowest-resource languages. But for a stronger baseline model that produces higher-quality translations on average, this noise can actively harm performance. Therefore, more carefully curated bilingual lexica, like \gatitos{}, will tend to will yield higher quality results when used for model training with bigger models, as evinced in Table \ref{tab:panlexqbig}.

\section{Does lexical augmentation fix common MT mistakes?}\label{sec:common_errors}

In evaluating the ``big'' models with $1.6$B parameters, we wished to see whether our preferred lexical data augmentation methods (GatiPanlexTokenPairs or CodeswitchMonoGatiPanlex) reduced several common types of MT errors. The errors we looked at were (1) null output, or the ``question mark phenomenon,'' where the model simply outputs some unrelated symbol (such as question marks) instead of actual text; (2) copying, where the model copies some or all of the source sentence in its prediction; and (3) repetition, where the model erroneously repeats the same word or phrase many times. There are other error types we could look at, like hallucination, but we stick with these three basic types for this paper. More precise definitions of these errors are given below.

The results of this analysis are given in Table \ref{tab:commonerrors}. For each error type, we computed the percentage of sentences that were affected by dividing the number of affected sentences by the total number of sentences in the eval set. \flores{} has $806248$ sentence pairs across all languages and \ntlevalset{} has $309887$.

\insertcommonerrorstable

% Alex: results from common mistakes analysis
\paragraph{Null output (question mark phenomenon)}
The first error type occurs when the model outputs only ``??', or some other arbitary character, as its prediction. Instances of this likely indicate catastrophic effects of out-of-domain phenomena for 0-shot translation. 

\paragraph{Copying}
Another common error is copying, where the model's prediction is close or identical to the source sentence. In measuring this phenomenon, we said that any prediction with $> 85\%$ character-level similarity to the source sentence was considered a copy. To measure character-level similarity, we took the multiset intersection of the character frequencies in the source and the character frequencies in the prediction, and then divided the size of the intersection by the number of characters in the source.

\paragraph{Repetition}
The last common error type we examined was repetition. To count these mistakes, we divided the total number of tokens in a sentence by the number of \textit{unique} tokens. If the ratio was $> 3$, we counted the prediction as an instance of erroneous repetition.

% \subsection{Results on common errors}

% The \flores{} and \ntlevalset{} eval sets agreed in this analysis: CodeswitchMonoGatiPanlexBig was best at reducing question mark errors ($-0.3\%$ for \flores{} and $-0.2\%$ for \ntlevalset{}), while GatiPanlexTokenPairsBig was best at reducing copying ($-0.9\%$ on \flores{}, $-0.4\%$ on \ntlevalset{}) and repetition ($-0.7\%$ on \flores{}, $-0.9\%$ on \ntlevalset{}).

% It isn't totally clear why the results are this way, but it could be that improving cross-lingual vocabulary coverage simply reduces catastrophic confusion in the model, although this doesn't explain why the raw token pair model is best at mitigating copying and repetition while CodeswitchMonoGatiPanlexBig does best in preventing question marks. Answering these questions is left for future work.

\subsection{Comparing sampling strategies for translating tokens} \label{token-sampling}

As one recalls from Section \ref{mca}, the Codeswitch augmentation works as follows: Let $D$ represent a multilingual lexicon containing word or phrase translation pairs for many languages. Given a source sentence $x = (x_{1}, x_{2}, . . ., x_{n})$ from monolingual corpus $X_{mono}$, we substitute each token in $x$ for its dictionary translation with probability $p_{tr}$. 

However, there is an issue with this formulation. Because the lexica we use do not have exhaustive coverage across languages, it is often the case that simply looping over $x$ and attempting to translate each token with probability $p_{tr}$ would result in translating a fraction of $x$ that is significantly less than $p_{tr}$. So in order to approximate this desired fraction $p_{tr}$ as closely as possible, we first count how many tokens in $x$ have dictionary translations. Let this number be $k$. We then compute the adjusted probability $\tilde p_{tr} = \max(\frac{np_{tr}}{k}, 1)$,  and sample from amongst the words in $x$ with translations with probability $\tilde p_{tr}$, to obtain the codeswitched sentence $x'$. When substituting a source word for its translation, we choose a translation uniformly at random from all available translations in all languages. Because of this, it is usually the case that $x$ is codeswitched into many languages. Finally, we train the model to reconstruct the monolingual sentence $x$ from $x'$ using the same sequence-to-sequence model and loss function as for the MT task.

In our experiments we use use $p_{tr} = 0.4$. We apply MCA to all $208$ languages in our corpus, but augment only half the available monolingual data and train the remaining half with MASS \citep{song-etal-2019-mass}, as done in the baseline training regime \citep{bapna-etal-2022-building}. We prepend a task token, \texttt{<2codeswitch>}, to the codeswitched sentences to cue the model to perform the MCA task, as well as language (\texttt{<2lang>}) and script (\texttt{<2script>}) tokens. The \texttt{<2lang>} and script \texttt{<2script>} tags are used in all models, including the baseline.

Since this augmentation samples each token with some probability, the number of tokens translated in a given sentence follows a binomial distribution. The Glowup augmentation, however, samples a number of tokens to translate uniformly at random from all possible translatable tokens. So one has a binomial distribution over N tokens sampled, and the other has a uniform distribution---does this make a difference?

To test this we trained a version of the CodeswitchMono model using uniform sampling. The average \chrf{} of the CodeswitchMonoUniform model was 0.1 to 0.2 higher on all four of the (\enxx{} \xxen{}) x (\flores{} \ntlevalset{}) directions. We conclude that this may have a slight benefit, but the difference is within the realm of noise, and does not affect the conclusions elsewhere in this paper.

\begin{figure}[!tbp]
  \centering
  \begin{minipage}[b]{0.5\textwidth}
    \includegraphics[width=\textwidth]{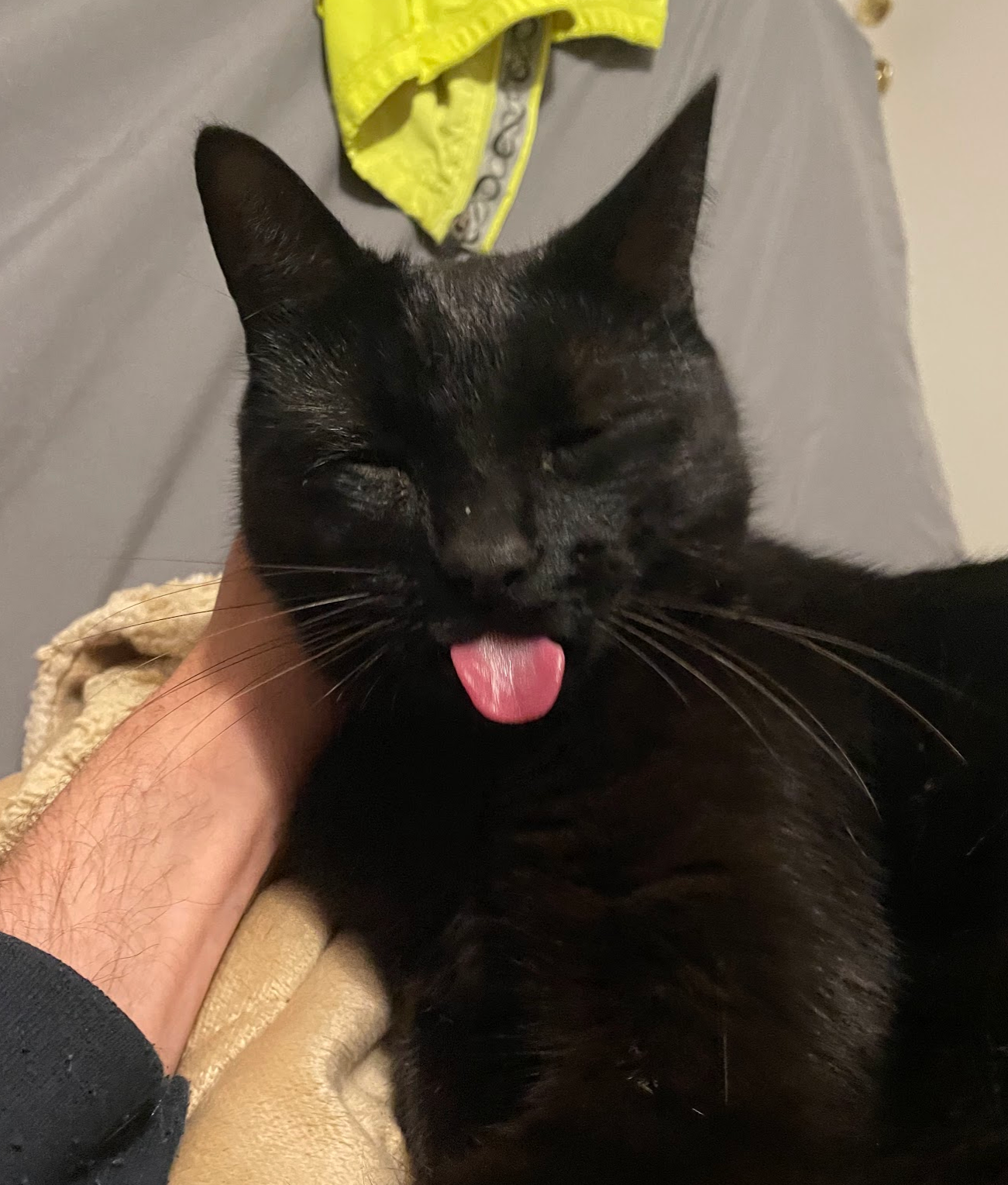}
  \end{minipage}
  \hfill
  \begin{minipage}[b]{0.45\textwidth}
    \includegraphics[width=\textwidth]{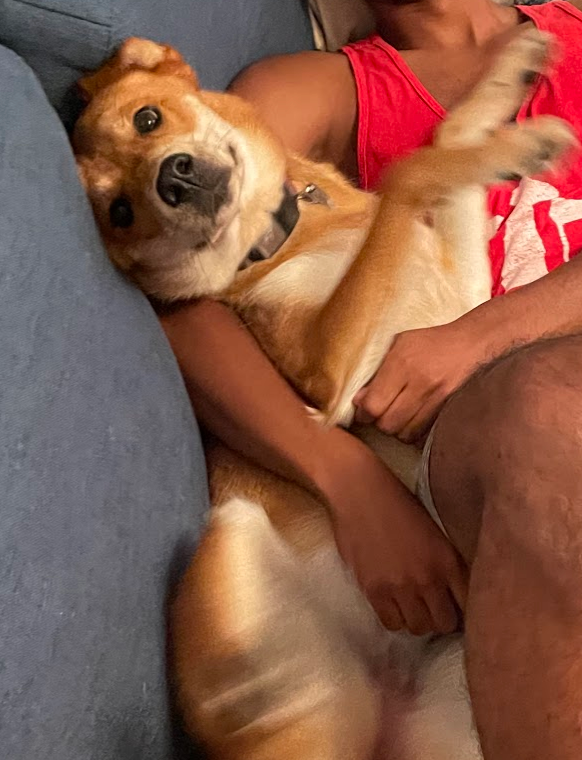}
  \end{minipage}
    \caption{Unsupervised Machine Translation models often confuse words that occur in similar contexts. A frequent example of this mistake is to translate one animal into another animal from a similar semantic category. For example, UNMT models frequently mis-translate ``cat'' (e.g. Odis on the left) as ``dog'', (e.g. Sandy on the right). We hope that multilingual lexica will help fix this sort of mistake, which is easy for a human to correct with a dictionary.}
\end{figure}

\section{Relationship between number of tokens and MT performance}

\begin{figure}[t]
    \centering
    \includegraphics[width=\textwidth]{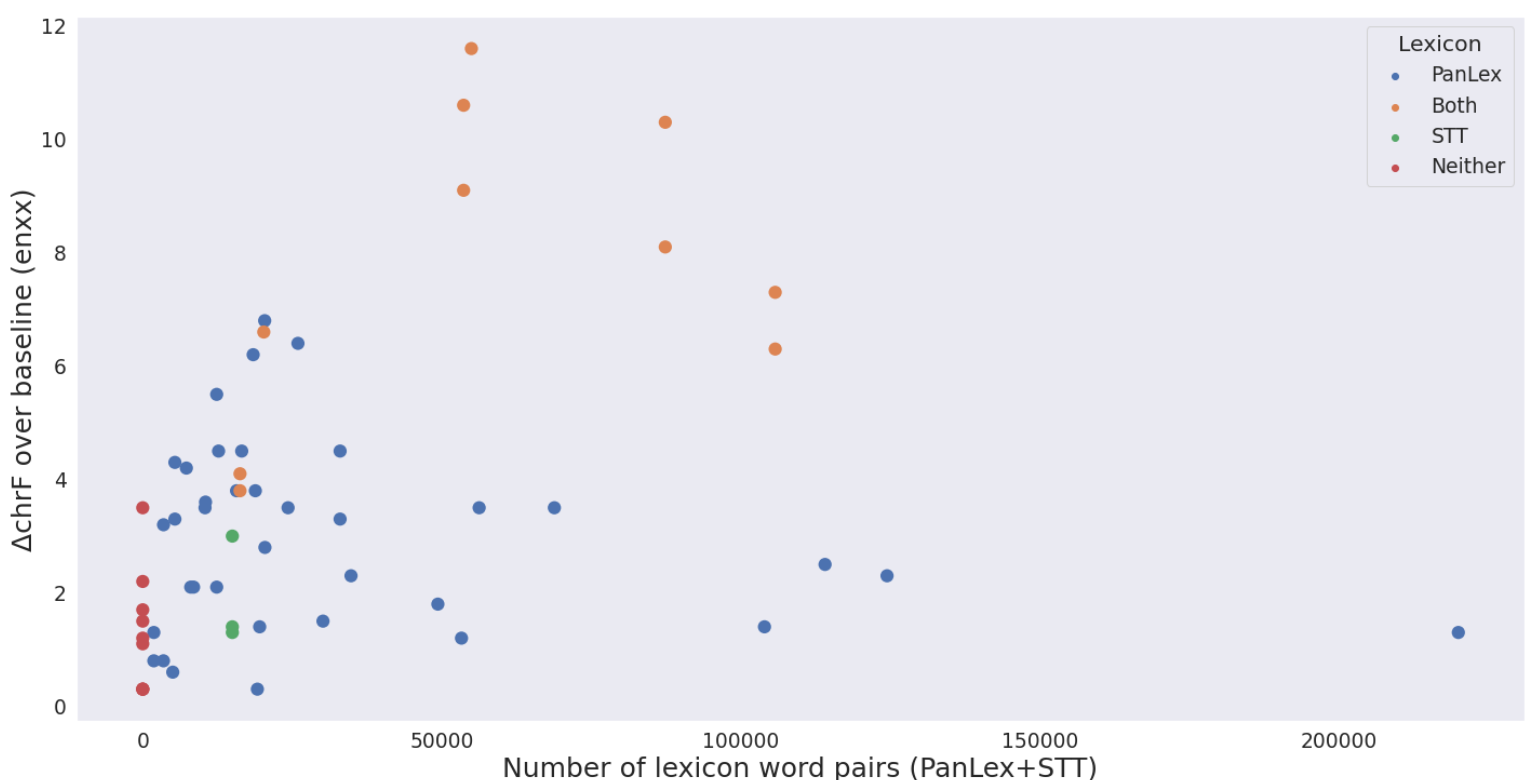}
    \caption{Number of lexicon word pairs in augmented data vs. $\Delta \chrf{}$ over baseline for unsupervised languages in the \enxx{} direction. Results for \flores{} and \ntlevalset{} are combined here.}
    \label{fig:token_pairs_enxx}
\end{figure}

\begin{figure}[t]
    \centering
    \includegraphics[width=\textwidth]{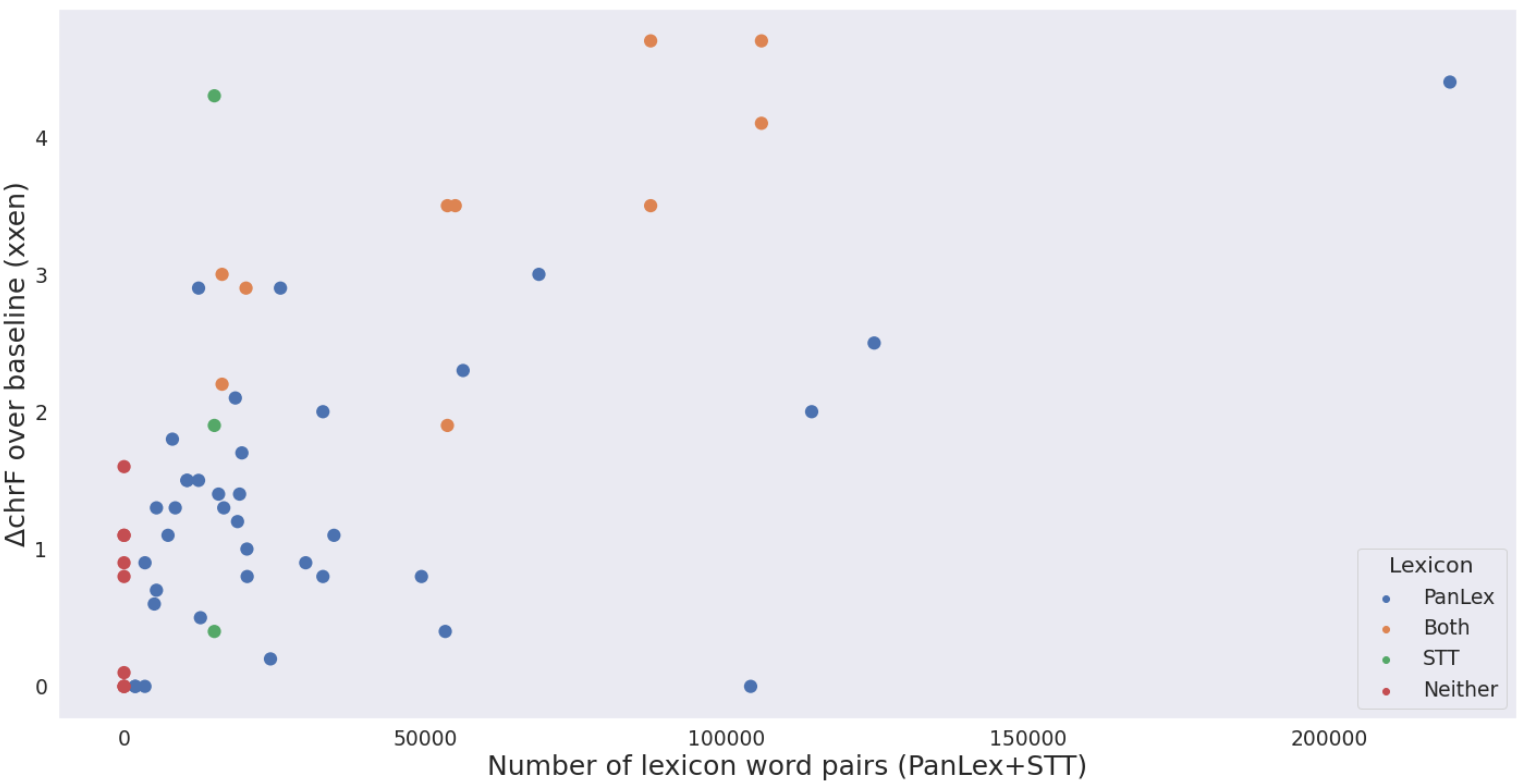}
    \caption{Number of lexicon word pairs in augmented data vs. $\Delta \chrf{}$ over baseline for unsupervised languages in the \xxen{} direction. Results for \flores{} and \ntlevalset{} are combined here.}
    \label{fig:token_pairs_xxen}
\end{figure}

We also graph the relationship between number of lexical word pairs and $\Delta \chrf{}$ in Figures \ref{fig:token_pairs_enxx} (\enxx{}) and \ref{fig:token_pairs_xxen} (\xxen{}) for URLs only. The results for \flores{} and \ntlevalset{} are combined in these plots. In both directions, we observe a moderate positive relationship between the number of lexical word pairs for a given language in the augmented data and the $\Delta \chrf{}$ over the baseline.

\section{Languages}

\subsection{Rationale for Language Choice}
\label{appendix:language-rationale}

Although this project is aligned with the 1000-language initiative from \citet{bapna-etal-2022-building}, we wanted to use smaller models for more rapid iteration, and as a result, commensurately smaller data and number of languages to fit comfortably in the model. Therefore, we chose to work with about 200 languages.

With this in mind, we also wanted to choose specifically those languages whose performance we could measure. Therefore, our approach was as follows:

\begin{itemize}[nosep]
    \item Include all languages with supervised (parallel) data, for maximal cross-lingual transfer
    \item Include all languages that have non-zero data and a \flores{} eval set
    \item Include all languages that have non-zero data and a \ntlevalset{} eval set
\end{itemize}

\insertmodeldescriptionsfull

\subsection{Complete Language data}

The following table gives a list of the languages used in our experiments, along with some linguistic and resource-related statistics. The numbers for data resources (i.e. Mono, Parallel, Panlex, and \gatitos{}) refer to the amount of data actually used in our experiments, \textit{not} necessarily the total amount of data available. For example, we subsampled the parallel and high-resource monolingual data we had available by a factor of $10$.

\insertlanguageslist

\section{Full results}
The full results on \flores{} for the various models we trained are available in Tables \ref{tab:fullresults-flores-enxx} and \ref{tab:fullresults-flores-xxen}. Model abbreviations are clarified in Table \ref{tab:modeldescriptions}. Scores from the NLLB model are included as reference, though keep in mind that the smaller research models in this paper will naturally have lower quality; even the ``Big'' models are 30x smaller, and not optimized with back-translation and so on.

\insertfullenxx
\insertfullxxen
\end{document}

%% file: tables.tex
\newcommand{\insertfloresenxxresults}{

\begin{table}[t]
\renewcommand\arraystretch{1.4}
    \centering
    \begin{tabularx}{\textwidth}{lXXXXX|XX}
    \toprule
         \textbf{Model} & $\mu$ & HRL & MRL & LRL & URL & LRL$_{\scaleto{GAT}{3pt}}$ & URL$_{\scaleto{GAT}{3pt}}$ \\\hline
        Baseline & 39.7 & 50.5 & 46.6 & 34.4 & 28.7 &  35.4 & 26.6 \\\hline
        GatiPanlexTokenPairs & \cellcolor{scale6} +0.4 & \cellcolor{scale4} -0.3 & \cellcolor{scale4} -0.1 & \cellcolor{scale6} +0.5 & \cellcolor{scale7} +1.4 &  \cellcolor{scale7} +1.8 & \cellcolor{scale9} +5.1 \\
CodeswitchMono & \cellcolor{scale6} +0.8 & \cellcolor{scale6} +0.8 & \cellcolor{scale6} +0.9 & \cellcolor{scale6} +0.8 & \cellcolor{scale6} +0.4 & \cellcolor{scale7} +1.9 & \cellcolor{scale9} +4.8 \\
CodeswitchParallel & \cellcolor{scale2} -1.4 & \cellcolor{scale2} -1.9 & \cellcolor{scale2} -1.7 & \cellcolor{scale2} -1.2 & \cellcolor{scale3} -0.9 &  \cellcolor{scale5} +0.2 & \cellcolor{scale8} +2.1 \\
CodeswitchMonoParallel & \cellcolor{scale4} -0.2 & \cellcolor{scale2} -1.0 & \cellcolor{scale3} -0.7 & \cellcolor{scale5} +0.0 & \cellcolor{scale7} +1.1 &  \cellcolor{scale7} +1.5 & \cellcolor{scale9} +5.2 \\
CodeswitchMonoGatiPanlex & \cellcolor{scale7} +1.0 & \cellcolor{scale6} +0.4 & \cellcolor{scale6} +0.6 & \cellcolor{scale7} +1.3 & \cellcolor{scale7} \textbf{+1.5} & \cellcolor{scale8} \textbf{+2.8} & \cellcolor{scale9} \textbf{+7.0} \\
GlowupMono & \cellcolor{scale7} +1.1 & \cellcolor{scale7} \textbf{+1.5} & \cellcolor{scale7} \textbf{+1.4} & \cellcolor{scale7} +1.0 & \cellcolor{scale6} +0.5 &  \cellcolor{scale8} +2.2 & \cellcolor{scale9} +5.6 \\
GlowupParallel & \cellcolor{scale2} -1.1 & \cellcolor{scale2} -2.0 & \cellcolor{scale2} -1.8 & \cellcolor{scale2} -1.1 & \cellcolor{scale6} +0.4 &  \cellcolor{scale6} +0.8 & \cellcolor{scale9} +3.0 \\
GlowupMonoParallel & \cellcolor{scale5} +0.3 & \cellcolor{scale5} +0.1 & \cellcolor{scale5} +0.2 & \cellcolor{scale5} +0.1 & \cellcolor{scale7} +1.0 &  \cellcolor{scale7} +1.3 & \cellcolor{scale9} +3.4 \\
GlowupMonoGatiPanlex & \cellcolor{scale7} \textbf{+1.2} & \cellcolor{scale7} +1.3 & \cellcolor{scale7} +1.2 & \cellcolor{scale7} \textbf{+1.4} & \cellcolor{scale6} +0.8 & \cellcolor{scale8} +2.5 & \cellcolor{scale9} +5.6 \\
    \bottomrule
    \end{tabularx}
    \caption{\enxx{} performance on the \flores{} test set, measured in $\Delta\chrf{}$ over the baseline.}
    \label{tab:floresenxx}
\end{table}
}

\newcommand{\insertfloresxxenresults}{
\begin{table}[]
\renewcommand\arraystretch{1.4}
    \centering
    \begin{tabular}{llllll|ll}
    \toprule
        \textbf{Model} & $\mu$ & HRL & MRL & LRL & URL &  LRL$_{\scaleto{GAT}{3pt}}$ & URL$_{\scaleto{GAT}{3pt}}$ \\\hline
        Baseline & 47.2 & 57.2 & 52.1 & 43.6 & 37.0  & 49.1 & 36.1 \\\hline
        GatiPanlexTokenPairs & \cellcolor{scale5} -0.0 & \cellcolor{scale4} -0.3 & \cellcolor{scale4} -0.3 & \cellcolor{scale5} +0.1 & \cellcolor{scale6} \textbf{+0.5} &  \cellcolor{scale7} +1.0 & \cellcolor{scale8} +2.4 \\
CodeswitchMono & \cellcolor{scale5} +0.2 & \cellcolor{scale6} +0.5 & \cellcolor{scale6} +0.4 & \cellcolor{scale5} +0.1 & \cellcolor{scale4} -0.1 & \cellcolor{scale6} +0.8 & \cellcolor{scale7} +1.5 \\
CodeswitchParallel & \cellcolor{scale2} -1.2 & \cellcolor{scale2} -1.6 & \cellcolor{scale2} -1.5 & \cellcolor{scale2} -1.2 & \cellcolor{scale4} -0.3 &  \cellcolor{scale5} +0.0 & \cellcolor{scale7} +1.2 \\
CodeswitchMonoParallel & \cellcolor{scale3} -0.8 & \cellcolor{scale3} -0.7 & \cellcolor{scale3} -0.9 & \cellcolor{scale3} -0.8 & \cellcolor{scale3} -0.7 &  \cellcolor{scale4} -0.1 & \cellcolor{scale7} +1.1 \\
CodeswitchMonoGatiPanlex & \cellcolor{scale5} +0.1 & \cellcolor{scale5} +0.3 & \cellcolor{scale5} +0.1 & \cellcolor{scale5} +0.3 & \cellcolor{scale4} -0.1 & \cellcolor{scale6} +0.9 & \cellcolor{scale8} \textbf{+2.9} \\
GlowupMono & \cellcolor{scale6} \textbf{+0.7} & \cellcolor{scale7} \textbf{+1.2} & \cellcolor{scale7} \textbf{+1.1} & \cellcolor{scale6} \textbf{+0.6} & \cellcolor{scale5} +0.0  & \cellcolor{scale7} \textbf{+1.5} & \cellcolor{scale8} +2.0 \\
GlowupParallel & \cellcolor{scale2} -1.3 & \cellcolor{scale2} -1.7 & \cellcolor{scale2} -1.6 & \cellcolor{scale2} -1.2 & \cellcolor{scale3} -0.6 & \cellcolor{scale4} -0.1 & \cellcolor{scale7} +1.2 \\
GlowupMonoParallel & \cellcolor{scale5} +0.1 & \cellcolor{scale5} +0.0 & \cellcolor{scale5} -0.0 & \cellcolor{scale5} -0.0 & \cellcolor{scale6} \textbf{+0.5} &  \cellcolor{scale6} +0.6 & \cellcolor{scale7} +1.7 \\
GlowupMonoGatiPanlex & \cellcolor{scale6} +0.6 & \cellcolor{scale7} +1.0 & \cellcolor{scale6} +0.8 & \cellcolor{scale6} \textbf{+0.6} & \cellcolor{scale4} -0.1 & \cellcolor{scale7} +1.3 & \cellcolor{scale7} +1.8 \\
    \bottomrule
    \end{tabular}
    \caption{\xxen{} performance on the FLORES-200 test set, measured in $\Delta\chrf{}$ over the baseline.}
    \label{tab:floresxxen}
\end{table}
}

\newcommand{\insertfloresenxxwinners}{
\begin{table}[]
\resizebox{\textwidth}{!}{%
\renewcommand\arraystretch{1.4}
    \centering
    \begin{tabular}{llllll}
    \toprule
         \textbf{Model} &&&&&  \\\hline
         GatiPanlexTokenPairs & ilo (+7.5) & nus (+6.8) & lus (+6.2) & bm (+5.5) & ts (+4.4) \\
         CodeswitchMono & bho (+8.6) & bo (+5.6) & lus (+5.2) & ilo (+5.0) & quy (+4.9) \\
         CodeswitchParallel & ilo (+3.7) & bho (+3.5) & as (+2.4) & lus (+2.3) & min (+1.9) \\
         CodeswitchMonoParallel & bho (+9.4) & ilo (+6.4) & bo (+6.2) & lus (+6.0) & ln (+4.9) \\
         CodeswitchMonoGatiPanlex & ilo (+9.1) & sr (+8.3) & bm (+8.1) & bho (+7.9) & bo (+6.8) \\
         GlowupMono & sr (+6.2) & acq (+4.5) & bo (+3.8) & aeb (+3.1) & bho (+3.1) \\
         GlowupParallel & bho (+4.4) & shn (+3.8) & sg (+3.7) & kac (+3.6) & kpb (+3.5) \\
         GlowupMonoParallel & bho (+12.3) & sr (+7.1) & sg (+4.5) & bo (+4.1) & nus (+3.3) \\
         GlowupMonoGatiPanlex & bho (+14.5) & bo (+8.0) & sr (+7.3) & bm (+5.7) & acq (4.2) \\
    \bottomrule
    \end{tabular}}
    \caption{Top $5$ biggest winners per model (en$\rightarrow$xx) on the FLORES-200 test set, measured in $\Delta\chrf{}$ over the baseline.}
    \label{tab:floresenxxwinners}
\end{table}
}

\newcommand{\insertfloresxxenwinners}{
\begin{table}[]
\resizebox{\textwidth}{!}{%
\renewcommand\arraystretch{1.4}
    \centering
    \begin{tabular}{llllll}
    \toprule
         \textbf{Model} &&&&& \\\hline
        GatiPanlexTokenPairs & bm (+3.5) & ts (+3.1) & lus (+2.7) & gn (+2.6) & tum (+2.1) \\ 
        CodeswitchMono & bo (+2.5) & lus (+2.4) & ti (+1.4) & ko (+1.3) & ay (+1.3) \\ 
        CodeswitchParallel & mni (+2.2) & min (+1.9) & kg (+1.7) & lus (+1.4) & kbp (+1.4) \\ 
        CodeswitchMonoParallel & bo (+2.2) & lus (+1.9) & ay (+1.4) & bm (+1.1) & ff (+1.0) \\ 
        CodeswitchMonoGatiPanlex & lus (+5.0) & bo (+3.5) & bm (+3.2) & gn (+2.8) & ba (+2.5) \\ 
        GlowupMono & bo (+2.4) & ti (+2.2) & ks (+2.2) & am (+2.0) & mai (+2.0) \\ 
        GlowupParallel & min (+2.1) & ee (+1.9) & kg (+1.6) & kac (+1.5) & kbp (+1.5) \\ 
        GlowupMonoParallel & lus (+2.9) & min (+2.5) & bo (+1.8) & ace (+1.8) & bug (+1.8) \\ 
        GlowupMonoGatiPanlex & lus (+2.2) & gn (+2.2) & ko (+1.9) & sa (+1.8) & ckb (+1.7) \\ 
    \bottomrule
    \end{tabular}}
    \caption{Top $5$ biggest winners per model (\xxen{}) on the FLORES-200 test set, measured in $\Delta\chrf{}$ over the baseline.}
    \label{tab:floresxxenwinners}
\end{table}
}

\newcommand{\insertntlenxxresults}{
\begin{table}[]
\renewcommand\arraystretch{1.4}
    \centering
    \begin{tabular}{lllll|ll}
    \toprule
        \textbf{Model} & $\mu$ & MRL & LRL & URL & LRL$_{\scaleto{GAT}{4pt}}$ & URL$_{\scaleto{GAT}{4pt}}$ \\\hline 
        Baseline & 21.4 & 28.7 & 22.9 & 19.1 & 17.1 & 20.0 \\\hline
GatiPanlexTokenPairs & \cellcolor{scale7} +1.7 & \cellcolor{scale6} +0.6 & \cellcolor{scale7} +1.6 & \cellcolor{scale7} +1.8 & \cellcolor{scale9} +3.6 & \cellcolor{scale9} +8.4 \\
CodeswitchMono & \cellcolor{scale7} +1.4 & \cellcolor{scale7} \textbf{+1.6} & \cellcolor{scale6} +0.8 & \cellcolor{scale7} +1.9 & \cellcolor{scale9} +3.2 & \cellcolor{scale9} +8.3 \\
CodeswitchParallel & \cellcolor{scale3} -0.5 & \cellcolor{scale2} -1.8 & \cellcolor{scale3} -0.9 & \cellcolor{scale5} -0.0 & \cellcolor{scale6} +0.9 & \cellcolor{scale9} +3.4 \\
CodeswitchMonoParallel & \cellcolor{scale7} +1.4 & \cellcolor{scale4} -0.1 & \cellcolor{scale6} +0.7 & \cellcolor{scale8} +2.2 & \cellcolor{scale8} +2.5 & \cellcolor{scale9} +6.1 \\
CodeswitchMonoGatiPanlex & \cellcolor{scale8} \textbf{+2.2} & \cellcolor{scale7} +1.1 & \cellcolor{scale8} \textbf{+2.1} & \cellcolor{scale8} \textbf{+2.3} & \cellcolor{scale9} \textbf{+3.9} & \cellcolor{scale9} +8.0 \\
GlowupMono & \cellcolor{scale5} +0.3 & \cellcolor{scale7} +1.0 & \cellcolor{scale6} +0.6 & \cellcolor{scale4} -0.2 & \cellcolor{scale8} +2.0 & \cellcolor{scale9} +7.6 \\
GlowupParallel & \cellcolor{scale4} -0.2 & \cellcolor{scale2} -1.6 & \cellcolor{scale3} -0.7 & \cellcolor{scale6} +0.6 & \cellcolor{scale7} +1.0 & \cellcolor{scale9} +3.0 \\
GlowupMonoParallel & \cellcolor{scale5} +0.2 & \cellcolor{scale4} -0.3 & \cellcolor{scale3} -0.7 & \cellcolor{scale7} +1.1 & \cellcolor{scale8} +2.1 & \cellcolor{scale9} +7.6 \\
GlowupMonoGatiPanlex & \cellcolor{scale7} +1.3 & \cellcolor{scale7} +1.4 & \cellcolor{scale7} +1.9 & \cellcolor{scale6} +0.7 & \cellcolor{scale9} +3.5 & \cellcolor{scale9} \textbf{+9.5} \\
    \bottomrule
    \end{tabular}
    \caption{\enxx{} performance on the \ntlevalset{} test set, measured in $\Delta\chrf{}$ over the baseline.}
    \label{tab:ntlenxx}
\end{table}
}

\newcommand{\insertntlxxenresults}{
\begin{table}[]
\renewcommand\arraystretch{1.4}
    \centering
    \begin{tabular}{lllll|ll}
    \toprule
        \textbf{Model} & $\mu$ & MRL & LRL & URL & LRL$_{\scaleto{GAT}{4pt}}$ & URL$_{\scaleto{GAT}{4pt}}$ \\\hline
        Baseline & 29.4 & 43.2 & 30.2 & 27.0 & 24.7 & 22.1 \\\hline 
GatiPanlexTokenPairs & \cellcolor{scale6} \textbf{+0.8} & \cellcolor{scale5} +0.0 & \cellcolor{scale6} \textbf{+0.5} & \cellcolor{scale7} \textbf{+1.1} & \cellcolor{scale7} \textbf{+1.1} & \cellcolor{scale9} +3.2 \\
CodeswitchMono & \cellcolor{scale5} +0.1 & \cellcolor{scale5} +0.1 & \cellcolor{scale5} +0.0 & \cellcolor{scale5} +0.2 & \cellcolor{scale6} +0.8 & \cellcolor{scale8} +2.2 \\
CodeswitchParallel & \cellcolor{scale3} -0.4 & \cellcolor{scale2} -1.2 & \cellcolor{scale3} -0.9 & \cellcolor{scale5} +0.2 & \cellcolor{scale5} +0.3 & \cellcolor{scale7} +1.8 \\
CodeswitchMonoParallel & \cellcolor{scale3} -0.5 & \cellcolor{scale3} -0.8 & \cellcolor{scale3} -0.7 & \cellcolor{scale4} -0.2 & \cellcolor{scale5} +0.2 & \cellcolor{scale7} +1.6 \\
CodeswitchMonoGatiPanlex & \cellcolor{scale6} +0.5 & \cellcolor{scale5} +0.0 & \cellcolor{scale6} \textbf{+0.5} & \cellcolor{scale6} +0.6 & \cellcolor{scale7} \textbf{+1.1} & \cellcolor{scale9} \textbf{+3.3} \\
GlowupMono & \cellcolor{scale5} -0.0 & \cellcolor{scale6} \textbf{+0.9} & \cellcolor{scale5} +0.2 & \cellcolor{scale3} -0.4 & \cellcolor{scale6} +0.5 & \cellcolor{scale7} +1.7 \\
GlowupParallel & \cellcolor{scale3} -0.6 & \cellcolor{scale2} -1.5 & \cellcolor{scale3} -0.9 & \cellcolor{scale4} -0.3 & \cellcolor{scale5} +0.0 & \cellcolor{scale7} +1.3 \\
GlowupMonoParallel & \cellcolor{scale5} +0.0 & \cellcolor{scale4} -0.2 & \cellcolor{scale4} -0.2 & \cellcolor{scale5} +0.2 & \cellcolor{scale5} +0.2 & \cellcolor{scale7} +1.6 \\
GlowupMonoGatiPanlex & \cellcolor{scale5} +0.2 & \cellcolor{scale6} +0.8 & \cellcolor{scale5} +0.3 & \cellcolor{scale5} +0.1 & \cellcolor{scale6} +0.8 & \cellcolor{scale9} \textbf{+3.3} \\
    \bottomrule
    \end{tabular}
    \caption{\xxen{} performance on the \ntlevalset{} test set, measured in $\Delta\chrf{}$ over the baseline.}
    \label{tab:ntlxxen}
\end{table}
}

\newcommand{\insertntlenxxwinners}{
\begin{table}[]
\resizebox{\textwidth}{!}{%
\renewcommand\arraystretch{1.4}
    \centering
    \begin{tabular}{llllll}
    \toprule
         \textbf{Model} &&&&&  \\\hline
        GatiPanlexTokenPairs & gom (+11.6) & ilo (+8.1) & meo (+7.7) & bm (+6.8) & lus (+6.6) \\ 
        CodeswitchMono & ady (+14.1) & av (+9.5) & meo (+6.8) & mad (+6.3) & gom (+5.3) \\ 
        CodeswitchParallel & tiv (+3.6) & min (+3.4) & as (+3.3) & iso (+3.2) & mfa (+2.9) \\ 
        CodeswitchMonoParallel & lus (+7.7) & ady (+6.8) & mad (+6.6) & bm (+6.0) & mfa (+5.7) \\ 
        CodeswitchMonoGatiPanlex & bho (+10.4) & gom (+9.9) & ilo (+9.8) & bm (+9.0) & lus (+8.9) \\ 
        GlowupMono & meo (+10.4) & bho (+6.6) & bo (+5.2) & gom (+5.0) & mfa (+4.1) \\ 
        GlowupParallel & bal (+4.1) & yua (+3.1) & meo (+3.0) & tiv (+2.9) & mni (+2.6) \\ 
        GlowupMonoParallel & meo (+7.1) & gom (+5.9) & mad (+3.9) & za (+3.7) & mni (+3.5) \\ 
        GlowupMonoGatiPanlex & meo (+12.6) & gom (+11.6) & as (+6.3) & ilo (+6.2) & kl (+5.4) \\ 
    \bottomrule
    \end{tabular}}
    \caption{Top $5$ biggest winners per model (en$\rightarrow$xx) on the \ntlevalset{} test set, measured in $\Delta\chrf{}$ over the baseline.}
    \label{tab:ntlenxxwinners}
\end{table}
}

\newcommand{\insertntlxxenwinners}{
\begin{table}[]
\resizebox{\textwidth}{!}{%
\renewcommand\arraystretch{1.4}
    \centering
    \begin{tabular}{llllll}
    \toprule
         \textbf{Model} &&&&&  \\\hline
        GatiPanlexTokenPairs & bm (+4.7) & mni-Mtei (+4.7) & gn (+3.8) & lus (+3.6) & ilo (+3.5) \\ 
        CodeswitchMono & av (+3.5) & mni-Mtei (+2.9) & yua (+2.5) & dv (+2.3) & lus (+2.2) \\ 
        CodeswitchParallel & mni (+3.0) & cv (+2.5) & av (+2.1) & lus (+1.9) & ee (+1.8) \\ 
        CodeswitchMonoParallel & mni-Mtei (+7.7) & av (+3.8) & lus (+2.4) & bm (+1.5) & chr (+1.5) \\ 
        CodeswitchMonoGatiPanlex & mni-Mtei (+5.4) & lus (+5.3) & gn (+3.5) & bm (+3.5) & dv (+3.4) \\ 
        GlowupMono & ti (+3.0) & bo (+1.8) & or (+1.6) & quc (+1.4) & dv (+1.4) \\ 
        GlowupParallel & ee (+2.2) & mad (+1.6) & yua (+1.6) & av (+1.6) & bm (+1.5) \\ 
        GlowupMonoParallel & mad (+2.8) & min (+2.2) & lus (+2.0) & quc (+1.6) & gom (+1.3) \\ 
        GlowupMonoGatiPanlex & mni-Mtei (+6.0) & doi (+5.4) & gom (+3.5) & dv (+3.3) & bm (+2.7) \\ 
    \bottomrule
    \end{tabular}}
    \caption{Top $5$ biggest winners per model (\xxen{}) on the \ntlevalset{} test set, measured in $\Delta\chrf{}$ over the baseline.}
    \label{tab:ntlxxenwinners}
\end{table}
}

\newcommand{\insertfloresenxxBIGwinners}{
\begin{table}[]
\resizebox{\textwidth}{!}{%
\renewcommand\arraystretch{1.4}
    \centering
    \begin{tabular}{llllll}
    \toprule
         \textbf{Model} &&&&&  \\\hline
        GatiPanlexTokenPairsBig & ts (+7.5) & din (+6.0) & ln (+ 5.8) & ilo (+5.3) & ay (+4.1) \\
        CodeswitchMonoGatiPanlexBig & ts (+6.9) & bm (+5.6) & ilo (+5.2) & ln (+4.8) & mag (+3.8) \\
    \bottomrule
    \end{tabular}}
    \caption{Top $5$ biggest winners per model (en$\rightarrow$xx) on the FLORES-200 test set for the 1.6B parameter models, measured in $\Delta\chrf{}$ over the baseline.}
    \label{tab:floresenxxwinnersBIG}
\end{table}
}

\newcommand{\insertfloresxxenBIGwinners}{
\begin{table}[]
\resizebox{\textwidth}{!}{%
\renewcommand\arraystretch{1.4}
    \centering
    \begin{tabular}{llllll}
    \toprule
         \textbf{Model} &&&&&  \\\hline
        GatiPanlexTokenPairsBig & tpi (+9.1) & mni (+6.2) & bm (+5.5) & ts (+3.5) & ay (+3.1)  \\
        CodeswitchMonoGatiPanlexBig & tpi (+5.1) & ay (+3.0) & mni (+2.2) & bm (+1.9) & ltg (+1.2) \\
    \bottomrule
    \end{tabular}}
    \caption{Top $5$ biggest winners per model (xx$\rightarrow$en) on the FLORES-200 test set for the 1.6B parameter models, measured in $\Delta\chrf{}$ over the baseline.}
    \label{tab:floresxxenwinnersBIG}
\end{table}
}

\newcommand{\insertntlenxxBIGwinners}{
\begin{table}[]
\resizebox{\textwidth}{!}{%
\renewcommand\arraystretch{1.4}
    \centering
    \begin{tabular}{llllll}
    \toprule
         \textbf{Model} &&&&&  \\\hline
         GatiPanlexTokenPairsBig & ts (+7.7) & gom (+6.1) & ilo (+6.0) & dv (+5.8) & bm (+5.1) \\
         CodeswitchMonoGatiPanlexBig & ts (+7.3) & bm (+7.0) & ilo (+6.8) & mni-Mtei (+5.5) & gom (+5.3) \\
    \bottomrule
    \end{tabular}}
    \caption{Top $5$ biggest winners per model (en$\rightarrow$xx) on the \ntlevalset{} test set for the 1.6B parameter models, measured in $\Delta\chrf{}$ over the baseline.}
    \label{tab:ntlenxxwinnersBIG}
\end{table}
}

\newcommand{\insertntlxxenBIGwinners}{
\begin{table}[]
\resizebox{\textwidth}{!}{%
\renewcommand\arraystretch{1.4}
    \centering
    \begin{tabular}{llllll}
    \toprule
        \textbf{Model} &&&&&  \\\hline
        GatiPanlexTokenPairsBig & cv (+8.7) & mni (+8.4) & bm (+6.6) & kl (+5.8) & ee (+5.0) \\
        CodeswitchMonoGatiPanlexBig & cv (+6.9) & kl (+6.2) & ce (+3.4) & av (+2.8) & chr (+2.7) \\
    \bottomrule
    \end{tabular}}
    \caption{Top $5$ biggest winners per model (xx$\rightarrow$en) on the \ntlevalset{} test set for the 1.6B parameter models, measured in $\Delta\chrf{}$ over the baseline.}
    \label{tab:ntlxxenwinnersBIG}
\end{table}
}

\newcommand{\insertexampletranslations}{
\begin{table}[]
    \centering
    \footnotesize
    \begin{tabularx}{\textwidth}{lX}
    \toprule
         \textbf{Source} & Khawvelah hian \textcolor{forestgreen}{tawng} chi \textcolor{forestgreen}{hrang} 5000 \textcolor{forestgreen}{chuang} a awma, heng zinga \textcolor{forestgreen}{tawng} 20 ai tam hi \textcolor{forestgreen}{chuan tawng} hmangtu \textcolor{forestgreen}{maktaduai} 50 aia tam a nei a ni. \\
         \textbf{Reference} & The world has over 5,000 different \textcolor{forestgreen}{languages}, \textcolor{forestgreen}{including} more than \textcolor{forestgreen}{twenty} with 50 \textcolor{forestgreen}{million} or more speakers. \\
         \textbf{Baseline} & There are 5000 houses in the city, 5000 houses in the city, and 5000 houses in the city. \\
         \textbf{Augment} & There are 5000 \textcolor{forestgreen}{languages} in the world, 20 of which are 50 \textcolor{forestgreen}{million languages}. \\\hline
         \textbf{Source} & Maani isumaqarpunga inuunerma sinnerani pilluartussanngorlunga.  \\
         \textbf{Reference} & Here I think I'll \textcolor{forestgreen}{remain} happy \textcolor{forestgreen}{throughout} my life.  \\
         \textbf{Baseline} & I believe I am the \textcolor{forestgreen}{only} one \textcolor{forestgreen}{who} has ever died. \\
         \textbf{Augment} & I mean, I'm \textcolor{forestgreen}{blessed with} a lifetime of joy. \\\hline
         \textbf{Source} & \foreignlanguage{russian}{ \textcolor{forestgreen}{нагахь} вы ж хьашто \textcolor{forestgreen}{хилар} в дӀадахар къамелан ма расширяемости, дазар с нами хула uservoice }\\
    \textbf{Reference} & If you are interested in continuing the conversation around extensibility, connect with us via uservoice \\
         \textbf{Baseline} & If you want to use the extension, see uservoice \\
         \textbf{Augment} & If you want to \textcolor{forestgreen}{learn} more about extensibility, contact us at uservoice \\
    \bottomrule
    \end{tabularx}
    \caption{Examples of translations where our lexical data augmentation methods appear to have helped the model choose the correct vocabulary, in Mizo (lus), Kalaallisut (kl), and Chechen (ce). Words appearing in the lexical data are colored \textcolor{forestgreen}{green}.}
    \label{tab:exampletranslations}
\end{table}
}

\newcommand{\insertgroundingmistakes}{
\begin{table}[]
    \centering
    \begin{tabularx}{\textwidth}{lX}
    \toprule
         \textbf{Source} & Saanbusaan yeroo baay'ee ayyaanaaf ni qopha'a  \\
         \textbf{Reference} & \textcolor{forestgreen}{Pie} is usually reserved for holidays.  \\
         \textbf{Model output} & \textcolor{red}{Sandwiches} are often served at festive occasions. \\\hline
         \textbf{Source} & m\textipa{E}ti lae nye \textipa{V}ukula la eye \textipa{V}u\textipa{N}utinunyala lae na be wòle z\textipa{O}z\textipa{O}m.  \\
         \textbf{Reference} & the \textcolor{forestgreen}{conductor} is the \textcolor{forestgreen}{driver} and the \textcolor{forestgreen}{engineer} keeps \textcolor{forestgreen}{it} going  \\
         \textbf{Model output} & The \textcolor{red}{rock badger} is the \textcolor{red}{astronaut}, and the \textcolor{red}{nuclear physicist} makes \textcolor{red}{him} the \textcolor{red}{astronaut} \\\hline
         \textbf{Source} & Mep\textipa{E} s\textipa{E} mey\textipa{E} aboa kraman dakoro p\textipa{E}.  \\
         \textbf{Reference} & I would want to be a  \textcolor{forestgreen}{dog} for a day.  \\
         \textbf{Model output} & I would want to be a \textcolor{red}{crocodile} for just one day. \\
    \bottomrule
    \end{tabularx}
    \caption{Examples of the sorts of mistakes we aim to fix in this project. Models often confuse distributionally similar words (such as ``pie'' and ``sandwiches''), and this problem may be mitigated by adding bilingual lexicon entries as grounding. The source languages in these examples are, from top to bottom: Oromo (om), Ewe (ee), and Bambara (bm).}
    \label{tab:groundingmistakes}
\end{table}
}

\newcommand{\insertvsparallel}{
\begin{table}[t]
\resizebox{\textwidth}{!}{%
\renewcommand\arraystretch{1.4}
    \centering
    \begin{tabular}{llllllllllll}
    \toprule
	& mean & 	ts &  	nso &  	lg &  	ee &  	bho &  	bm &  	ff &  	gn &  	ti &  	om \\  
	\hline
    \enxx{} Parallel toks & 375169  & 2296670 &	606853 &	315905 &	275417 &	154023 &	148549 &	130953 &	112958 &	43511 &	42018  \\
    \enxx{} Bilex toks & 16324 & 8015 &	4500 &	8298 &	6755 &	24665 &	24665 &	70854 &	4500 &	6484 &	4500 \\
    \hline \hline
    \multicolumn{11}{c}{\flores{} \enxx{}} \\ 
Baseline &	21.1 & 	33.2 & 	31.9 & 	30.1 & 	26.9 & 	14.7 & 	15 & 	19.1 & 	19.1 & 	5.9 & 	15.5 \\ 
GatiPanlexTokenPairs &	24.4 & 	37.3 & 	31.9 & 	32.7 & 	30.5 & 	25.5 & 	22.6 & 	19.6 & 	19.4 & 	8.1 & 	16.1 \\ 
Parallel &	33.6 & 	45.6 & 	48 & 	39.4 & 	28 & 	40.6 & 	30.7 & 	25 & 	39.2 & 	15.7 & 	24.1 \\ 
Parallel + GatiPanlex &	34.2 & 	45.3 & 	45.3 & 	38.2 & 	31.1 & 	40.8 & 	30.1 & 	24.9 & 	39.9 & 	18.1 & 	28.8 \\

    \multicolumn{11}{c}{\ntlevalset{} \enxx{}} \\ 
Baseline &	20.2 & 	32.1 & 	28.3 & 	27.6 & 	23.5 & 	15.8 & 	13.5 & 	23.9 & 	15.6 & 	6.9 & 	14.6 \\ 
GatiPanlexTokenPairs &	23.9 & 	37.3 & 	28.7 & 	31.3 & 	28.1 & 	26.2 & 	22.5 & 	23.5 & 	16.1 & 	9.3 & 	15.6 \\ 
Parallel &	29.5 & 	42.9 & 	39.9 & 	35.5 & 	26.9 & 	38.5 & 	30.3 & 	23.6 & 	28.2 & 	11.1 & 	18 \\ 
Parallel + GatiPanlexTokenPairs &	30.1 & 	42.6 & 	37.9 & 	35 & 	29.6 & 	38.7 & 	30.4 & 	22.6 & 	30.1 & 	13.1 & 	20.8 \\  
    \bottomrule
    \end{tabular}}
    \caption{Comparing improvements from token-pair data to the oracle: training on trusted parallel data.}
    \label{tab:vsparallel}
\end{table}
}

\newcommand{\insertbangforbuck}{
\begin{table}[t]
\resizebox{\textwidth}{!}{%
\renewcommand\arraystretch{1.4}
    \centering
    \begin{tabular}{lllllllllllllll}
    \toprule
\textsc{Total Tokens} &	&	&	1.8M &	2.1M &	2.6M &	3.6M &	3.7M &	5.4M &	6.2M &	14.7M &	14.8M &	16M &	16.9M &	26.1M \\
&	$\mu_{lrl}$ &	$\mu_{url}$ &	ff &	mni-Mtei &	kri &	doi &	bm &	ay &	gom &	bho &	kl &	ee &	qu &	ts \\
$\Delta$ \flores{} &	\textbf{+1.6} &	\textbf{+4.9} &	+0.1 & 	- & 	- & 	- & 	+5.0 & 	+0.8 & 	- & 	+4.5 & 	- & 	+2.9 & 	- & 	+4.2 \\ 
$\Delta$ \ntlevalset{} &	\textbf{+2.3} &	\textbf{+5.2} &	+0.0 & 	+0.7 & 	+5.1 & 	+1.5 & 	+7.6 & 	-0.1 & 	+11.0 & 	+5.1 & 	+4.8 & 	+3.9 & 	+1.7 & 	+5.3 \\
\hline											\textsc{Total Tokens}	&	26.4M &	27M &	28M &	40M &	41M &	52M &	52M &	80M &	115M &	124M &	157M &	167M &	204M &	505M \\
&	ak &	mai &	ln &	lg &	gn &	nso &	ilo &	ti &	om &	sa &	dv &	lus &	as &	ckb \\
$\Delta$ \flores{} &	- &	-1.2 & 	+1.3 & 	+1.6 & 	-0.4 & 	-0.5 & 	+7.0 & 	+0.9 & 	-0.2 & 	+0.2 & 	- & 	+8.6 & 	+3.0 & 	+3.3 \\ 
$\Delta$ \ntlevalset{} &	+2.0 & 	-1.2 & 	1.2 & 	+3.0 & 	+0.5 & 	+0.2 & 	+8.0 & 	+1.7 & 	+0.1 & 	+0.4 & 	+6.1 & 	+9.3 & 	+3.4 & 	+2.7 \\ 

    \bottomrule
    \end{tabular}}
    \caption{Improvements on languages with \gatitos{} data when training ONLY on \gatitos{} data in the \enxx{} direction. Sorted by total training tokens (mono, parallel, and bilex), in millions.}
    \label{tab:bangforbuck}
\end{table}
}

\newcommand{\insertbigbig}{
\begin{table*}[]
\renewcommand\arraystretch{1.4}
    \centering
    \begin{tabular}{lllllll}
& 	HRL & 	MRL & 	LRL & 	URL & 	LRL$_{GAT}$ &	URL$_{GAT}$ \\
\toprule						
 \multicolumn{7}{c}{\flores{} \enxx{}} \\ 
 \hline
BaselineBig &	56.1 &	51.2 &	38.5 &	35.5 &	24.4 &	36.0 \\
\hline						
GatiPanlexTokenPairsBig &	 \cellcolor{scale4} -0.5 &	 \cellcolor{scale5} -0.3 &	\cellcolor{scale6} +0.1 &	 \cellcolor{scale5} -0.3 &	\cellcolor{scale8} +1.2 &	\cellcolor{scale9} +2.2 \\
CodeswitchMonoGatiPanlexBig &	\cellcolor{scale6} +0.1 &	\cellcolor{scale6} +0.1 &	\cellcolor{scale6} +0.3 &	 \cellcolor{scale3} -0.3 &	\cellcolor{scale7} +0.9 &	\cellcolor{scale10} +3.4 \\
 \multicolumn{7}{c}{\ntlevalset{} \enxx{}} \\ 	
\hline
BaselineBig &	- &	34.1 &	26.9 &	25.5 &	23.9 &	27.1 \\
\hline						
GatiPanlexTokenPairsBig &	- &	\cellcolor{scale6} +0.2 &	\cellcolor{scale7} +0.6 &	\cellcolor{scale7} +0.7 &	\cellcolor{scale8} +1.6 &	\cellcolor{scale10} +3.5 \\
CodeswitchMonoGatiPanlexBig &	- &	\cellcolor{scale7} +0.4 &	\cellcolor{scale7} +0.7 &	\cellcolor{scale6} +0.2 &	\cellcolor{scale8} +1.4 &	\cellcolor{scale10} +4.2 \\
 \multicolumn{7}{c}{\flores{} \xxen{}} \\ 		
 \hline	
BaselineBig &	62.7 &	58.1 &	50.2 &	44.4 &	41.3 &	41.5 \\
\hline						
GatiPanlexTokenPairsBig &	 \cellcolor{scale4} -0.6 &	 \cellcolor{scale4} -0.5 &	 \cellcolor{scale5} -0.2 &	 \cellcolor{scale5} -0.0 &	\cellcolor{scale6} +0.2 &	\cellcolor{scale8} +1.1 \\
CodeswitchMonoGatiPanlexBig &	\cellcolor{scale6} +0.0 &	 \cellcolor{scale5} -0.2 &	 \cellcolor{scale4} -0.6 &	 \cellcolor{scale3} -1.1 &	 \cellcolor{scale4} -0.6 &	\cellcolor{scale7} +0.4 \\
  \multicolumn{7}{c}{\ntlevalset{} \xxen{}} \\ 	
  \hline	
BaselineBig &- 	& 51.2 &	35.9 &	32.9 &	33.9 &	38.8 \\
\hline						
GatiPanlexTokenPairsBig &	- &	 \cellcolor{scale4} -0.5 &	 \cellcolor{scale5} -0.0 &	\cellcolor{scale8} +1.1 &	\cellcolor{scale7} +0.4 &	\cellcolor{scale8} +1.7 \\
CodeswitchMonoGatiPanlexBig &	- &	 \cellcolor{scale4} -0.6 &	 \cellcolor{scale4} -0.7 &	\cellcolor{scale6} +0.2 &	 \cellcolor{scale4} -0.4 &	\cellcolor{scale7} +0.4 \\
    \bottomrule
    \end{tabular}
    \caption{Average \chrf{} scores by resource category for the larger models, reported in delta relative to baseline. These models are trained with 10x the data and 3x the parameters. We see that the gains as a whole are washed out somewhat, but for those low-resource and unsupervised language pairs with the more trusted \gatitos{} data (last two columns), there is still a noticeable gain.}
    \label{tab:bigbig}
\end{table*}
}

\newcommand{\insertpanlexq}{

\begin{table}[]
 \centering
 \begin{tabular}{|p{1cm}|p{3cm}|p{2cm}|p{5.1cm}|}
 \hline
  & Win \xxen{} & Neut. \xxen{} & Loss \xxen{} \\
 \hline
  Win \enxx{}  &	 \cellcolor{scale8} aa ace av \textbf{ay} bci \textbf{bm} bo cv \textbf{doi} \textbf{dv} dyu dz \textbf{ee gn gom ilo} kbp \textbf{kl} \textbf{kri lg lus} mad min \textbf{mni-Mtei nso} nus \textbf{om} quc quy sg \textbf{ts} yua &	 \cellcolor{scale7} cjk \textbf{ckb} ny \textbf{ti} tn &	 \cellcolor{scale6} acm acq aeb af am apc ar ar-MA arz \textbf{as} awa az ba bbc be bg \textbf{bho} bn bs ca ce ceb cs cy da de el eo et eu fa-AF fi fil fj fo fr ga gl gu hr hu hy id is iso it iw ja jv ka kk km kn ko ks ku lb lo lt ltg lv mag mfa mg mk ml mn mni mr ms mt nl no pa pt rn ro ru rw scn si sk sl sn so sq sr su sv sw ta te tg th tr tt uk ur uz vi war xh yue zh zu \\
\hline			
  Neut. \enxx{}  &	 \cellcolor{scale7} \textbf{ak} bug pag \textbf{qu} tpi &	 \cellcolor{scale6} bew kmb or pcm sat-Beng sm &	 \cellcolor{scale4} ban brx-Beng es fa gd ha hi ht ig ky \textbf{ln} \textbf{mai} mi my ne oc pap pl ps sd skr st tk ug vec wo yi yo zza \\
\hline			
  Loss \enxx{}  &	 \cellcolor{scale6} din \textbf{ff} fon kg \textbf{sa} tum &	 \cellcolor{scale4} ady ber kac &	 \cellcolor{scale3} ahr ber-Latn hne la shn \\
 \hline
 \end{tabular}
    \caption{Languages sorted by whether it helps or hurts to include PanLex and \gatitos{}, for the smaller models (Transformer Big, $475$M). \gatitos{} languages bolded.}
  \label{tab:panlexqsmall}
 \end{table}

\begin{table}[]
 \centering
 \begin{tabular}{|p{1cm}|p{2.5cm}|p{5cm}|p{4cm}|}
 \hline
  & Win \xxen{} & Neut. \xxen{} & Loss \xxen{} \\
 \hline
  Win \enxx{}  &	 \cellcolor{scale8} \textbf{ay bm} cv \textbf{dv ee gom ilo kl kri} ltg \textbf{mni-Mtei sa} &	 \cellcolor{scale7} awa bci din \textbf{doi} fo ga \textbf{lus} mag quy tt &	 \cellcolor{scale6} aeb ahr \textbf{ak} ba \textbf{ckb} dyu kg \textbf{lg ln} mfa pag pap rn sat-Beng skr su \textbf{ts} yua \\
\hline			
  Neut. \enxx{}  &	 \cellcolor{scale7} ce km mg tpi zza &	 \cellcolor{scale6} af am ar az ban be bg \textbf{bho} bn bs ca cs cy da de el eo es et eu fa fa-AF \textbf{ff} fi fil fr gd gl gu ha hi hne hr ht hu hy id is it iw ja ka kk kn ko ky lb lo lt lv mad mi mk ml mn mr ms mt ne nl no ny pa pcm pl pt \textbf{qu} ro ru si sk sl so sq sr sv sw ta te tg th tk tr uk ur uz vi yi yo zh zu &	 \cellcolor{scale4} ace acm acq apc ar-MA arz bew brx-Beng bug ceb dz fj ig jv ks \textbf{nso} oc \textbf{om} or ps scn sd sm sn st ug vec xh yue \\
\hline			
  Loss \enxx{}  &	 \cellcolor{scale6} ady av ber-Latn min mni &	 \cellcolor{scale4} aa bbc cjk \textbf{gn} kmb my sg tn tum &	 \cellcolor{scale3} \textbf{as} ber bo fon iso kac kbp ku la \textbf{mai} nus quc rw shn \textbf{ti} war wo \\

 \hline
 \end{tabular}
    \caption{Languages sorted by whether it helps or hurts to include PanLex and \gatitos{}, for the bigger models (Transformer 1.6B). \gatitos{} languages bolded.}
 \label{tab:panlexqbig}
 \end{table}
}

\newcommand{\insertregressiontab}{
\begin{table}[]
    \centering
    \begin{tabularx}{\textwidth}{lXXXXX}
    \toprule
         & No. PanLex & No. \textsc{Gat.} & No. mono & Intercept & Adjusted $R^2$ \\\hline
         $\Delta$\flores{} & $6.55$e-$5$ & $1.96$e-$4$ & $-6.71$e-$7$ & $1.93$ & $0.40$ \\
         $\Delta$\ntlevalset{} & $6.59$e-$6$ & $2.58$e-$4$ & $-7.78$e-$8$ & $2.91$ & $0.23$ \\
    \bottomrule
    \end{tabularx}
    \caption{Linear regression results in the \enxx{} direction. Here, the dependent variable is $\Delta \chrf{}$ over the baseline, and the independent variables are: number of PanLex word pairs for a given language (No. PanLex), number of \gatitos{} word pairs (No. \gatitos{}), and number of monolingual sentences (No. mono).}
    \label{tab:tokensregression}
\end{table}
}

\newcommand{\inserthitrate}{
\begin{table}[t]
\resizebox{\textwidth}{!}{%
\renewcommand\arraystretch{1.4}
    \centering
    \begin{tabular}{l|llll|llll}
    \toprule
\multicolumn{1}{c}{} & \multicolumn{4}{c}{\flores{}} & \multicolumn{4}{c}{\ntlevalset{}} \\\hline
cat. &	 A $\in$ GAT. &	A $\notin$ GAT. &	colors &	\#s &	A $\in$ GAT. &	A  $\notin$ GAT. &	colors &	\#s \\
BaselineBig &	36.6 &	36.6 &	55.3 &	63.3 &	33.4 &	25.8 &	35.4 &	53.5 \\
CodeswitchMonoGatiPanlexBig &	45.5 &	40.7 &	66.5 &	66.8 &	44.9 &	32.5 &	47.8 &	58.0 \\
$\Delta$ &	+8.9 &	+4.0 &	+11.1 &	+3.6 &	+11.5 &	+6.7 &	+12.4 &	4.5 \\
    \bottomrule
    \end{tabular}}
    \caption{Comparing token hit-rate on classes of nouns known to have issues for UNMT models, along with a weak control of numbers. There are large improvements for animals in the \gatitos{} training lexicon ($A \in  GAT$), as well as their complementary distribution, animals not in \gatitos{} ($A \not\in  GAT$), and colors. Number hit-rate has a minor bump.}
    \label{tab:hitrate}
\end{table}
}

%% file: tables_appendix.tex
\newcommand{\insertmodeldescriptionsfull}{

\begin{table}[]
 \centering
 \begin{tabular}{ll|p{5cm}}
Model Name &	Abbr. &	Description \\
\hline
Baseline &	B &	Trained on MASS $+$ translation \\
GatiPanlexTokenPairs &	T &	Baseline $+$ 5\% token pairs \\
GatiPanlexTokenPairsSamp75 &	T$_{75}$ &	token-pairs sampled to 75\% \\
GatiPanlexTokenPairsSamp50 &	T$_{50}$ &	token-pairs sampled to 50\% \\
GatiPanlexTokenPairsSamp25 &	T$_{25}$ &	token-pairs sampled to 25\% \\
GatiTokenPairs &	T$_{GAT}$ &	token pairs with only \gatitos \\
CodeswitchMono &	C$_{M}$ &	See \S \ref{methods} \\
CodeswitchParallel &	C$_{P}$ &	See \S \ref{methods} \\
CodeswitchMonoParallel &	C$_{MP}$ &	See \S \ref{methods} \\
CodeswitchMonoGatiPanlex &	C$_{M}$\textnormal{T} &	See \S \ref{methods} \\
GlowupMono &	G$_{M}$ &	See \S \ref{methods} \\
GlowupParallel &	G$_{P}$ &	See \S \ref{methods} \\
GlowupMonoParallel &	G$_{MP}$ &	See \S \ref{methods} \\
GlowupMonoGatiPanlex &	G$_{M}$\textnormal{T} &	See \S \ref{methods} \\
BaselineBig &	B$_{BIG}$ &	Big version of the Baseline (\S \ref{sec:bigger_models}) \\
GatiPanlexTokenPairsBig &	T$_{BIG}$ &	Big version of the TP (\S \ref{sec:bigger_models}) \\
CodeswitchMonoGatiPanlexBig &	(C$_{M}+$\textnormal{T})$_{BIG}$ &	Big version of C$_{M}$\textnormal{T} (\S \ref{sec:bigger_models}) \\
NLLB &	NL &	NLLB 54B model \\
 \hline
 \end{tabular}
    \caption{Names, abbreviations, and descriptions of the full model results in Tables \ref{tab:fullresults-flores-xxen} and \ref{tab:fullresults-flores-enxx}.}
  \label{tab:modeldescriptions}
 \end{table}
 }

\newcommand{\insertlanguageslist}{

{\fontsize{6.4}{8}\selectfont
\begin{xltabular}[H]{\linewidth}{lllllllllll}
\toprule
    \centering
         BCP-47 & Language & Cat. & Mono & Parallel & PanLex & \gatitos & Speak. & Script & Cont. & Family   \\\hline
         \endhead
en &	English &	HRL &	738.8M &	4726.4M &	3.6M &	0 &	984M &	Latn &	Europe &	Indo-European \\
es &	Spanish &	HRL &	175.1M &	585.6M &	2.4M &	4K &	528M &	Latn &	Europe &	Indo-European \\
de &	German &	HRL &	169.3M &	389.3M &	2.8M &	4K &	130M &	Latn &	Europe &	Indo-European \\
id &	Indonesian &	HRL &	95M &	97.4M &	1.1M &	4K &	198M &	Latn &	Asia &	Austronesian \\
yue &	Cantonese &	MRL &	83.1M &	405K &	180K &	0 &	84M &	Hant &	Asia &	Sino-Tibetan \\
hu &	Hungarian &	HRL &	72.4M &	50.9M &	1.3M &	0 &	13M &	Latn &	Europe &	Indo-European \\
pl &	Polish &	HRL &	68.7M &	152.8M &	1.4M &	4K &	41M &	Latn &	Europe &	Indo-European \\
zh &	Mandarin &	HRL &	67.6M &	215.7M &	6K &	4K &	1092M &	Hans &	Asia &	Sino-Tibetan \\
mi &	Maori &	MRL &	67.4M &	1.3M &	474K &	0 &	50K &	Latn &	Oceania &	Austronesian \\
ko &	Korean &	HRL &	67.2M &	128.1M &	1.3M &	5K &	77M &	Kore &	Asia &	Koreanic \\
ja &	Japanese &	HRL &	65.2M &	307.7M &	2.2M &	4K &	128M &	Jpan &	Asia &	Japonic \\
lo &	Lao &	MRL &	59.4M &	817K &	184K &	0 &	30M &	Laoo &	Asia &	Kra-Dai \\
ru &	Russian &	HRL &	57M &	294.7M &	2.8M &	4K &	268M &	Cyrl &	Europe &	Indo-European \\
gd &	Scottish Gaelic &	MRL &	56.9M &	4M &	324K &	0 &	57K &	Latn &	Europe &	Indo-European \\
tr &	Turkish &	HRL &	56.5M &	159.2M &	1.3M &	4K &	71M &	Latn &	Asia &	Turkic \\
so &	Somali &	MRL &	56.4M &	1.3M &	112K &	0 &	16M &	Latn &	Africa &	Afro-Asiatic \\
th &	Thai &	HRL &	53.1M &	69.1M &	1.6M &	4K &	61M &	Thai &	Asia &	Kra-Dai \\
ha &	Hausa &	MRL &	50.6M &	1.8M &	202K &	0 &	80M &	Latn &	Africa &	Afro-Asiatic \\
it &	Italian &	HRL &	49.1M &	245.5M &	2M &	4K &	66M &	Latn &	Europe &	Indo-European \\
pt &	Portuguese &	HRL &	48.5M &	240.7M &	1.7M &	4K &	230M &	Latn &	Europe &	Indo-European \\
vi &	Vietnamese &	HRL &	47.7M &	94.2M &	825K &	4K &	68M &	Latn &	Asia &	Austroasiatic \\
fr &	French &	HRL &	45M &	481.6M &	2.5M &	4K &	230M &	Latn &	Europe &	Indo-European \\
ceb &	Cebuano &	MRL &	43.9M &	9.2M &	62K &	0 &	20M &	Latn &	Asia &	Austronesian \\
yo &	Yoruba &	MRL &	43.4M &	847K &	244K &	0 &	50M &	Latn &	Africa &	Niger-Congo \\
sd &	Sindhi &	MRL &	42.3M &	1.6M &	43K &	0 &	26M &	Arab &	Asia &	Indo-European \\
co &	Corsican &	MRL &	41.2M &	1.5M &	148K &	0 &	150K &	Latn &	Europe &	Indo-European \\
mg &	Malagasy &	MRL &	40.2M &	3.6M &	116K &	0 &	25M &	Latn &	Africa &	Austronesian \\
ms &	Malay &	HRL &	39.4M &	53.8M &	0 &	0 &	77M &	Latn &	Asia &	Austronesian \\
ar-MA &	Mor. Arabic &	URL &	35.5M &	0 &	0 &	0 &	52M &	Arab &	Africa &	Afro-Asiatic \\
bew &	Betawi &	URL &	33.3M &	0 &	5K &	0 &	5M &	Latn &	Asia &	Malay Creole \\
ny &	Nyanja &	MRL &	32.9M &	1.2M &	26K &	0 &	12M &	Latn &	Africa &	Niger-Congo \\
nl &	Dutch &	HRL &	32M &	258M &	1.6M &	4K &	22M &	Latn &	Europe &	Indo-European \\
uk &	Ukrainian &	HRL &	32M &	75.3M &	892K &	0 &	35M &	Cyrl &	Europe &	Indo-European \\
sv &	Swedish &	HRL &	31.1M &	122.6M &	1.3M &	0 &	12M &	Latn &	Europe &	Indo-European \\
haw &	Hawaiian &	MRL &	30.4M &	698K &	156K &	0 &	24K &	Latn &	Americas &	Austronesian \\
ro &	Romanian &	HRL &	30.1M &	45M &	841K &	0 &	24M &	Latn &	Europe &	Indo-European \\
cs &	Czech &	HRL &	30M &	106.9M &	1.6M &	0 &	13M &	Latn &	Europe &	Indo-European \\
hmn &	Hmong &	MRL &	27.7M &	4.9M &	4K &	0 &	4M &	Latn &	Europe &	Hmong-Mien \\
yi &	Yiddish &	MRL &	27.6M &	760K &	0 &	0 &	2M &	Hebr &	Europe &	Indo-European \\
fa &	Persian &	HRL &	27.5M &	45.2M &	0 &	0 &	53M &	Arab &	Asia &	Indo-European \\
ig &	Igbo &	MRL &	27.4M &	647K &	48K &	0 &	27M &	Latn &	Africa &	Niger-Congo \\
lv &	Latvian &	MRL &	26M &	22.4M &	0 &	0 &	2M &	Latn &	Europe &	Indo-European \\
ar &	Arabic &	HRL &	25.8M &	116.7M &	0 &	4K &	310M &	Arab &	Asia &	Afro-Asiatic \\
ckb &	Sorani &	MRL &	25.1M &	155K &	53K &	4K &	7M &	Arab &	Asia &	Indo-European \\
tt &	Tatar &	MRL &	25M &	557K &	128K &	0 &	5M &	Cyrl &	Europe &	Turkic \\
sm &	Samoan &	MRL &	23.9M &	502K &	58K &	0 &	510K &	Latn &	Oceania &	Austronesian \\
zu &	Zulu &	MRL &	23.4M &	2.3M &	144K &	0 &	12M &	Latn &	Africa &	Niger-Congo \\
no &	Norwegian &	HRL &	23.1M &	85.8M &	4K &	0 &	5M &	Latn &	Europe &	Indo-European \\
st &	Sesotho &	MRL &	22.6M &	1.2M &	35K &	0 &	6M &	Latn &	Africa &	Niger-Congo \\
ta &	Tamil &	MRL &	22M &	11.1M &	247K &	0 &	76M &	Taml &	Asia &	Dravidian \\
or &	Odia (Oriya) &	MRL &	21.2M &	169K &	16K &	0 &	35M &	Orya &	Asia &	Indo-European \\
sn &	Shona &	MRL &	18.5M &	958K &	102K &	0 &	8M &	Latn &	Africa &	Niger-Congo \\
bo &	Tibetan &	LRL &	17.8M &	282K &	56K &	0 &	1M &	Tibt &	Asia &	Sino-Tibetan \\
el &	Greek &	HRL &	17.3M &	54M &	1.2M &	0 &	13M &	Grek &	Europe &	Indo-European \\
fi &	Finnish &	HRL &	17.1M &	48.6M &	2.1M &	0 &	6M &	Latn &	Europe &	Uralic \\
hi &	Hindi &	HRL &	16.5M &	75.7M &	449K &	4K &	381M &	Deva &	Asia &	Indo-European \\
xh &	Xhosa &	LRL &	15.8M &	697K &	57K &	0 &	8M &	Latn &	Africa &	Niger-Congo \\
mr &	Marathi &	MRL &	15.6M &	8.1M &	154K &	0 &	75M &	Deva &	Asia &	Indo-European \\
sk &	Slovak &	HRL &	15.6M &	63.9M &	1.1M &	0 &	7M &	Latn &	Europe &	Indo-European \\
hy &	Armenian &	MRL &	15.4M &	6.9M &	751K &	0 &	5M &	Armn &	Asia &	Indo-European \\
kk &	Kazakh &	MRL &	15.3M &	6.6M &	241K &	0 &	13M &	Cyrl &	Asia &	Turkic \\
da &	Danish &	HRL &	15.2M &	78.9M &	545K &	0 &	6M &	Latn &	Europe &	Indo-European \\
mk &	Macedonian &	MRL &	15M &	6.6M &	363K &	0 &	2M &	Cyrl &	Europe &	Indo-European \\
bg &	Bulgarian &	MRL &	14.9M &	37.4M &	693K &	0 &	8M &	Cyrl &	Europe &	Indo-European \\
sr &	Serbian &	MRL &	14.1M &	30.6M &	206K &	0 &	8M &	Cyrl &	Europe &	Indo-European \\
ml &	Malayalam &	MRL &	13.9M &	6.4M &	163K &	0 &	34M &	Mlym &	Asia &	Dravidian \\
az &	Azerbaijani &	MRL &	13.5M &	19.6M &	0 &	0 &	23M &	Latn &	Asia &	Turkic \\
is &	Icelandic &	MRL &	13.4M &	15.8M &	620K &	0 &	310K &	Latn &	Europe &	Indo-European \\
te &	Telugu &	MRL &	12.7M &	8M &	311K &	0 &	79M &	Telu &	Asia &	Dravidian \\
ne &	Nepali &	MRL &	11.6M &	9.7M &	0 &	0 &	16M &	Deva &	Asia &	Indo-European \\
mzn &	Mazanderani &	URL &	11.6M &	0 &	29K &	0 &	6M &	Arab &	Asia &	Indo-European \\
meo &	Kedah Malay &	URL &	11.3M &	0 &	0 &	0 &	3M &	Latn &	Asia &	Austronesian \\
et &	Estonian &	MRL &	11.2M &	30.1M &	0 &	0 &	1M &	Latn &	Europe &	Uralic \\
iw &	Hebrew &	HRL &	10.8M &	57.1M &	707K &	0 &	5M &	Hebr &	Asia &	Afro-Asiatic \\
rw &	Kinyarwanda &	LRL &	10.1M &	803K &	56K &	0 &	10M &	Latn &	Africa &	Niger-Congo \\
mn &	Mongolian &	LRL &	10M &	4.1M &	0 &	0 &	5M &	Cyrl &	Asia &	Mongolic \\
ur &	Urdu &	MRL &	9.9M &	15.2M &	400K &	0 &	163M &	Arab &	Asia &	Indo-European \\
apc &	N. Lev. Arabic &	URL &	9.7M &	0 &	0 &	0 &	15M &	Arab &	Asia &	Afro-Asiatic \\
hr &	Croatian &	MRL &	9.7M &	17.4M &	738K &	0 &	7M &	Latn &	Europe &	Indo-European \\
fil &	Filipino &	MRL &	9.5M &	25.3M &	61K &	0 &	45M &	Latn &	Asia &	Austronesian \\
as &	Assamese &	LRL &	9.3M &	575K &	78K &	4K &	15M &	Beng &	Asia &	Indo-European \\
arz &	Egyptian Arabic &	URL &	9.2M &	0 &	101K &	0 &	58M &	Arab &	Africa &	Afro-Asiatic \\
fo &	Faroese &	LRL &	9.2M &	26K &	254K &	0 &	66K &	Latn &	Europe &	Indo-European \\
pap &	Papiamento &	URL &	9.1M &	0 &	99K &	0 &	341K &	Latn &	Americas &	Portuguese Creole \\
fa-AF &	Dari &	LRL &	8.7M &	10K &	0 &	0 &	21M &	Arab &	Asia &	Indo-European \\
acm &	Mesop. Arabic &	URL &	8.7M &	0 &	11K &	0 &	15M &	Arab &	Asia &	Afro-Asiatic \\
lt &	Lithuanian &	MRL &	8.6M &	30.9M &	611K &	0 &	3M &	Latn &	Europe &	Indo-European \\
lus &	Mizo &	URL &	8.3M &	0 &	85K &	4K &	688K &	Latn &	Asia &	Sino-Tibetan \\
be &	Belarusian &	LRL &	8.3M &	6.5M &	508K &	0 &	3M &	Cyrl &	Europe &	Indo-European \\
kn &	Kannada &	LRL &	8M &	5.8M &	116K &	0 &	47M &	Knda &	Asia &	Dravidian \\
dv &	Dhivehi &	LRL &	7.9M &	1K &	34K &	4K &	300K &	Thaa &	Asia &	Indo-European \\
my &	Burmese &	LRL &	7.8M &	5.5M &	121K &	0 &	43M &	Mymr &	Asia &	Sino-Tibetan \\
oc &	Occitan &	LRL &	7.5M &	6K &	2.4M &	0 &	500K &	Latn &	Europe &	Indo-European \\
bn &	Bengali &	MRL &	7.3M &	21.7M &	259K &	0 &	262M &	Beng &	Asia &	Indo-European \\
af &	Afrikaans &	MRL &	7.1M &	12.7M &	258K &	0 &	18M &	Latn &	Africa &	Indo-European \\
eu &	Basque &	LRL &	7M &	6.4M &	792K &	0 &	540K &	Latn &	Europe &	Language isolate \\
gu &	Gujarati &	LRL &	6.9M &	5.8M &	198K &	0 &	47M &	Gujr &	Asia &	Indo-European \\
gl &	Galician &	MRL &	6.4M &	13.1M &	383K &	0 &	2M &	Latn &	Europe &	Indo-European \\
sa &	Sanskrit &	LRL &	6.2M &	11K &	168K &	4K &	100K &	Deva &	Asia &	Indo-European \\
sl &	Slovenian &	MRL &	5.9M &	27M &	672K &	0 &	2M &	Latn &	Europe &	Indo-European \\
ug &	Uyghur &	LRL &	5.7M &	526K &	116K &	0 &	10M &	Arab &	Asia &	Turkic \\
ba &	Bashkir &	LRL &	5.6M &	303K &	112K &	0 &	1M &	Cyrl &	Europe &	Turkic \\
si &	Sinhala &	LRL &	5.6M &	6.5M &	88K &	0 &	16M &	Sinh &	Asia &	Indo-European \\
om &	Oromo &	LRL &	5.6M &	203K &	10K &	4K &	24M &	Latn &	Africa &	Afro-Asiatic \\
zza &	Zaza &	URL &	5.3M &	0 &	0 &	0 &	2M &	Latn &	Asia &	Indo-European \\
uz &	Uzbek &	MRL &	5.3M &	11.2M &	0 &	0 &	34M &	Latn &	Asia &	Turkic \\
sw &	Swahili &	MRL &	5.2M &	10.5M &	0 &	0 &	150M &	Latn &	Africa &	Niger-Congo \\
km &	Khmer &	MRL &	5.1M &	8M &	188K &	0 &	17M &	Khmr &	Asia &	Austroasiatic \\
ky &	Kyrgyz &	LRL &	4.9M &	2.9M &	174K &	0 &	5M &	Cyrl &	Asia &	Turkic \\
am &	Amharic &	LRL &	4.7M &	2.8M &	71K &	0 &	26M &	Ethi &	Africa &	Afro-Asiatic \\
vec &	Venetian &	URL &	4.4M &	0 &	202K &	0 &	4M &	Latn &	Europe &	Indo-European \\
ca &	Catalan &	MRL &	4.4M &	32.9M &	803K &	0 &	9M &	Latn &	Europe &	Indo-European \\
tk &	Turkmen &	LRL &	4M &	416K &	177K &	0 &	7M &	Latn &	Asia &	Turkic \\
ti &	Tigrinya &	LRL &	3.9M &	67K &	45K &	4K &	8M &	Ethi &	Africa &	Afro-Asiatic \\
pa &	Punjabi &	LRL &	3.7M &	3.3M &	89K &	0 &	29M &	Guru &	Asia &	Indo-European \\
sq &	Albanian &	MRL &	3.2M &	10.6M &	269K &	0 &	13M &	Latn &	Europe &	Indo-European \\
ka &	Georgian &	MRL &	3M &	11.7M &	482K &	0 &	4M &	Geor &	Asia &	Kartvelian \\
cv &	Chuvash &	URL &	2.8M &	0 &	121K &	0 &	1M &	Cyrl &	Europe &	Turkic \\
ilo &	Ilocano &	URL &	2.6M &	0 &	41K &	4K &	9M &	Latn &	Asia &	Austronesian \\
bal &	Baluchi &	URL &	2.5M &	0 &	16K &	0 &	8M &	Arab &	Asia &	Indo-European \\
eo &	Esperanto &	LRL &	2.4M &	7.5M &	1.3M &	0 &	2M &	Latn &	Europe &	Constructed \\
cy &	Welsh &	LRL &	2.2M &	6.4M &	448K &	0 &	590K &	Latn &	Europe &	Indo-European \\
la &	Latin &	LRL &	2.2M &	2.2M &	740K &	0 &	0 &	Latn &	Europe &	Indo-European \\
dz &	Dzongkha &	LRL &	2.1M &	260K &	31K &	0 &	200K &	Tibt &	Asia &	Sino-Tibetan \\
mt &	Maltese &	LRL &	2.1M &	7.3M &	247K &	0 &	470K &	Latn &	Europe &	Afro-Asiatic \\
tn &	Tswana &	LRL &	2M &	66K &	100K &	0 &	8M &	Latn &	Africa &	Niger-Congo \\
lg &	Luganda &	LRL &	2M &	3K &	31K &	4K &	4M &	Latn &	Africa &	Niger-Congo \\
ht &	Haitian &	LRL &	2M &	3.4M &	177K &	0 &	8M &	Latn &	Americas &	French Creole \\
nso &	Sepedi &	LRL &	1.9M &	798K &	13K &	4K &	5M &	Latn &	Africa &	Niger-Congo \\
ps &	Pashto &	LRL &	1.8M &	2.1M &	0 &	0 &	50M &	Arab &	Asia &	Indo-European \\
za &	Zhuang &	URL &	1.7M &	0 &	0 &	0 &	15M &	Latn &	Asia &	Kra-Dai \\
ga &	Irish &	LRL &	1.7M &	4M &	428K &	0 &	1M &	Latn &	Europe &	Indo-European \\
tpi &	Tok Pisin &	LRL &	1.7M &	3.3M &	62K &	0 &	120K &	Latn &	Oceania &	English Creole \\
pcm &	Nigerian Pidgin &	LRL &	1.6M &	24K &	4K &	0 &	40M &	Latn &	Africa &	English Creole \\
lb &	Luxembourgish &	LRL &	1.5M &	4.6M &	183K &	0 &	420K &	Latn &	Europe &	Indo-European \\
ku &	Kurmanji &	LRL &	1.5M &	2.1M &	0 &	0 &	15M &	Latn &	Asia &	Indo-European \\
tg &	Tajik &	LRL &	1.4M &	1.5M &	194K &	0 &	8M &	Cyrl &	Asia &	Indo-European \\
ln &	Lingala &	LRL &	1.4M &	5K &	97K &	4K &	58M &	Latn &	Africa &	Niger-Congo \\
ce &	Chechen &	URL &	1.4M &	0 &	113K &	0 &	1M &	Cyrl &	Europe &	NE Caucasian \\
mai &	Maithili &	URL &	1.3M &	0 &	0 &	4K &	65M &	Deva &	Asia &	Indo-European \\
jv &	Javanese &	LRL &	1.3M &	6.2M &	128K &	0 &	84M &	Latn &	Asia &	Austronesian \\
ts &	Tsonga &	LRL &	1.3M &	2K &	9K &	4K &	13M &	Latn &	Africa &	Niger-Congo \\
fj &	Fijian &	LRL &	1.3M &	6K &	44K &	0 &	339K &	Latn &	Oceania &	Austronesian \\
ak &	Twi &	LRL &	1.3M &	38K &	115K &	4K &	11M &	Latn &	Africa &	Niger-Congo \\
ber-Latn &	Tamazight &	URL &	1.2M &	0 &	2K &	0 &	30M &	Latn &	Africa &	Afro-Asiatic \\
su &	Sundanese &	LRL &	1.2M &	2.7M &	90K &	0 &	34M &	Latn &	Asia &	Austronesian \\
fy &	Western Frisian &	LRL &	1.1M &	4.8M &	90K &	0 &	850K &	Latn &	Europe &	Indo-European \\
skr &	Saraiki &	URL &	974K &	0 &	0 &	0 &	20M &	Arab &	Asia &	Indo-European \\
bbc &	Batak Toba &	URL &	932K &	0 &	23K &	0 &	2M &	Latn &	Asia &	Austronesian \\
war &	Waray (PHs) &	URL &	902K &	0 &	48K &	0 &	3M &	Latn &	Asia &	Austronesian \\
gn &	Guarani &	LRL &	861K &	1.3M &	6K &	4K &	5M &	Latn &	Americas &	Tupian \\
qu &	Quechua &	LRL &	842K &	2K &	46K &	4K &	9M &	Latn &	Americas &	Quechuan \\
bug &	Buginese &	URL &	797K &	0 &	19K &	0 &	6M &	Latn &	Asia &	Austronesian \\
ee &	Ewe &	LRL &	796K &	4K &	90K &	4K &	4M &	Latn &	Africa &	Niger-Congo \\
ltg &	Latgalian &	URL &	796K &	0 &	34K &	0 &	170K &	Latn &	Europe &	Indo-European \\
kl &	Kalaallisut &	LRL &	741K &	500 &	48K &	4K &	56K &	Latn &	Americas &	Eskimo-Aleut \\
bho &	Bhojpuri &	LRL &	734K &	4K &	0 &	4K &	60M &	Deva &	Asia &	Indo-European \\
ar-Latn &	Arabic &	URL &	634K &	0 &	4K &	0 &	3M &	Latn &	Asia &	Afro-Asiatic \\
pag &	Pangasinan &	URL &	594K &	0 &	19K &	0 &	1M &	Latn &	Asia &	Austronesian \\
shn &	Shan &	URL &	566K &	0 &	16K &	0 &	3M &	Mymr &	Asia &	Kra-Dai \\
min &	Minangkabau &	URL &	533K &	0 &	12K &	0 &	6M &	Latn &	Asia &	Austronesian \\
cjk &	Chokwe &	URL &	494K &	0 &	8K &	0 &	983K &	Latn &	Africa &	Niger-Congo \\
yua &	Yucateco &	URL &	419K &	0 &	67K &	0 &	766K &	Latn &	Americas &	Mayan \\
sg &	Sango &	URL &	410K &	0 &	32K &	0 &	400K &	Latn &	Africa &	Ngbandi Creole \\
iso &	Isoko &	URL &	409K &	0 &	3K &	0 &	420K &	Latn &	Africa &	Niger-Congo \\
kac &	Kachin &	URL &	402K &	0 &	10K &	0 &	940K &	Latn &	Asia &	Sino-Tibetan \\
kg &	Kongo &	LRL &	376K &	5K &	15K &	0 &	7M &	Latn &	Africa &	Niger-Congo \\
gom &	Goan Konkani &	URL &	311K &	0 &	37K &	4K &	2M &	Deva &	Asia &	Indo-European \\
bs &	Bosnian &	MRL &	311K &	22.6M &	112K &	0 &	2M &	Cyrl &	Europe &	Indo-European \\
av &	Avaric &	URL &	301K &	0 &	216K &	0 &	760K &	Cyrl &	Europe &	Northeast Caucasian \\
tiv &	Tiv &	URL &	297K &	0 &	13K &	0 &	2M &	Latn &	Africa &	Niger-Congo \\
ady &	Adyghe &	URL &	296K &	0 &	25K &	0 &	575K &	Cyrl &	Europe &	Northwest Caucasian \\
wo &	Wolof &	LRL &	289K &	290K &	94K &	0 &	4M &	Latn &	Africa &	Niger-Congo \\
hne &	Chhattisgarhi &	URL &	269K &	0 &	0 &	0 &	18M &	Deva &	Asia &	Indo-European \\
ay &	Aymara &	LRL &	267K &	600 &	91K &	4K &	3M &	Latn &	Americas &	Aymaran \\
quc &	K'iche' &	URL &	250K &	0 &	54K &	0 &	2M &	Latn &	Americas &	Mayan \\
ace &	Achinese &	URL &	226K &	0 &	18K &	0 &	4M &	Latn &	Asia &	Austronesian \\
acq &	Mesop. Arabic &	URL &	216K &	0 &	3K &	0 &	7M &	Arab &	Asia &	Afro-Asiatic \\
fon &	Fon &	URL &	197K &	0 &	8K &	0 &	1M &	Latn &	Africa &	Niger-Congo \\
ban &	Balinese &	LRL &	188K &	9K &	30K &	0 &	3M &	Latn &	Asia &	Austronesian \\
bm &	Bambara &	URL &	187K &	0 &	73K &	4K &	14M &	Latn &	Africa &	Mande \\
doi &	Dogri &	URL &	179K &	0 &	0 &	4K &	2M &	Deva &	Asia &	Indo-European \\
tum &	Tumbuka &	LRL &	171K &	4K &	1K &	0 &	2M &	Latn &	Africa &	Niger-Congo \\
bci &	Baoulé &	URL &	152K &	0 &	20K &	0 &	2M &	Latn &	Africa &	Niger-Congo \\
quy &	Ayacucho Quechua &	URL &	140K &	0 &	94K &	0 &	900K &	Latn &	Americas &	Quechuan \\
mad &	Madurese &	URL &	138K &	0 &	18K &	0 &	7M &	Latn &	Asia &	Austronesian \\
awa &	Awadhi &	URL &	136K &	0 &	0 &	0 &	38M &	Deva &	Asia &	Indo-European \\
dyu &	Dyula &	URL &	130K &	0 &	5K &	4K &	3M &	Latn &	Africa &	Mande \\
kbp &	Kabiyè &	URL &	129K &	0 &	10K &	0 &	1M &	Latn &	Africa &	Niger-Congo \\
kri &	Krio &	URL &	129K &	0 &	10K &	4K &	496K &	Latn &	Africa &	English Creole \\
rn &	Rundi &	LRL &	125K &	2K &	27K &	0 &	9M &	Latn &	Africa &	Niger-Congo \\
mni &	Manipuri &	URL &	106K &	0 &	2K &	0 &	1M &	Beng &	Asia &	Sino-Tibetan \\
mni-Mtei &	Manipuri  &	URL &	103K &	0 &	1K &	4K &	1M &	Mtei &	Asia &	Sino-Tibetan \\
ber &	Tamazight &	URL &	96K &	0 &	0 &	0 &	30M &	Tfng &	Africa &	Afro-Asiatic \\
kmb &	Kimbundu &	URL &	94K &	0 &	7K &	0 &	4M &	Latn &	Africa &	Niger-Congo \\
scn &	Sicilian &	URL &	92K &	0 &	149K &	0 &	5M &	Latn &	Europe &	Indo-European \\
ff &	Fulah &	LRL &	86K &	4K &	30K &	4K &	50M &	Latn &	Africa &	Niger-Congo \\
aa &	Afar &	URL &	82K &	0 &	52K &	0 &	4M &	Latn &	Africa &	Afro-Asiatic \\
ks &	Kashmiri &	LRL &	71K &	1K &	24K &	0 &	6M &	Arab &	Asia &	Indo-European \\
mag &	Magahi &	URL &	66K &	0 &	0 &	0 &	14M &	Deva &	Asia &	Indo-European \\
chr &	Cherokee &	LRL &	63K &	76K &	61K &	0 &	13K &	Cher &	Americas &	Iroquoian \\
din &	Dinka &	URL &	62K &	0 &	3K &	0 &	1M &	Latn &	Africa &	Nilo-Saharan \\
aeb &	Tunisian Arabic &	URL &	48K &	0 &	13K &	0 &	11M &	Arab &	Africa &	Afro-Asiatic \\
ahr &	Ahirani &	URL &	24K &	0 &	0 &	0 &	2M &	Deva &	Asia &	Indo-European \\
nus &	Nuer &	URL &	24K &	0 &	20K &	0 &	890K &	Latn &	Africa &	Nilo-Saharan \\
mfa &	Pattani Malay &	URL &	7K &	0 &	0 &	0 &	1000K &	Arab &	Asia &	Austronesian \\
sat-Beng &	Santali &	URL &	7K &	0 &	0 &	0 &	6M &	Beng &	Asia &	Austroasiatic \\
brx-Beng &	Bodo (India) &	URL &	4K &	0 &	0 &	0 &	1M &	Beng &	Asia &	Sino-Tibetan \\
        \bottomrule
    \label{tab:languageslist}
\end{xltabular}
}
}

\newcommand{\insertcommonerrorstable}{
\begin{table}[H]
    \centering
    \begin{tabular}{lccc}
    \toprule
         & \% question marks & \% near copy & \% repetition  \\\hline
         && \flores & \\
         BaselineBig & 0.7 & 4.5 & 3.4 \\
         GatiPanlexTokenPairsBig & 0.9 & \textbf{3.6} & \textbf{2.7} \\
         CodeswitchMonoGatiPanlexBig & \
         \textbf{0.4} & 4.7 & 3.2 \\\hline
         && \ntlevalset & \\
         BaselineBig & 0.6 & 2.2 & 3.1 \\
         GatiPanlexTokenPairsBig & 0.6 & \textbf{1.8} & \textbf{2.2} \\
         CodeswitchMonoGatiPanlexBig & \textbf{0.4} & 2.3 & 2.6 \\
    \bottomrule
    \end{tabular}
    \caption{The frequencies of three common error types in MT in each of the eval sets, as a percentage of the total sentences in the set that had each issue (lower is better). Exact descriptions of the error types are given in Section \ref{sec:common_errors}.}
    \label{tab:commonerrors}
\end{table}
}

\newcommand{\insertfullenxx}{
\scriptsize
\setlength{\tabcolsep}{3pt}
\begin{longtable}{llllllllllllllllllll}
lang &	cat. &	B &	T &	T$_{75}$ &	T$_{50}$ &	T$_{25}$ &	T$_{Gatitos}$ &	C$_M$ &	C$_P$ &	C$_{MP}$ &	C$_M$ T &	G$_M$ &	G$_P$ &	G$_{MP}$ &	G$_M$ T &	B$_{BIG}$ &	T$_{BIG}$ &	(C$_M$ T)$_{BIG}$ & NLLB \\
\hline	
\endhead
$\mu$ &	mean &	39.7 &	40.0 &	40.0 &	39.6 &	39.8 &	39.7 &	40.4 &	38.3 &	39.5 &	40.7 &	40.8 &	38.6 &	40.0 &	40.7 &	44.6 &	44.4 &	44.7 &	50.7 \\
HRL &	mean &	50.5 &	50.1 &	50.5 &	50.1 &	50.1 &	50.1 &	51.3 &	48.5 &	49.4 &	50.9 &	52.0 &	48.4 &	50.6 &	51.6 &	56.0 &	55.5 &	56.1 &	58.5 \\
MRL &	mean &	46.6 &	46.4 &	46.5 &	46.4 &	46.4 &	46.4 &	47.5 &	44.9 &	45.8 &	47.2 &	48.0 &	44.8 &	46.8 &	47.6 &	51.2 &	50.9 &	51.3 &	53.8 \\
LRL &	mean &	34.4 &	34.9 &	34.9 &	34.6 &	34.6 &	34.6 &	35.2 &	33.3 &	34.4 &	35.7 &	35.4 &	33.4 &	34.5 &	35.6 &	38.5 &	38.6 &	38.8 &	50.3 \\
URL &	mean &	28.7 &	30.2 &	29.5 &	28.8 &	29.5 &	29.1 &	29.2 &	27.8 &	29.9 &	30.2 &	29.2 &	29.2 &	29.7 &	29.5 &	34.8 &	34.5 &	34.5 &	40.1 \\
OOM & mean	 &	6.2 &	12.6 &	10.9 &	13.2 &	12.3 &	12.5 &	13.7 &	9.7 &	12.2 &	12.2 &	14.2 &	8.8 &	14.8 &	10.6 &	11.0 &	12.9 &	13.0 & - \\	
$\Delta_{UR}$ &	mean &	30.4 &	31.6 &	31.1 &	30.6 &	31.0 &	30.8 &	31.0 &	29.4 &	31.3 &	31.9 &	31.1 &	30.4 &	31.2 &	31.3 &	35.9 &	35.7 &	35.8 &	43.2 \\
--- & --- & --- & --- & --- & --- & --- & --- & --- & --- & --- & --- & --- & --- & --- & --- & --- & --- & --- & --- \\
\end{longtable}
}

\newcommand{\insertfullxxen}{
\scriptsize
\setlength{\tabcolsep}{3pt}
\begin{longtable}{llllllllllllllllllll}
lang &	cat. &	B &	T &	T$_{75}$ &	T$_{50}$ &	T$_{25}$ &	T$_{GAT}$ &	C$_M$ &	C$_P$ &	C$_{MP}$ &	C$_M$ \textnormal{T} &	G$_M$ &	G$_P$ &	G$_{MP}$ &	G$_M$ \textnormal{T} &	B$_{BIG}$ &	T$_{BIG}$ &	(C$_M$ \textnormal{T})$_{BIG}$ &	NLLB \\
\hline								
\endhead 
$\mu$ &	mean &	47.2 &	47.2 &	47.1 &	46.7 &	46.8 &	46.4 &	47.4 &	46.0 &	46.4 &	47.3 &	47.9 &	45.9 &	47.3 &	47.8 &	53.3 &	53.0 &	52.9 &	59.9 \\
HRL &	mean &	57.2 &	56.9 &	56.9 &	56.8 &	56.8 &	56.5 &	57.7 &	55.6 &	56.5 &	57.5 &	58.4 &	55.5 &	57.2 &	58.1 &	62.6 &	62.0 &	62.6 &	65.7 \\
MRL &	mean &	52.1 &	51.8 &	51.9 &	51.5 &	51.8 &	51.3 &	52.5 &	50.6 &	51.2 &	52.2 &	53.2 &	50.5 &	52.1 &	52.9 &	58.1 &	57.7 &	58.0 &	63.3 \\
LRL &	mean &	43.6 &	43.8 &	43.5 &	43.0 &	43.5 &	42.8 &	43.8 &	42.4 &	42.8 &	43.9 &	44.3 &	42.4 &	43.6 &	44.3 &	50.2 &	50.0 &	49.7 &	60.4 \\
URL &	mean &	37.0 &	37.5 &	37.2 &	36.7 &	36.6 &	36.4 &	36.9 &	36.7 &	36.2 &	36.9 &	37.0 &	36.4 &	37.5 &	37.1 &	43.4 &	43.3 &	42.3 &	49.2 \\
OOM &	mean &	32.5 &	32.5 &	32.5 &	31.8 &	32.1 &	31.5 &	33.0 &	31.5 &	31.4 &	32.7 &	33.0 &	31.4 &	32.8 &	32.8 &	37.9 &	37.5 &	37.3 &	- \\
$\Delta_{UL}$ &	URL &	39.0 &	39.4 &	39.1 &	38.6 &	38.6 &	38.3 &	38.9 &	38.4 &	38.2 &	39.0 &	39.2 &	38.2 &	39.3 &	39.3 &	45.4 &	45.3 &	44.5 &	52.6 \\
--- & --- & --- & --- & --- & --- & --- & --- & --- & --- & --- & --- & --- & --- & --- & --- & --- & --- & --- & --- \\
\end{longtable}
}

%% file: main.bbl
\begin{thebibliography}{83}
\providecommand{\natexlab}[1]{#1}
\providecommand{\url}[1]{\texttt{#1}}
\expandafter\ifx\csname urlstyle\endcsname\relax
  \providecommand{\doi}[1]{doi: #1}\else
  \providecommand{\doi}{doi: \begingroup \urlstyle{rm}\Url}\fi

\bibitem[{ Antonov, Alexander}({2022})]{chuvashcorpus}
{ Antonov, Alexander}.
\newblock {Chuvash-Russian parallel corpus}.
\newblock \url{https://github.com/AlAntonov/chv_corpus}, {2022}.

\bibitem[Aharoni et~al.(2019)Aharoni, Johnson, and
  Firat]{aharoni-etal-2019-massively}
Roee Aharoni, Melvin Johnson, and Orhan Firat.
\newblock Massively multilingual neural machine translation.
\newblock In \emph{Proceedings of the 2019 Conference of the North {A}merican
  Chapter of the Association for Computational Linguistics: Human Language
  Technologies, Volume 1 (Long and Short Papers)}, pp.\  3874--3884,
  Minneapolis, Minnesota, June 2019. Association for Computational Linguistics.
\newblock \doi{10.18653/v1/N19-1388}.
\newblock URL \url{https://aclanthology.org/N19-1388}.

\bibitem[Akera et~al.(2022)Akera, Mukiibi, Naggayi, Babirye, Owomugisha,
  Nsumba, Nakatumba-Nabende, Bainomugisha, Mwebaze, and
  Quinn]{akera2022machine}
Benjamin Akera, Jonathan Mukiibi, Lydia~Sanyu Naggayi, Claire Babirye, Isaac
  Owomugisha, Solomon Nsumba, Joyce Nakatumba-Nabende, Engineer Bainomugisha,
  Ernest Mwebaze, and John Quinn.
\newblock Machine translation for african languages: Community creation of
  datasets and models in uganda.
\newblock In \emph{3rd Workshop on African Natural Language Processing}, 2022.
\newblock URL \url{https://openreview.net/forum?id=BK-z5qzEU-9}.

\bibitem[{Andersen, J{\' o}gvan}({2021})]{faroesecorpus}
{Andersen, J{\' o}gvan}.
\newblock {Sprotin Translations}.
\newblock \url{https://github.com/Sprotin/translations}, {2021}.

\bibitem[Ardila et~al.(2020)Ardila, Branson, Davis, Kohler, Meyer, Henretty,
  Morais, Saunders, Tyers, and Weber]{ardila-etal-2020-common}
Rosana Ardila, Megan Branson, Kelly Davis, Michael Kohler, Josh Meyer, Michael
  Henretty, Reuben Morais, Lindsay Saunders, Francis Tyers, and Gregor Weber.
\newblock Common voice: A massively-multilingual speech corpus.
\newblock In \emph{Proceedings of the Twelfth Language Resources and Evaluation
  Conference}, pp.\  4218--4222, Marseille, France, May 2020. European Language
  Resources Association.
\newblock ISBN 979-10-95546-34-4.
\newblock URL \url{https://aclanthology.org/2020.lrec-1.520}.

\bibitem[Arivazhagan et~al.(2019)Arivazhagan, Bapna, Firat, Lepikhin, Johnson,
  Krikun, Chen, Cao, Foster, Cherry, Macherey, Chen, and
  Wu]{arivazhagan2019massively}
Naveen Arivazhagan, Ankur Bapna, Orhan Firat, Dmitry Lepikhin, Melvin Johnson,
  Maxim Krikun, Mia~Xu Chen, Yuan Cao, George~F. Foster, Colin Cherry, Wolfgang
  Macherey, Zhifeng Chen, and Yonghui Wu.
\newblock Massively multilingual neural machine translation in the wild:
  Findings and challenges.
\newblock \emph{CoRR}, abs/1907.05019, 2019.
\newblock URL \url{http://arxiv.org/abs/1907.05019}.

\bibitem[Artetxe \& Schwenk(2018)Artetxe and
  Schwenk]{artetxe-etal-2018-massively}
Mikel Artetxe and Holger Schwenk.
\newblock Massively multilingual sentence embeddings for zero-shot
  cross-lingual transfer and beyond.
\newblock \emph{CoRR}, abs/1812.10464, 2018.
\newblock URL \url{http://arxiv.org/abs/1812.10464}.

\bibitem[Artetxe et~al.(2017)Artetxe, Labaka, Agirre, and
  Cho]{artetxe-etal-2017-unsupervised}
Mikel Artetxe, Gorka Labaka, Eneko Agirre, and Kyunghyun Cho.
\newblock Unsupervised neural machine translation.
\newblock \emph{CoRR}, abs/1710.11041, 2017.
\newblock URL \url{http://arxiv.org/abs/1710.11041}.

\bibitem[Artetxe et~al.(2019)Artetxe, Labaka, and
  Agirre]{artetxe-etal-2019-effective}
Mikel Artetxe, Gorka Labaka, and Eneko Agirre.
\newblock An effective approach to unsupervised machine translation.
\newblock In \emph{Proceedings of the 57th Annual Meeting of the Association
  for Computational Linguistics}, pp.\  194--203, Florence, Italy, July 2019.
  Association for Computational Linguistics.
\newblock \doi{10.18653/v1/P19-1019}.
\newblock URL \url{https://aclanthology.org/P19-1019}.

\bibitem[Bahdanau et~al.(2015)Bahdanau, Cho, and
  Bengio]{bahdanau-etal-2015-neural}
Dzmitry Bahdanau, Kyunghyun Cho, and Yoshua Bengio.
\newblock Neural machine translation by jointly learning to align and
  translate.
\newblock In Yoshua Bengio and Yann LeCun (eds.), \emph{3rd International
  Conference on Learning Representations, {ICLR} 2015, San Diego, CA, USA, May
  7-9, 2015, Conference Track Proceedings}, 2015.
\newblock URL \url{http://arxiv.org/abs/1409.0473}.

\bibitem[Ba{\~n}{\'o}n et~al.(2020)Ba{\~n}{\'o}n, Chen, Haddow, Heafield,
  Hoang, Espl{\`a}-Gomis, Forcada, Kamran, Kirefu, Koehn, Ortiz~Rojas,
  Pla~Sempere, Ram{\'\i}rez-S{\'a}nchez, Sarr{\'\i}as, Strelec, Thompson,
  Waites, Wiggins, and Zaragoza]{banon-etal-2020-paracrawl}
Marta Ba{\~n}{\'o}n, Pinzhen Chen, Barry Haddow, Kenneth Heafield, Hieu Hoang,
  Miquel Espl{\`a}-Gomis, Mikel~L. Forcada, Amir Kamran, Faheem Kirefu, Philipp
  Koehn, Sergio Ortiz~Rojas, Leopoldo Pla~Sempere, Gema
  Ram{\'\i}rez-S{\'a}nchez, Elsa Sarr{\'\i}as, Marek Strelec, Brian Thompson,
  William Waites, Dion Wiggins, and Jaume Zaragoza.
\newblock {P}ara{C}rawl: Web-scale acquisition of parallel corpora.
\newblock In \emph{Proceedings of the 58th Annual Meeting of the Association
  for Computational Linguistics}, pp.\  4555--4567, Online, July 2020.
  Association for Computational Linguistics.
\newblock \doi{10.18653/v1/2020.acl-main.417}.
\newblock URL \url{https://aclanthology.org/2020.acl-main.417}.

\bibitem[{Bapna} et~al.(2022){Bapna}, {Caswell}, {Kreutzer}, {Firat}, {van
  Esch}, {Siddhant}, {Niu}, {Baljekar}, {Garcia}, {Macherey}, {Breiner},
  {Axelrod}, {Riesa}, {Cao}, {Chen}, {Macherey}, {Krikun}, {Wang}, {Gutkin},
  {Shah}, {Huang}, {Chen}, {Wu}, and {Hughes}]{bapna-etal-2022-building}
Ankur {Bapna}, Isaac {Caswell}, Julia {Kreutzer}, Orhan {Firat}, Daan {van
  Esch}, Aditya {Siddhant}, Mengmeng {Niu}, Pallavi {Baljekar}, Xavier
  {Garcia}, Wolfgang {Macherey}, Theresa {Breiner}, Vera {Axelrod}, Jason
  {Riesa}, Yuan {Cao}, Mia~Xu {Chen}, Klaus {Macherey}, Maxim {Krikun}, Pidong
  {Wang}, Alexander {Gutkin}, Apurva {Shah}, Yanping {Huang}, Zhifeng {Chen},
  Yonghui {Wu}, and Macduff {Hughes}.
\newblock {Building Machine Translation Systems for the Next Thousand
  Languages}.
\newblock \emph{arXiv e-prints}, art. arXiv:2205.03983, May 2022.

\bibitem[Baziotis et~al.(2020)Baziotis, Haddow, and
  Birch]{baziotis-etal-2020-language}
Christos Baziotis, Barry Haddow, and Alexandra Birch.
\newblock Language model prior for low-resource neural machine translation.
\newblock In \emph{Proceedings of the 2020 Conference on Empirical Methods in
  Natural Language Processing (EMNLP)}, pp.\  7622--7634, Online, November
  2020. Association for Computational Linguistics.
\newblock \doi{10.18653/v1/2020.emnlp-main.615}.
\newblock URL \url{https://aclanthology.org/2020.emnlp-main.615}.

\bibitem[Caswell et~al.(2019)Caswell, Chelba, and
  Grangier]{caswell-etal-2019-tagged}
Isaac Caswell, Ciprian Chelba, and David Grangier.
\newblock Tagged back-translation.
\newblock In \emph{Proceedings of the Fourth Conference on Machine Translation
  (Volume 1: Research Papers)}, pp.\  53--63, Florence, Italy, August 2019.
  Association for Computational Linguistics.
\newblock \doi{10.18653/v1/W19-5206}.
\newblock URL \url{https://aclanthology.org/W19-5206}.

\bibitem[Caswell et~al.(2020)Caswell, Breiner, van Esch, and
  Bapna]{caswell2020language}
Isaac Caswell, Theresa Breiner, Daan van Esch, and Ankur Bapna.
\newblock Language id in the wild: Unexpected challenges on the path to a
  thousand-language web text corpus, 2020.
\newblock URL \url{https://arxiv.org/abs/2010.14571}.

\bibitem[{Chaudhary} et~al.(2020){Chaudhary}, {Raman}, {Srinivasan}, and
  {Chen}]{chaudhary-etal-2020-dict}
Aditi {Chaudhary}, Karthik {Raman}, Krishna {Srinivasan}, and Jiecao {Chen}.
\newblock {DICT-MLM: Improved Multilingual Pre-Training using Bilingual
  Dictionaries}.
\newblock \emph{arXiv e-prints}, art. arXiv:2010.12566, October 2020.

\bibitem[Cheng et~al.(2021)Cheng, Wang, Jiang, and
  Macherey]{cheng-etal-2021-self}
Yong Cheng, Wei Wang, Lu~Jiang, and Wolfgang Macherey.
\newblock Self-supervised and supervised joint training for resource-rich
  machine translation.
\newblock In Marina Meila and Tong Zhang (eds.), \emph{Proceedings of the 38th
  International Conference on Machine Learning}, volume 139 of
  \emph{Proceedings of Machine Learning Research}, pp.\  1825--1835. PMLR,
  18--24 Jul 2021.
\newblock URL \url{https://proceedings.mlr.press/v139/cheng21b.html}.

\bibitem[Chiruzzo et~al.(2022)Chiruzzo, G{\'o}ngora, Alvarez, Gim{\'e}nez-Lugo,
  Ag{\"u}ero-Torales, and Rodr{\'\i}guez]{chiruzzo-etal-2022-jojajovai}
Luis Chiruzzo, Santiago G{\'o}ngora, Aldo Alvarez, Gustavo Gim{\'e}nez-Lugo,
  Marvin Ag{\"u}ero-Torales, and Yliana Rodr{\'\i}guez.
\newblock Jojajovai: A parallel {G}uarani-{S}panish corpus for {MT}
  benchmarking.
\newblock In \emph{Proceedings of the Thirteenth Language Resources and
  Evaluation Conference}, pp.\  2098--2107, Marseille, France, June 2022.
  European Language Resources Association.
\newblock URL \url{https://aclanthology.org/2022.lrec-1.226}.

\bibitem[Conneau et~al.(2017)Conneau, Lample, Ranzato, Denoyer, and
  J{\'e}gou]{conneau2017word}
Alexis Conneau, Guillaume Lample, Marc'Aurelio Ranzato, Ludovic Denoyer, and
  Herv{\'e} J{\'e}gou.
\newblock Word translation without parallel data.
\newblock \emph{arXiv preprint arXiv:1710.04087}, 2017.

\bibitem[Emezue \& Dossou(2020)Emezue and Dossou]{emezue-dossou-2020-ffr}
Chris~Chinenye Emezue and Femi Pancrace~Bonaventure Dossou.
\newblock {FFR} v1.1: {F}on-{F}rench neural machine translation.
\newblock In \emph{Proceedings of the The Fourth Widening Natural Language
  Processing Workshop}, pp.\  83--87, Seattle, USA, July 2020. Association for
  Computational Linguistics.
\newblock \doi{10.18653/v1/2020.winlp-1.21}.
\newblock URL \url{https://aclanthology.org/2020.winlp-1.21}.

\bibitem[Espl{\`a}-Gomis(2009)]{espla-gomis-2009-bitextor}
Miquel Espl{\`a}-Gomis.
\newblock Bitextor: a free/open-source software to harvest translation memories
  from multilingual websites.
\newblock In \emph{Beyond Translation Memories: New Tools for Translators
  Workshop}, Ottawa, Canada, August 26-30 2009.
\newblock URL \url{https://aclanthology.org/2009.mtsummit-btm.6}.

\bibitem[Fan et~al.(2022)Fan, Bhosale, Schwenk, Ma, El-Kishky, Goyal, Baines,
  Celebi, Wenzek, Chaudhary, Goyal, Birch, Liptchinsky, Edunov, Grave, Auli,
  and Joulin]{m2m-100}
Angela Fan, Shruti Bhosale, Holger Schwenk, Zhiyi Ma, Ahmed El-Kishky,
  Siddharth Goyal, Mandeep Baines, Onur Celebi, Guillaume Wenzek, Vishrav
  Chaudhary, Naman Goyal, Tom Birch, Vitaliy Liptchinsky, Sergey Edunov,
  Edouard Grave, Michael Auli, and Armand Joulin.
\newblock Beyond english-centric multilingual machine translation.
\newblock \emph{J. Mach. Learn. Res.}, 22\penalty0 (1), jul 2022.
\newblock ISSN 1532-4435.

\bibitem[Feldman \& Coto-Solano(2020)Feldman and
  Coto-Solano]{feldman-coto-solano-2020-neural}
Isaac Feldman and Rolando Coto-Solano.
\newblock Neural machine translation models with back-translation for the
  extremely low-resource indigenous language {B}ribri.
\newblock In \emph{Proceedings of the 28th International Conference on
  Computational Linguistics}, pp.\  3965--3976, Barcelona, Spain (Online),
  December 2020. International Committee on Computational Linguistics.
\newblock \doi{10.18653/v1/2020.coling-main.351}.
\newblock URL \url{https://aclanthology.org/2020.coling-main.351}.

\bibitem[Firat et~al.(2016)Firat, Cho, and Bengio]{firat-etal-2016-multi}
Orhan Firat, Kyunghyun Cho, and Yoshua Bengio.
\newblock Multi-way, multilingual neural machine translation with a shared
  attention mechanism.
\newblock In \emph{Proceedings of the 2016 Conference of the North {A}merican
  Chapter of the Association for Computational Linguistics: Human Language
  Technologies}, pp.\  866--875, San Diego, California, June 2016. Association
  for Computational Linguistics.
\newblock \doi{10.18653/v1/N16-1101}.
\newblock URL \url{https://aclanthology.org/N16-1101}.

\bibitem[Freitag et~al.(2022{\natexlab{a}})Freitag, Rei, Mathur, kiu Lo,
  Stewart, Avramidis, Foster, Lavie, and Martins]{freitag-2022-stop}
Markus Freitag, Ricardo Rei, Nitika Mathur, Chi kiu Lo, Craig Stewart,
  Eleftherios Avramidis, George Foster, Alon Lavie, and Andr{\:' e} F.~T.
  Martins.
\newblock Results of wmt22 metrics shared task: Stop using bleu – neural
  metrics are better and more robust.
\newblock 2022{\natexlab{a}}.
\newblock URL \url{https://aclanthology.org/2022.wmt-1.2.pdf}.

\bibitem[Freitag et~al.(2022{\natexlab{b}})Freitag, Vilar, Grangier, Cherry,
  and Foster]{freitag-etal-2022-natural}
Markus Freitag, David Vilar, David Grangier, Colin Cherry, and George Foster.
\newblock A natural diet: Towards improving naturalness of machine translation
  output.
\newblock In \emph{Findings of the Association for Computational Linguistics:
  ACL 2022}, pp.\  3340--3353, Dublin, Ireland, May 2022{\natexlab{b}}.
  Association for Computational Linguistics.
\newblock \doi{10.18653/v1/2022.findings-acl.263}.
\newblock URL \url{https://aclanthology.org/2022.findings-acl.263}.

\bibitem[Ghazvininejad et~al.(2023)Ghazvininejad, Gonen, and
  Zettlemoyer]{dipmt2023}
Marjan Ghazvininejad, Hila Gonen, and Luke Zettlemoyer.
\newblock Dictionary-based phrase-level prompting of large language models for
  machine translation, 2023.
\newblock URL \url{https://arxiv.org/abs/2302.07856}.

\bibitem[Goyal et~al.(2021)Goyal, Gao, Chaudhary, Chen, Wenzek, Ju, Krishnan,
  Ranzato, Guzm\'{a}n, and Fan]{flores101}
Naman Goyal, Cynthia Gao, Vishrav Chaudhary, Peng-Jen Chen, Guillaume Wenzek,
  Da~Ju, Sanjana Krishnan, Marc'Aurelio Ranzato, Francisco Guzm\'{a}n, and
  Angela Fan.
\newblock The flores-101 evaluation benchmark for low-resource and multilingual
  machine translation.
\newblock 2021.

\bibitem[Gulcehre et~al.(2017)Gulcehre, Firat, Xu, Cho, and
  Bengio]{gulcehre-etal-2017-on}
Caglar Gulcehre, Orhan Firat, Kelvin Xu, Kyunghyun Cho, and Yoshua Bengio.
\newblock On integrating a language model into neural machine translation.
\newblock \emph{Computer Speech \& Language}, 45:\penalty0 137--148, 2017.
\newblock ISSN 0885-2308.
\newblock \doi{https://doi.org/10.1016/j.csl.2017.01.014}.
\newblock URL
  \url{https://www.sciencedirect.com/science/article/pii/S0885230816301395}.

\bibitem[Guzm\'{a}n et~al.(2019)Guzm\'{a}n, Chen, Ott, Pino, Lample, Koehn,
  Chaudhary, and Ranzato]{guzman-etal-2019-two}
Francisco Guzm\'{a}n, Peng-Jen Chen, Myle Ott, Juan Pino, Guillaume Lample,
  Philipp Koehn, Vishrav Chaudhary, and Marc'Aurelio Ranzato.
\newblock Two new evaluation datasets for low-resource machine translation:
  Nepali-english and sinhala-english.
\newblock 2019.

\bibitem[Hadgu et~al.(2022)Hadgu, Gebremeskel, and Aregawi]{hornMT}
Asmelash Hadgu, Gebrekirstos Gebremeskel, and Abel Aregawi.
\newblock {HornMT}.
\newblock \url{https://github.com/asmelashteka/HornMT}, 2022.
\newblock Accessed: 2023-03-24.

\bibitem[Heffernan et~al.(2022)Heffernan, Çelebi, and
  Schwenk]{heffernan-etal-2022-bitext}
Kevin Heffernan, Onur Çelebi, and Holger Schwenk.
\newblock Bitext mining using distilled sentence representations for
  low-resource languages, 2022.
\newblock URL \url{https://arxiv.org/abs/2205.12654}.

\bibitem[Jiao et~al.(2023)Jiao, Wang, Huang, Wang, and
  Tu]{jiao-etal-2023-chatgpt}
Wenxiang Jiao, Wenxuan Wang, Jen-tse Huang, Xing Wang, and Zhaopeng Tu.
\newblock Is chatgpt a good translator? a preliminary study, 2023.
\newblock URL \url{https://arxiv.org/abs/2301.08745}.

\bibitem[Johnson et~al.(2017)Johnson, Schuster, Le, Krikun, Wu, Chen, Thorat,
  Vi{\'e}gas, Wattenberg, Corrado, Hughes, and Dean]{johnson-etal-2017-googles}
Melvin Johnson, Mike Schuster, Quoc~V. Le, Maxim Krikun, Yonghui Wu, Zhifeng
  Chen, Nikhil Thorat, Fernanda Vi{\'e}gas, Martin Wattenberg, Greg Corrado,
  Macduff Hughes, and Jeffrey Dean.
\newblock {G}oogle{'}s multilingual neural machine translation system: Enabling
  zero-shot translation.
\newblock \emph{Transactions of the Association for Computational Linguistics},
  5:\penalty0 339--351, 2017.
\newblock \doi{10.1162/tacl_a_00065}.
\newblock URL \url{https://aclanthology.org/Q17-1024}.

\bibitem[Kamholz et~al.(2014)Kamholz, Pool, and
  Colowick]{kamholz-etal-2014-panlex}
David Kamholz, Jonathan Pool, and Susan Colowick.
\newblock {P}an{L}ex: Building a resource for panlingual lexical translation.
\newblock In \emph{Proceedings of the Ninth International Conference on
  Language Resources and Evaluation ({LREC}'14)}, pp.\  3145--3150, Reykjavik,
  Iceland, May 2014. European Language Resources Association (ELRA).
\newblock URL
  \url{http://www.lrec-conf.org/proceedings/lrec2014/pdf/1029_Paper.pdf}.

\bibitem[Khatri et~al.(2021)Khatri, Murthy, Banerjee, and
  Bhattacharyya]{yu2021simple}
Jyotsana Khatri, Rudra Murthy, Tamali Banerjee, and Pushpak Bhattacharyya.
\newblock Simple measures of bridging lexical divergence help unsupervised
  neural machine translation for low-resource languages.
\newblock 2021.
\newblock URL
  \url{https://link.springer.com/article/10.1007/s10590-021-09292-y}.

\bibitem[Kocmi et~al.(2021)Kocmi, Federmann, Grundkiewicz, Junczys-Dowmunt,
  Matsushita, and Menezes]{kocmi-etal-2021-ship}
Tom Kocmi, Christian Federmann, Roman Grundkiewicz, Marcin Junczys-Dowmunt,
  Hitokazu Matsushita, and Arul Menezes.
\newblock To ship or not to ship: An extensive evaluation of automatic metrics
  for machine translation.
\newblock In \emph{Proceedings of the Sixth Conference on Machine Translation},
  pp.\  478--494, Online, November 2021. Association for Computational
  Linguistics.
\newblock URL \url{https://aclanthology.org/2021.wmt-1.57}.

\bibitem[Kreutzer et~al.(2022)Kreutzer, Caswell, Wang, Wahab, van Esch,
  Ulzii-Orshikh, Tapo, Subramani, Sokolov, Sikasote, Setyawan, Sarin, Samb,
  Sagot, Rivera, Rios, Papadimitriou, Osei, Suarez, Orife, Ogueji, Rubungo,
  Nguyen, M{\"u}ller, M{\"u}ller, Muhammad, Muhammad, Mnyakeni, Mirzakhalov,
  Matangira, Leong, Lawson, Kudugunta, Jernite, Jenny, Firat, Dossou, Dlamini,
  de~Silva, {\c{C}}abuk~Ball{\i}, Biderman, Battisti, Baruwa, Bapna, Baljekar,
  Azime, Awokoya, Ataman, Ahia, Ahia, Agrawal, and
  Adeyemi]{kreutzer-etal-2022-quality}
Julia Kreutzer, Isaac Caswell, Lisa Wang, Ahsan Wahab, Daan van Esch,
  Nasanbayar Ulzii-Orshikh, Allahsera Tapo, Nishant Subramani, Artem Sokolov,
  Claytone Sikasote, Monang Setyawan, Supheakmungkol Sarin, Sokhar Samb,
  Beno{\^\i}t Sagot, Clara Rivera, Annette Rios, Isabel Papadimitriou, Salomey
  Osei, Pedro~Ortiz Suarez, Iroro Orife, Kelechi Ogueji, Andre~Niyongabo
  Rubungo, Toan~Q. Nguyen, Mathias M{\"u}ller, Andr{\'e} M{\"u}ller,
  Shamsuddeen~Hassan Muhammad, Nanda Muhammad, Ayanda Mnyakeni, Jamshidbek
  Mirzakhalov, Tapiwanashe Matangira, Colin Leong, Nze Lawson, Sneha Kudugunta,
  Yacine Jernite, Mathias Jenny, Orhan Firat, Bonaventure F.~P. Dossou, Sakhile
  Dlamini, Nisansa de~Silva, Sakine {\c{C}}abuk~Ball{\i}, Stella Biderman,
  Alessia Battisti, Ahmed Baruwa, Ankur Bapna, Pallavi Baljekar, Israel~Abebe
  Azime, Ayodele Awokoya, Duygu Ataman, Orevaoghene Ahia, Oghenefego Ahia,
  Sweta Agrawal, and Mofetoluwa Adeyemi.
\newblock Quality at a glance: An audit of web-crawled multilingual datasets.
\newblock \emph{Transactions of the Association for Computational Linguistics},
  10:\penalty0 50--72, 2022.
\newblock \doi{10.1162/tacl_a_00447}.
\newblock URL \url{https://aclanthology.org/2022.tacl-1.4}.

\bibitem[Kumar et~al.(2022)Kumar, Kumar, and Mishra]{kumar2022dictnmt}
Nalin Kumar, Deepak Kumar, and Subhankar Mishra.
\newblock Dict-nmt: Bilingual dictionary based nmt for extremely low resource
  languages, 2022.

\bibitem[Kuwanto et~al.(2021)Kuwanto, Aky{\"{u}}rek, Tourni, Li, and
  Wijaya]{kuwanto2021low}
Garry Kuwanto, Afra~Feyza Aky{\"{u}}rek, Isidora~Chara Tourni, Siyang Li, and
  Derry Wijaya.
\newblock Low-resource machine translation for low-resource languages:
  Leveraging comparable data, code-switching and compute resources.
\newblock \emph{CoRR}, abs/2103.13272, 2021.
\newblock URL \url{https://arxiv.org/abs/2103.13272}.

\bibitem[Lample et~al.(2017)Lample, Conneau, Denoyer, and
  Ranzato]{lample2017unsupervised}
Guillaume Lample, Alexis Conneau, Ludovic Denoyer, and Marc'Aurelio Ranzato.
\newblock Unsupervised machine translation using monolingual corpora only.
\newblock \emph{arXiv preprint arXiv:1711.00043}, 2017.

\bibitem[Lewis et~al.(2020)Lewis, Liu, Goyal, Ghazvininejad, Mohamed, Levy,
  Stoyanov, and Zettlemoyer]{lewis-etal-2020-bart}
Mike Lewis, Yinhan Liu, Naman Goyal, Marjan Ghazvininejad, Abdelrahman Mohamed,
  Omer Levy, Veselin Stoyanov, and Luke Zettlemoyer.
\newblock {BART}: Denoising sequence-to-sequence pre-training for natural
  language generation, translation, and comprehension.
\newblock In \emph{Proceedings of the 58th Annual Meeting of the Association
  for Computational Linguistics}, pp.\  7871--7880, Online, July 2020.
  Association for Computational Linguistics.
\newblock \doi{10.18653/v1/2020.acl-main.703}.
\newblock URL \url{https://aclanthology.org/2020.acl-main.703}.

\bibitem[Lin et~al.(2021)Lin, Lin, Zhang, and Dai]{lin2021bilingual}
Yusen Lin, Jiayong Lin, Shuaicheng Zhang, and Haoying Dai.
\newblock Bilingual dictionary-based language model pretraining for neural
  machine translation.
\newblock \emph{CoRR}, abs/2103.07040, 2021.
\newblock URL \url{https://arxiv.org/abs/2103.07040}.

\bibitem[Lin et~al.(2020)Lin, Pan, Wang, Qiu, Feng, Zhou, and
  Li]{lin-etal-2020-pre}
Zehui Lin, Xiao Pan, Mingxuan Wang, Xipeng Qiu, Jiangtao Feng, Hao Zhou, and
  Lei Li.
\newblock Pre-training multilingual neural machine translation by leveraging
  alignment information.
\newblock In \emph{Proceedings of the 2020 Conference on Empirical Methods in
  Natural Language Processing (EMNLP)}, pp.\  2649--2663, Online, November
  2020. Association for Computational Linguistics.
\newblock \doi{10.18653/v1/2020.emnlp-main.210}.
\newblock URL \url{https://aclanthology.org/2020.emnlp-main.210}.

\bibitem[Liu et~al.(2020)Liu, Gu, Goyal, Li, Edunov, Ghazvininejad, Lewis, and
  Zettlemoyer]{liu-etal-2020-multilingual-denoising}
Yinhan Liu, Jiatao Gu, Naman Goyal, Xian Li, Sergey Edunov, Marjan
  Ghazvininejad, Mike Lewis, and Luke Zettlemoyer.
\newblock Multilingual denoising pre-training for neural machine translation.
\newblock \emph{Transactions of the Association for Computational Linguistics},
  8:\penalty0 726--742, 2020.
\newblock \doi{10.1162/tacl_a_00343}.
\newblock URL \url{https://aclanthology.org/2020.tacl-1.47}.

\bibitem[Liu et~al.(2021)Liu, Winata, and Fung]{liu-etal-2021-continual}
Zihan Liu, Genta~Indra Winata, and Pascale Fung.
\newblock Continual mixed-language pre-training for extremely low-resource
  neural machine translation.
\newblock In \emph{Findings of the Association for Computational Linguistics:
  ACL-IJCNLP 2021}, pp.\  2706--2718, Online, August 2021. Association for
  Computational Linguistics.
\newblock \doi{10.18653/v1/2021.findings-acl.239}.
\newblock URL \url{https://aclanthology.org/2021.findings-acl.239}.

\bibitem[Mager et~al.(2021)Mager, Oncevay, Ebrahimi, Ortega, Rios, Fan,
  Gutierrez-Vasques, Chiruzzo, Gim{\'e}nez-Lugo, Ramos, Meza~Ruiz, Coto-Solano,
  Palmer, Mager-Hois, Chaudhary, Neubig, Vu, and
  Kann]{mager-etal-2021-findings}
Manuel Mager, Arturo Oncevay, Abteen Ebrahimi, John Ortega, Annette Rios,
  Angela Fan, Ximena Gutierrez-Vasques, Luis Chiruzzo, Gustavo
  Gim{\'e}nez-Lugo, Ricardo Ramos, Ivan~Vladimir Meza~Ruiz, Rolando
  Coto-Solano, Alexis Palmer, Elisabeth Mager-Hois, Vishrav Chaudhary, Graham
  Neubig, Ngoc~Thang Vu, and Katharina Kann.
\newblock Findings of the {A}mericas{NLP} 2021 shared task on open machine
  translation for indigenous languages of the {A}mericas.
\newblock In \emph{Proceedings of the First Workshop on Natural Language
  Processing for Indigenous Languages of the Americas}, pp.\  202--217, Online,
  June 2021. Association for Computational Linguistics.
\newblock \doi{10.18653/v1/2021.americasnlp-1.23}.
\newblock URL \url{https://aclanthology.org/2021.americasnlp-1.23}.

\bibitem[Maheshwari et~al.(2022)Maheshwari, Sharma, Jyothi, and
  Ramakrishnan]{maheshwari-etal-2022-dictdis}
Ayush Maheshwari, Piyush Sharma, Preethi Jyothi, and Ganesh Ramakrishnan.
\newblock Dictdis: Dictionary constrained disambiguation for improved nmt,
  2022.
\newblock URL \url{https://arxiv.org/abs/2210.06996}.

\bibitem[Malon(2021)]{malon2021overcoming}
Christopher Malon.
\newblock Overcoming poor word embeddings with word definitions.
\newblock \emph{CoRR}, abs/2103.03842, 2021.
\newblock URL \url{https://arxiv.org/abs/2103.03842}.

\bibitem[Michon et~al.(2020)Michon, Crego, and
  Senellart]{michon-etal-2020-integrating}
Elise Michon, Josep Crego, and Jean Senellart.
\newblock Integrating domain terminology into neural machine translation.
\newblock In \emph{Proceedings of the 28th International Conference on
  Computational Linguistics}, pp.\  3925--3937, Barcelona, Spain (Online),
  December 2020. International Committee on Computational Linguistics.
\newblock \doi{10.18653/v1/2020.coling-main.348}.
\newblock URL \url{https://aclanthology.org/2020.coling-main.348}.

\bibitem[Mukiibi et~al.(2021)Mukiibi, Claire, and Joyce]{makereremt2021}
Jonathan Mukiibi, Babirye Claire, and Nakatumba-Nabende Joyce.
\newblock {The Makerere MT Corpus: English to Luganda parallel corpus}, May
  2021.
\newblock URL \url{https://doi.org/10.5281/zenodo.5089560}.

\bibitem[Nag et~al.(2020)Nag, Kale, Lakshminarasimhan, and
  Singhavi]{nag2020incorporating}
Sreyashi Nag, Mihir Kale, Varun Lakshminarasimhan, and Swapnil Singhavi.
\newblock Incorporating bilingual dictionaries for low resource semi-supervised
  neural machine translation.
\newblock \emph{CoRR}, abs/2004.02071, 2020.
\newblock URL \url{https://arxiv.org/abs/2004.02071}.

\bibitem[Niehues(2021)]{niehues2021continuous}
Jan Niehues.
\newblock Continuous learning in neural machine translation using bilingual
  dictionaries.
\newblock \emph{CoRR}, abs/2102.06558, 2021.
\newblock URL \url{https://arxiv.org/abs/2102.06558}.

\bibitem[NLLBTeam et~al.(2022)NLLBTeam, Costa-jussà, Cross, Çelebi, Elbayad,
  Heafield, Heffernan, Kalbassi, Lam, Licht, Maillard, Sun, Wang, Wenzek,
  Youngblood, Akula, Barrault, Gonzalez, Hansanti, Hoffman, Jarrett, Sadagopan,
  Rowe, Spruit, Tran, Andrews, Ayan, Bhosale, Edunov, Fan, Gao, Goswami,
  Guzmán, Koehn, Mourachko, Ropers, Saleem, Schwenk, and Wang]{nllb2022}
NLLBTeam, Marta~R. Costa-jussà, James Cross, Onur Çelebi, Maha Elbayad,
  Kenneth Heafield, Kevin Heffernan, Elahe Kalbassi, Janice Lam, Daniel Licht,
  Jean Maillard, Anna Sun, Skyler Wang, Guillaume Wenzek, Al~Youngblood, Bapi
  Akula, Loic Barrault, Gabriel~Mejia Gonzalez, Prangthip Hansanti, John
  Hoffman, Semarley Jarrett, Kaushik~Ram Sadagopan, Dirk Rowe, Shannon Spruit,
  Chau Tran, Pierre Andrews, Necip~Fazil Ayan, Shruti Bhosale, Sergey Edunov,
  Angela Fan, Cynthia Gao, Vedanuj Goswami, Francisco Guzmán, Philipp Koehn,
  Alexandre Mourachko, Christophe Ropers, Safiyyah Saleem, Holger Schwenk, and
  Jeff Wang.
\newblock No language left behind: Scaling human-centered machine translation.
\newblock 2022.

\bibitem[Pan et~al.(2021)Pan, Wang, Wu, and Li]{pan-etal-2021-contrastive}
Xiao Pan, Mingxuan Wang, Liwei Wu, and Lei Li.
\newblock Contrastive learning for many-to-many multilingual neural machine
  translation.
\newblock In \emph{Proceedings of the 59th Annual Meeting of the Association
  for Computational Linguistics and the 11th International Joint Conference on
  Natural Language Processing (Volume 1: Long Papers)}, pp.\  244--258, Online,
  August 2021. Association for Computational Linguistics.
\newblock \doi{10.18653/v1/2021.acl-long.21}.
\newblock URL \url{https://aclanthology.org/2021.acl-long.21}.

\bibitem[Post(2018)]{post-2018-call}
Matt Post.
\newblock A call for clarity in reporting {BLEU} scores.
\newblock In \emph{Proceedings of the Third Conference on Machine Translation:
  Research Papers}, pp.\  186--191, Brussels, Belgium, October 2018.
  Association for Computational Linguistics.
\newblock \doi{10.18653/v1/W18-6319}.
\newblock URL \url{https://aclanthology.org/W18-6319}.

\bibitem[Qin et~al.(2020)Qin, Ni, Zhang, and Che]{ijcai2020p0533}
Libo Qin, Minheng Ni, Yue Zhang, and Wanxiang Che.
\newblock Cosda-ml: Multi-lingual code-switching data augmentation for
  zero-shot cross-lingual nlp.
\newblock In Christian Bessiere (ed.), \emph{Proceedings of the Twenty-Ninth
  International Joint Conference on Artificial Intelligence, {IJCAI-20}}, pp.\
  3853--3860. International Joint Conferences on Artificial Intelligence
  Organization, 7 2020.
\newblock \doi{10.24963/ijcai.2020/533}.
\newblock URL \url{https://doi.org/10.24963/ijcai.2020/533}.
\newblock Main track.

\bibitem[Reid \& Artetxe(2022)Reid and Artetxe]{reid-artetxe-2022-paradise}
Machel Reid and Mikel Artetxe.
\newblock {PARADISE}: Exploiting parallel data for multilingual
  sequence-to-sequence pretraining.
\newblock In \emph{Proceedings of the 2022 Conference of the North American
  Chapter of the Association for Computational Linguistics: Human Language
  Technologies}, pp.\  800--810, Seattle, United States, July 2022. Association
  for Computational Linguistics.
\newblock \doi{10.18653/v1/2022.naacl-main.58}.
\newblock URL \url{https://aclanthology.org/2022.naacl-main.58}.

\bibitem[Resnik \& Smith(2003)Resnik and Smith]{resnik-smith-2003-web}
Philip Resnik and Noah~A. Smith.
\newblock The web as a parallel corpus.
\newblock \emph{Computational Linguistics}, 29\penalty0 (3):\penalty0 349--380,
  2003.
\newblock \doi{10.1162/089120103322711578}.
\newblock URL \url{https://aclanthology.org/J03-3002}.

\bibitem[Schwenk et~al.(2021{\natexlab{a}})Schwenk, Chaudhary, Sun, Gong, and
  Guzm{\'a}n]{schwenk-etal-2021-wikimatrix}
Holger Schwenk, Vishrav Chaudhary, Shuo Sun, Hongyu Gong, and Francisco
  Guzm{\'a}n.
\newblock {W}iki{M}atrix: Mining 135{M} parallel sentences in 1620 language
  pairs from {W}ikipedia.
\newblock In \emph{Proceedings of the 16th Conference of the European Chapter
  of the Association for Computational Linguistics: Main Volume}, pp.\
  1351--1361, Online, April 2021{\natexlab{a}}. Association for Computational
  Linguistics.
\newblock \doi{10.18653/v1/2021.eacl-main.115}.
\newblock URL \url{https://aclanthology.org/2021.eacl-main.115}.

\bibitem[Schwenk et~al.(2021{\natexlab{b}})Schwenk, Wenzek, Edunov, Grave,
  Joulin, and Fan]{schwenk-etal-2021-ccmatrix}
Holger Schwenk, Guillaume Wenzek, Sergey Edunov, Edouard Grave, Armand Joulin,
  and Angela Fan.
\newblock {CCM}atrix: Mining billions of high-quality parallel sentences on the
  web.
\newblock In \emph{Proceedings of the 59th Annual Meeting of the Association
  for Computational Linguistics and the 11th International Joint Conference on
  Natural Language Processing (Volume 1: Long Papers)}, pp.\  6490--6500,
  Online, August 2021{\natexlab{b}}. Association for Computational Linguistics.
\newblock \doi{10.18653/v1/2021.acl-long.507}.
\newblock URL \url{https://aclanthology.org/2021.acl-long.507}.

\bibitem[Sellam et~al.(2020)Sellam, Das, and Parikh]{sellam-2020-bleurt}
Thibault Sellam, Dipanjan Das, and Ankur~P. Parikh.
\newblock {BLEURT:} learning robust metrics for text generation.
\newblock \emph{CoRR}, abs/2004.04696, 2020.
\newblock URL \url{https://arxiv.org/abs/2004.04696}.

\bibitem[Sennrich et~al.(2016)Sennrich, Haddow, and
  Birch]{sennrich-etal-2016-improving}
Rico Sennrich, Barry Haddow, and Alexandra Birch.
\newblock Improving neural machine translation models with monolingual data.
\newblock In \emph{Proceedings of the 54th Annual Meeting of the Association
  for Computational Linguistics (Volume 1: Long Papers)}, pp.\  86--96, Berlin,
  Germany, August 2016. Association for Computational Linguistics.
\newblock \doi{10.18653/v1/P16-1009}.
\newblock URL \url{https://aclanthology.org/P16-1009}.

\bibitem[Shakirov \& Kunafin(2023)Shakirov and Kunafin]{bashkircorpus}
Iskander Shakirov and Aigiz Kunafin.
\newblock Bashkir-russian parallel corpora.
\newblock 2023.

\bibitem[Siddhant et~al.(2020)Siddhant, Bapna, Cao, Firat, Chen, Kudugunta,
  Arivazhagan, and Wu]{siddhant-etal-2020-leveraging}
Aditya Siddhant, Ankur Bapna, Yuan Cao, Orhan Firat, Mia Chen, Sneha Kudugunta,
  Naveen Arivazhagan, and Yonghui Wu.
\newblock Leveraging monolingual data with self-supervision for multilingual
  neural machine translation.
\newblock In \emph{Proceedings of the 58th Annual Meeting of the Association
  for Computational Linguistics}, pp.\  2827--2835, Online, July 2020.
  Association for Computational Linguistics.
\newblock \doi{10.18653/v1/2020.acl-main.252}.
\newblock URL \url{https://aclanthology.org/2020.acl-main.252}.

\bibitem[Siddhant et~al.(2022)Siddhant, Bapna, Firat, Cao, Chen, Caswell, and
  Garcia]{siddhant-etal-2022-towards}
Aditya Siddhant, Ankur Bapna, Orhan Firat, Yuan Cao, Mia~Xu Chen, Isaac
  Caswell, and Xavier Garcia.
\newblock Towards the next 1000 languages in multilingual machine translation:
  Exploring the synergy between supervised and self-supervised learning.
\newblock \emph{CoRR}, abs/2201.03110, 2022.
\newblock URL \url{https://arxiv.org/abs/2201.03110}.

\bibitem[Song et~al.(2019{\natexlab{a}})Song, Zhang, Yu, Luo, Wang, and
  Zhang]{song-etal-2019-code}
Kai Song, Yue Zhang, Heng Yu, Weihua Luo, Kun Wang, and Min Zhang.
\newblock Code-switching for enhancing {NMT} with pre-specified translation.
\newblock In \emph{Proceedings of the 2019 Conference of the North {A}merican
  Chapter of the Association for Computational Linguistics: Human Language
  Technologies, Volume 1 (Long and Short Papers)}, pp.\  449--459, Minneapolis,
  Minnesota, June 2019{\natexlab{a}}. Association for Computational
  Linguistics.
\newblock \doi{10.18653/v1/N19-1044}.
\newblock URL \url{https://aclanthology.org/N19-1044}.

\bibitem[Song et~al.(2019{\natexlab{b}})Song, Tan, Qin, Lu, and
  Liu]{song-etal-2019-mass}
Kaitao Song, Xu~Tan, Tao Qin, Jianfeng Lu, and Tie-Yan Liu.
\newblock {MASS}: Masked sequence to sequence pre-training for language
  generation.
\newblock In Kamalika Chaudhuri and Ruslan Salakhutdinov (eds.),
  \emph{Proceedings of the 36th International Conference on Machine Learning},
  volume~97 of \emph{Proceedings of Machine Learning Research}, pp.\
  5926--5936. PMLR, 09--15 Jun 2019{\natexlab{b}}.
\newblock URL \url{https://proceedings.mlr.press/v97/song19d.html}.

\bibitem[Susanto et~al.(2020)Susanto, Chollampatt, and
  Tan]{susanto2020lexically}
Raymond~Hendy Susanto, Shamil Chollampatt, and Liling Tan.
\newblock Lexically constrained neural machine translation with levenshtein
  transformer.
\newblock \emph{CoRR}, abs/2004.12681, 2020.
\newblock URL \url{https://arxiv.org/abs/2004.12681}.

\bibitem[Sutskever et~al.(2014)Sutskever, Vinyals, and
  Le]{sutskever-etal-2014-sequence}
Ilya Sutskever, Oriol Vinyals, and Quoc~V Le.
\newblock Sequence to sequence learning with neural networks.
\newblock In Z.~Ghahramani, M.~Welling, C.~Cortes, N.~Lawrence, and K.Q.
  Weinberger (eds.), \emph{Advances in Neural Information Processing Systems},
  volume~27. Curran Associates, Inc., 2014.
\newblock URL
  \url{https://proceedings.neurips.cc/paper/2014/file/a14ac55a4f27472c5d894ec1c3c743d2-Paper.pdf}.

\bibitem[Sutton(2019)]{sutton_2019}
Richard Sutton.
\newblock The bitter lesson.
\newblock Mar 2019.
\newblock
  \doi{https://www.cs.utexas.edu/~eunsol/courses/data/bitter_lesson.pdf}.

\bibitem[Tiedemann(2020)]{tatoeba}
J{\"{o}}rg Tiedemann.
\newblock The tatoeba translation challenge - realistic data sets for low
  resource and multilingual {MT}.
\newblock \emph{CoRR}, abs/2010.06354, 2020.
\newblock URL \url{https://arxiv.org/abs/2010.06354}.

\bibitem[{Tlisha, Nart}({2022})]{abkhazcorpus}
{Tlisha, Nart}.
\newblock {Multilingual Parallel Corpus}.
\newblock \url{https://github.com/danielinux7/Multilingual-Parallel-Corpus},
  {2022}.
\newblock Accessed: 2023-03-24.

\bibitem[Uszkoreit et~al.(2010)Uszkoreit, Ponte, Popat, and
  Dubiner]{uszkoreit-etal-2010-large}
Jakob Uszkoreit, Jay Ponte, Ashok Popat, and Moshe Dubiner.
\newblock Large scale parallel document mining for machine translation.
\newblock In \emph{Proceedings of the 23rd International Conference on
  Computational Linguistics (Coling 2010)}, pp.\  1101--1109, Beijing, China,
  August 2010. Coling 2010 Organizing Committee.
\newblock URL \url{https://aclanthology.org/C10-1124}.

\bibitem[Vaswani et~al.(2017)Vaswani, Shazeer, Parmar, Uszkoreit, Jones, Gomez,
  Kaiser, and Polosukhin]{vaswani-etal-2017-attention}
Ashish Vaswani, Noam Shazeer, Niki Parmar, Jakob Uszkoreit, Llion Jones,
  Aidan~N Gomez, \L~ukasz Kaiser, and Illia Polosukhin.
\newblock Attention is all you need.
\newblock In I.~Guyon, U.~Von Luxburg, S.~Bengio, H.~Wallach, R.~Fergus,
  S.~Vishwanathan, and R.~Garnett (eds.), \emph{Advances in Neural Information
  Processing Systems}, volume~30. Curran Associates, Inc., 2017.
\newblock URL
  \url{https://proceedings.neurips.cc/paper/2017/file/3f5ee243547dee91fbd053c1c4a845aa-Paper.pdf}.

\bibitem[Wang et~al.(2022)Wang, Ruder, and Neubig]{wang-etal-2022-expanding}
Xinyi Wang, Sebastian Ruder, and Graham Neubig.
\newblock Expanding pretrained models to thousands more languages via
  lexicon-based adaptation.
\newblock In \emph{Proceedings of the 60th Annual Meeting of the Association
  for Computational Linguistics (Volume 1: Long Papers)}, pp.\  863--877,
  Dublin, Ireland, May 2022. Association for Computational Linguistics.
\newblock \doi{10.18653/v1/2022.acl-long.61}.
\newblock URL \url{https://aclanthology.org/2022.acl-long.61}.

\bibitem[Wanjawa et~al.(2022)Wanjawa, Wanzare, Indede, McOnyango, Ombui, and
  Muchemi]{wanjawa2022kencorpus}
Barack Wanjawa, Lilian Wanzare, Florence Indede, Owen McOnyango, Edward Ombui,
  and Lawrence Muchemi.
\newblock Kencorpus: A kenyan language corpus of swahili, dholuo and luhya for
  natural language processing tasks, 2022.

\bibitem[Wu et~al.(2016)Wu, Schuster, Chen, Le, Norouzi, Macherey, Krikun, Cao,
  Gao, Macherey, Klingner, Shah, Johnson, Liu, Łukasz Kaiser, Gouws, Kato,
  Kudo, Kazawa, Stevens, Kurian, Patil, Wang, Young, Smith, Riesa, Rudnick,
  Vinyals, Corrado, Hughes, and Dean]{wu-etal-2016-google}
Yonghui Wu, Mike Schuster, Zhifeng Chen, Quoc~V. Le, Mohammad Norouzi, Wolfgang
  Macherey, Maxim Krikun, Yuan Cao, Qin Gao, Klaus Macherey, Jeff Klingner,
  Apurva Shah, Melvin Johnson, Xiaobing Liu, Łukasz Kaiser, Stephan Gouws,
  Yoshikiyo Kato, Taku Kudo, Hideto Kazawa, Keith Stevens, George Kurian,
  Nishant Patil, Wei Wang, Cliff Young, Jason Smith, Jason Riesa, Alex Rudnick,
  Oriol Vinyals, Greg Corrado, Macduff Hughes, and Jeffrey Dean.
\newblock Google's neural machine translation system: Bridging the gap between
  human and machine translation.
\newblock \emph{CoRR}, abs/1609.08144, 2016.
\newblock URL \url{http://arxiv.org/abs/1609.08144}.

\bibitem[Xia et~al.(2019)Xia, Kong, Anastasopoulos, and
  Neubig]{xia2019generalized}
Mengzhou Xia, Xiang Kong, Antonios Anastasopoulos, and Graham Neubig.
\newblock Generalized data augmentation for low-resource translation.
\newblock \emph{CoRR}, abs/1906.03785, 2019.
\newblock URL \url{http://arxiv.org/abs/1906.03785}.

\bibitem[Yang et~al.(2021)Yang, Yin, Ma, Huang, Zhang, Li, and
  Wei]{yang-etal-2021-multilingual}
Jian Yang, Yuwei Yin, Shuming Ma, Haoyang Huang, Dongdong Zhang, Zhoujun Li,
  and Furu Wei.
\newblock Multilingual agreement for multilingual neural machine translation.
\newblock In \emph{Proceedings of the 59th Annual Meeting of the Association
  for Computational Linguistics and the 11th International Joint Conference on
  Natural Language Processing (Volume 2: Short Papers)}, pp.\  233--239,
  Online, August 2021. Association for Computational Linguistics.
\newblock \doi{10.18653/v1/2021.acl-short.31}.
\newblock URL \url{https://aclanthology.org/2021.acl-short.31}.

\bibitem[Yang et~al.(2020)Yang, Hu, Han, Huang, and Ju]{yang-etal-2020-csp}
Zhen Yang, Bojie Hu, Ambyera Han, Shen Huang, and Qi~Ju.
\newblock {CSP}:code-switching pre-training for neural machine translation.
\newblock In \emph{Proceedings of the 2020 Conference on Empirical Methods in
  Natural Language Processing (EMNLP)}, pp.\  2624--2636, Online, November
  2020. Association for Computational Linguistics.
\newblock \doi{10.18653/v1/2020.emnlp-main.208}.
\newblock URL \url{https://aclanthology.org/2020.emnlp-main.208}.

\bibitem[Yu et~al.(2021)Yu, Zhu, Fang, Yu, Wang, Xu, Zeng, and
  Jiang]{yu2021dictbert}
Wenhao Yu, Chenguang Zhu, Yuwei Fang, Donghan Yu, Shuohang Wang, Yichong Xu,
  Michael Zeng, and Meng Jiang.
\newblock Dict-bert: Enhancing language model pre-training with dictionary.
\newblock \emph{CoRR}, abs/2110.06490, 2021.
\newblock URL \url{https://arxiv.org/abs/2110.06490}.

\bibitem[Zhong \& Chiang(2020)Zhong and Chiang]{xing2020look}
Xing~Jie Zhong and David Chiang.
\newblock Look it up: Bilingual and monolingual dictionaries improve neural
  machine translation.
\newblock \emph{CoRR}, abs/2010.05997, 2020.
\newblock URL \url{https://arxiv.org/abs/2010.05997}.

\end{thebibliography}
